\documentclass[sigconf,nonacm]{acmart}
\usepackage{times}  % DO NOT CHANGE THIS
\usepackage{helvet}  % DO NOT CHANGE THIS
\usepackage{courier}  % DO NOT CHANGE THIS

\usepackage{graphicx} % DO NOT CHANGE THIS
\usepackage{natbib}  % DO NOT CHANGE THIS AND DO NOT ADD ANY OPTIONS TO IT
\usepackage{caption} % DO NOT CHANGE THIS AND DO NOT ADD ANY OPTIONS TO IT
\usepackage{algorithm}
\usepackage{algorithmic}
\usepackage{appendix}
\usepackage{tikz}
\usetikzlibrary{positioning,calc}
\usepackage{newfloat}
\usepackage{listings}
\DeclareCaptionStyle{ruled}{labelfont=normalfont,labelsep=colon,strut=off} % DO NOT CHANGE THIS
\floatstyle{ruled}
\newfloat{listing}{tb}{lst}{}
\floatname{listing}{Listing}
\usepackage{times}
\usepackage{soul}
\usepackage{url}
\usepackage{stackengine}
\usepackage{subcaption}
\usepackage{xcolor}
\newcommand{\independent}{\perp \!\!\! \perp}
\usepackage{amsmath,bm}
\newcommand{\commentout}[1]{%
}
\newcommand{\indep}{\perp \!\!\! \perp}
\newcommand{\fixme}[1]{{\textcolor{red}{\textit{#1}}}}
\newtheorem{problem}{Problem}

\title{Contagion Effect Estimation Using Proximal Embeddings}
\author{Zahra Fatemi}
\email{zfatem2@uic.edu}
\affiliation{%
  \institution{Department of Computer Science\\University of Illinois Chicago}
  \country{}
}
\author{Elena Zheleva}
\email{ezheleva@uic.edu}
\affiliation{%
  \institution{Department of Computer Science\\University of Illinois Chicago}
  \country{}
}

\setcopyright{none}
\setcopyright{none}
\begin{document}

\begin{abstract}

Contagion effect refers to the causal effect of peers' behavior on the outcome of an individual in social networks. Contagion can be confounded due to latent homophily which makes contagion effect estimation very hard: nodes in a homophilic network tend to have ties to peers with similar attributes and can behave similarly without influencing one another. One way to account for latent homophily is by considering proxies for the unobserved confounders. However, as we demonstrate in this paper, existing proxy-based methods for contagion effect estimation have a very high variance when the proxies are high-dimensional. To address this issue, we introduce a novel framework, \textit{Proximal Embeddings (ProEmb)}, that integrates variational autoencoders with adversarial networks to create low-dimensional representations of high-dimensional proxies and help with identifying contagion effects. While VAEs have been used previously for representation learning in causal inference, a novel aspect of our approach is the additional component of adversarial networks to balance the representations of different treatment groups, which is essential in causal inference from observational data where these groups typically come from different distributions. We empirically show that our method significantly increases the accuracy and reduces the variance of contagion effect estimation in observational network data compared to state-of-the-art methods.

% aaai 24:   Contagion effect refers to the causal effect of peers' behavior on the outcome of an individual in social networks. While prominent methods for estimating contagion effects in observational studies often assume that there are no unmeasured confounders, contagion can be confounded due to latent homophily: nodes in a homophilic network tend to have ties to peers with similar attributes and can behave similarly without influencing one another. One way to account for latent homophily is by considering proxies for the unobserved confounders. However, in the presence of high-dimensional proxies, proxy-based methods can lead to substantially biased estimation of contagion effects, as we demonstrate in this paper. To tackle this issue, we introduce the novel \textit{Proximal Embeddings (ProEmb)}, a framework that integrates Variational Autoencoders (VAEs) and adversarial networks to generate balanced low-dimensional representations of high-dimensional proxies for different treatment groups and identifies contagion effects in the presence of unobserved network confounders. We empirically show that our method significantly increases the accuracy of contagion effect estimation in observational network data compared to state-of-the-art methods.
\end{abstract}

\maketitle

\section{Introduction}
The goal of causal inference is to estimate the effect of an intervention on individuals' outcomes. Traditionally, causal inference has relied on the assumption of no interference, which states that any individual’s response to treatment depends only on their own treatment and not on the treatment of others.
However, individuals can impact each other through their interactions. Contagion is a type of interference that is defined as the influence of neighbors' actions on the actions of an individual.
Contagion effect estimation plays a central role in understanding how social environments shape personal actions, behavior, and attitudes \citep{fowler-jhe08,bramoulle-joe09,eckles-pnas16}.  
%Some applications of contagion effect estimation in real-world domains include the spread of obesity from person to person \citep{christakis-njm07,krauth-cje05}, the spread of person-to-person smoking behavior and quitting decisions \citep{christakis-njm08}, understanding and quantifying the role of one’s social environment in trusting fake news in online social media \citep{torres-sig18,sterrett-dj19}, the influence of peers on new product adoption in online marketing \citep{bailey-nber19}, and the influence of peers on education outcomes in online learning systems \citep{lai-det19}.
Some real-world applications of contagion effect estimation include studying the spread of obesity from one person to another \citep{christakis-njm07,krauth-cje05}, analyzing the spread of smoking behavior among peers \citep{christakis-njm08}, and understanding the role of one's social environment in trusting fake news on online social media platforms \citep{torres-sig18}.
%, and assessing the influence of peers on educational outcomes in online learning systems \citep{lai-det19}.

Despite their importance, identification (i.e. determining causal estimands based on observational data distribution \citep{pearl-book09}) and estimation of contagion effects are challenging due to the presence of unmeasured confounders, variables that affect both the treatment and the outcome of interest. 
%For instance, an individual's political inclination  (e.g., liberal or conservative) can act as an unobserved confounder influencing the opinion of users toward policies related to social distancing.
%In the running example, race can influence both the political affiliation party and the voting decisions of people. 
 A common source of confounding in networks is latent homophily
 \citep{shalizi-smr11,manski-jres93,vanderweele-cafsr13,ogburn-arxiv17}, the tendency of ties to form between individuals with similar unobserved attributes. 
 When contagion effects are confounded with latent homophily, it is hard to tell if any changes in the individual's outcome are the result of neighbors' influence or the similarity between the individual and neighbors characteristics. For example, people with similar political affiliations would be more likely to interact on social media (e.g., Twitter) and they may express similar opinions (e.g., agree or disagree with social distancing policies during a pandemic), not because one influences the other but because they share similar political views in the first place.
%In some networks contagion is confounded with latent homophily, making it hard to estimate it from observational data.
%In a homophilous network, a node is more likely to be similar to its neighbors.
%Nodes in a homophilous network tend to have ties to nodes with similar attributes.

To identify and estimate contagion effects in the presence of unobserved confounders, existing approaches look for observed variables that can be considered as valid proxies of the unobserved confounders \citep{miao-biometrika18,tchetgen-arxiv20,egami-zrxiv2021}. 
%A recent study proposed a method that utilizes two sets of negative control proxies to nonparametrically identify contagion effects in the presence of unmeasured confounders \citep{egami-zrxiv2021}. 
%\citep{egami-zrxiv2021} employs two sets of negative control proxies to nonparametrically identify contagion effects in the presence of unmeasured confounders. 
However, such approaches can perform poorly on real-world observational data, 
%with high-dimensional proxies,
%\citep{deaner-arxiv21},
such as web and social media, in which a high-dimensional covariate space is the norm. High-dimensional control proxies (e.g., tweet words of a user) lead to a sparse vector of model parameters and 
higher asymptotic bias and variance of the estimation \citep{de-oup11}. 
Another source of variance is selection bias \cite{guo-wsdm20,shalit-pml17,assaad-pmlr21,johansson-icml16}. Selection bias occurs when there is a mismatch in attribute distribution between the treatment and control groups. For instance, a treatment group can comprise mostly individuals who prioritize their health and have friends who follow social distancing guidelines, while the control group comprises of individuals who do not prioritize their health and have friends who largely disregard social distancing measures. A common method for dealing with selection bias in observational studies is matching, where a balanced sample is created by identifying similar units from the opposite treatment group.
%treatment and control nodes are matched based on their similarity of attributes ~\cite{stuart-ss10}
%Matching methods create a balanced sample by searching for “similar” units from the opposite treatment group
However, matching tends to encounter scalability issues when applied to high-dimensional data \citep{abadie-eco06,assaad-pmlr21}.

%this approach applies to low-dimensional settings with a small number of proxies and confounders. 
%\textbf{Present Work.} 
To address high-dimensionality and selection bias in real-world contagion estimation settings, we introduce \textit{\textbf{Pro}ximal \textbf{Emb}eddings (ProEmb)}, a framework for inferring contagion effects in homophilic networks. ProEmb learns embeddings of high-dimensional proxies for unobserved confounders.
It combines variational autoencoders (VAEs) and adversarial networks \citep{goodfellow-neurips14,mescheder-pmlr17} to map high-dimensional proxies to a probability distribution over the latent space
with the goal of obtaining a balanced low-dimensional proxy representation. While the use of VAEs for causal effect estimation is not new ~\cite{louizos-neurips17,grari-ijc22,kim-aaai21,im-arxiv21}, our framework has two novel components. The first one is in defining and developing the first solution to the problem of contagion estimation with high-dimensional proxies, an important problem in real-world contagion estimation scenarios. The second one is the novel enhancement of VAEs with adversarial networks, similar to matching ~\cite{stuart-ss10}, which play the important role of addressing the selection bias in treatment groups and is of independent interest for causal effect estimation beyond contagion. In addition to being meaningful for causal inference, this enhancement is crucial for the empirical performance of the estimator.
%To the best of our knowledge, this work is the first that integrates VAEs and adversarial networks to improve contagion effect estimation in networks with unobserved confounders.
%For the counterfactual model, we use a meta-learner ~\citep{kunzel-nas19} to train two different outcome models for treatment and control nodes. 
%While the idea of using proxies for estimating contagion effects is not new \cite{egami-zrxiv2021}, making this idea practical for high-dimensional applications suffering from selection bias is novel.

Our framework consists of three main components: 1) an embedding learning component that maps high-dimensional proxies to low-dimensional latent representations, 2) a representation balancing component that addresses the representation mismatch between treatment and control groups in the learned latent space, and 3) a counterfactual learning component, where an outcome model is trained to infer the potential outcomes of nodes based on the treatment and the low-dimensional representation of proxies. 
The primary goal of dimensionality reduction is to reduce bias and variance in the estimation by decreasing sparsity and the number of model parameters that need to be estimated.

\textbf{Key idea and highlights.} To summarize, this paper makes the following main contributions:
\begin{itemize}
\item We formulate the problem of contagion effect estimation in networks when latent homophily and high-dimensional proxies of latent confounders are present.
\item We propose a novel framework for contagion effect estimation, ProEmb, that integrates VAEs, adversarial networks, and meta-learners for balanced representation learning and contagion effect estimation.
%Our framework maps high-dimensional proxies to a balanced low-dimensional latent space distribution and utilizes meta-learners for contagion effect estimation.
\item Through empirical analysis, we demonstrate that existing methods for inferring contagion effects using high-dimensional proxies are prone to high bias and variance, while our proposed approach exhibits remarkable performance improvements over state-of-the-art techniques.
%\item 
\end{itemize}

\commentout{
While the idea of using proxies for estimating contagion effects is not new,
making this idea practical for high-dimensional social media applications is novel. We achieve this by developing a framework that finds the embedding of proxies by mapping high-dimensional proxies to low-dimensional latent space and using meta-learners to estimate contagion effects.}

%The rest of the paper is structured as follows. Section 2 provides background for causal effect estimation and rela- tional d-separation. Sections 3 and 4 introduce Relational Covariate Adjustment and discuss practical issues of imple- mentation. Section 5 compares the estimates of RCA to es- timates obtained via experimentation using multiple graph structures with data simulated under multiple functional forms, and shows that the performance of RCA can be com- petitive with experimental results.

%identifying observed proxies for the unobserved confounders and assume 
%covariate measurements are at best proxies
\section{Related Work}
\label{related}
%Causal inference in networks has attracted significant attention in recent years \citep{mcfowland-jasa21,ogburn-cspss18}. 
 
Here, we review prior studies that focus on causal inference in observational network data.
%mitigate confounding bias to infer causal effect in networks.
%\textbf{Contagion effect estimation in networks.} 
%Graphical models provide a way to represent interdependency between the variables in causal inference studies.
%have shed light on the identification of causal effects in many settings. 
 \citet{ogburn-ss14} study the role of structural causal models in causal effect estimation in the presence of different types of interference. 
%and deploy them determine the adjustment set to make the causal effect of interest identifiable.
%describe how to draw graphs in the presence of different types of interference and how to use these graphs to determine the adjustment set to make the causal effect of interest identifiable.
 \citet{shalizi-smr11} show that in networks formed by latent homophily, contagion, and homophily can be confounded and the causal effect is not always identifiable. 
Another study shows that controlling for the cluster assignment of nodes can help \citep{shalizi-arxiv16}. 
%\citep{toulis-mlr13}
%\textbf{Causal inference using proxy variables.}
A recent study deploys negative control outcome and exposure variables to estimate contagion effects in low-dimensional settings \citep{egami-zrxiv2021}. Our work
builds upon this work and focuses on estimating contagion effects in datasets with high-dimensional proxies.

%\textbf{Causal inference using embeddings.}
%in i.i.d data \citep{louizos-neurips17}
Recently, a series of methods have been proposed to leverage representation learning to relax the strong ignorability assumption in networked data.
%\citep{mcfowland-jasa21} propose a two-stage approach to measure the contagion in networks generated by a stochastic block model or a latent space model
\citet{guo-wsdm20} propose the Network Deconfounder framework where network structure and the observed features of nodes are mapped to a latent representation space to capture the influence of hidden confounders. \citet{veitch-neurips20} propose a procedure for estimating treatment effects using network embeddings by
reducing the causal estimation problem to a semi-supervised prediction of the treatments and outcomes and using embedding models for the semi-supervised prediction. \citet{cristali-neurips21} use node embeddings learned from the network structure for estimating contagion effects in a different setting where covariates and the network structure are unobserved. However, these works either do not consider interference \cite{guo-wsdm20, veitch-neurips20}, or selection bias \cite{cristali-neurips21}.
%This work presents a different approach from the method described in this paper. 
%However, these studies assume no contagion effects in their experiments.
%\fixme{Some studies propose methodologies for measuring direct interference (i.e, the causal effect of neighbor's treatment on the node's outcome) which is a different causal estimand from the estimand of this paper \citep{ma-kdd22,jiang-icikm22}.}

%Some of few studies deploy embedding techniques to  

%\textbf{Representation balancing.}
%A line of research leverages weighting-based methods to balance covariates distribution \cite{assaad-pmlr21,li-ijb13,crump-bio09,li-jasa18}.
%In weighting method, such as Inverse Probability Weighting (IPW), different importance weights are assigned to nodes with the objective of aligning covariate distributions across various treatment groups. 
Methods to improve the distribution mismatch between treatment groups include combining weighting with representation learning \citep{hassanpour-ijcai19,li-neurips17,guo-wsdm20}, linear ridge regression with representation learning \citep{jiang-icikm22}, and a discriminator component that balances the representation of the observed confounders \citep{jiang-icikm22}. %Our method is different is a way that it balances the respresentation generated by VAEs from high dimensional proxies of the unobserved confoundeds.
Our approach is distinct in that it balances the proxy representations generated by VAEs with adversarial networks.

%In contrast, our approach accounts for unobserved confounders in contagion estimation.
%(I wasn't sure what is distinct about their model? are they doing contagion?)

%A different line of research combines weighting with representation learning to improve the distribution mismatch in causal effect estimation \cite{hassanpour-ijcai19,li-neurips17,guo-wsdm20}.
%Guo et al. introduce an optimization function that integrates the minimization of IPM in the representation learning component \cite{guo-wsdm20}.
% \citet{shalit-pml17} 
%propose a data representation learning model that improves selection bias by fitting employing linear ridge regression and enforcing a relative error bound to address representation mismatch. 
% \citet{jiang-icikm22} propose the NetEst framework, which targets the representation learning for observed confounders. This approach integrates a discriminator component to tackle representation mismatch. It is worth mentioning that their methodology differs from ours in two key aspects. First, they assume that there are no unmeasured confounders, whereas our approach accounts for their presence. Second, they employ a distinct causal model that differs from the one presented in this paper.

Several studies have utilized VAEs to estimate proxies for confounding variables in non-network data. Louizos et al. \cite{louizos-neurips17} leverage VAEs for the purpose of inferring latent variables proxies that help with estimating individual treatment effects. Grari et al. \cite{grari-ijc22} integrate VAEs with an adversarial training component aimed at acquiring a proxy for latent sensitive information, such as gender. Their approach differs from our framework in the sense that adversarial training focuses on guaranteeing the independence of the generated latent space from the unobserved sensitive variable. In contrast, our approach utilizes the discriminator component of an adversarial network to achieve a balance in the representation of treatment and control groups.
\section{Problem Description}
\label{problem}
In this section, we present our data and causal model, the causal estimand, different types of proxy variables, and issues with contagion effect estimation in high-dimensional settings.
\subsection{Data model}
We assume a graph $G=(\mathbf{V},\mathbf{E})$ that consists of a set of $|\mathbf{V}|$ nodes and a set of edges $\mathbf{E}=\{ e_{ij}\}$, where $e_{ij}$ denotes that there is an edge between node $v_i\in \mathbf{V}$ and node $v_j\in \mathbf{V}$.
Each node has an observed $n$-dimensional vector of attributes, $\mathbf{Z}_i$, unobserved characteristics, $\mathbf{U}_i$, and outcomes in two consecutive time steps, $Y_{i,t-1} \in \{0,1\}$, and $Y_{i,t} \in \mathbb{R}$.
If node $v_i$ is activated at time $t-1$ (e.g., obeyed social distancing policies), then $Y_{i,t-1}=1$.
%If node $v_i$ is activated at time $t$, then $Y_{i,t} \leftarrow Y_{i,t} + \delta$ where $\delta \in \mathbb{R}$.
%If due to contagion effects $Y_{i,t} \rightarrow Y_{i,t}+\delta$, node $v_i$ is activated at time $t$.
%Node $v_i$ is activated at time $t-1$, if $Y_{i,t-1}=1$ and is activated at time $t$ if its outcome $Y_{i,t}$ has changed.
%We assume that $Y_{i,t-1}$ is unobserved. 
%Let $\mathbf{U}_i$ be the set of hidden confounders and $\mathbf{Z}_i$ be the set of observed attributes of node $v_i$. 
%The data can be on dyads i.e., pairs of two individuals, where $|N_i|=1$ or network where $|N_i|>1$.
%Let $A$ indicate the adjacency matrix where 
%for each $e_{i,j} \in E$, $A_{i,j}=1$ and 
%N_i={v_j|v_j\in V & \exists e_ij}
Let $\mathbf{N}_i=\{v_j|v_j\in \mathbf{V} \ \ \& \ \ \exists \ \ e_{ij} \in \mathbf{E}\}$  denote the set of neighbors of node $v_i$ and 
$\mathbf{A}_i$
%$=\{1\}^{|\mathbf{N}_i|}$
be the adjacency vector for node $v_i$ where $A_{ij}=1$ if  $\exists e_{ij}$.
%\fixme{$C_{i,ngb}=\{A_{i,j}: j\in{N_i}\}$.}
For each node, there exists a set of neighbors' hidden characteristics $\mathbf{U}_{ngb}$, a set of neighbors' observed attributes $\mathbf{Z}_{ngb}$, and two sets of neighbors' outcomes $\mathbf{Y}_{ngb,t-1}$ and $\mathbf{Y}_{ngb,t}$. 
We further assume that edges form by latent homophily on $\mathbf{U}$ which means that nodes with more similar unobserved characteristics are more likely to connect ~\citep{shalizi-arxiv16}. 
\subsection{Causal Model}
\commentout{
Structural Causal Models (SCMs) are graphical representations of cause-effect relationships between variables that allow reasoning about the identification of the effect of interest \citep{pearl-book09}. Causal Graphs are Directed Acyclic Graphs (DAGs) that represent SCMs, with variables as nodes and cause-effect relationships as edges.
 If variable Y is a child of variable X, we say that Y is caused by X.
 %, or that X is the direct cause of Y.
 }
Following \citet{egami-zrxiv2021}, we assume the causal graph depicted in Fig. \ref{fig:causalmodel}, where the connections are formed based on the similarity of the unobserved homophilic attributes. 
However, Egami and Tchetgen consider a chain graph \cite{lauritzen-jrs02} to represent network dependence of unobserved confounders across units. Hidden variables are represented by dashed circles. 

In each connection, "ego" refers to a node whose contagion effects we estimate, and "peer" refers to a node that influences the ego's outcome.
In this paper, treatment is the set of peer outcomes $\mathbf{Y}_{ngb,t-1}$ and the outcome is the ego's outcome $Y_{i,t}$. 
The potential outcome of node $v_i$ under contagion effects is defined as the value that $Y_{i,t}$ would take if peer's outcome $\mathbf{Y}_{ngb,t-1}$ had been set to $y$. The factual outcome $Y_{i,t}^F$ refers to the observed outcome of an individual when $\mathbf{Y}_{ngb,t-1}=y$ 
%under treatment $y$ 
%($Y_{i,t}^F=Y_{i,t}(y)$)
 and the counterfactual outcome $Y_{i,t}^{CF}$ shows the unobserved response of an individual when $\mathbf{Y}_{ngb,t-1}=1-y$.
 %\fixme{ $1-y$ ($Y_{i,t}^{CF}=Y_{i,t}(1-y)$)}.

Given a set of activated neighbors $\hat{\mathbf{N}}_i \subseteq \mathbf{N}_i$, we define $h: \{0,1\}^{|\mathbf{N}_i|}  \rightarrow \{0,1\}$
%$h(\mathbf{Y}_{ngb,t-1})$
as a function over the neighbors' activations which maps the neighbors' activations to a binary value.
%This function could be based on an aggregate of neighbors' activations which maps to a binary value.
%, or more sophisticated functions, such as a linear threshold model ~\citep{granovetter-ajs78}. 
In this paper,
we consider the ego-networks connection model where multiple activated peers may exist ($|\mathbf{N}_i|\geq1$). 
%, and function $h$ maps the vector of neighbors' activations to binary values.
Dyads, i.e., pairs of two individuals, are a special case of the ego-networks model where for every node $v_i$, $|\mathbf{N}_i|=1$ and an activated peer can activate an ego.

%two types of connection models: 1) dyads, i.e., pairs of two individuals, where for every node $v_i$, $|\mathbf{N}_i|=1$ and an activated peer can activate an ego,
%and 2) ego-networks where multiple activated peers may exist ($|\mathbf{N}_i|\geq1$) and function $h$ maps the vector of neighbors' activations to a real number. 
%This function makes our causal model distinct from the one presented in \cite{egami-zrxiv2021} where multiple neighbors can impact the outcome a node independently. 

\subsection{Contagion Effect Estimation}
%The fundamental problem of causal effect is that the 
%The potential outcome of i under treatment tr is defined as 
%ego — a unit on whom we estimate a causal peer effect — and define k = 2 to be the peer — a unit whose effect on the ego we estimate. 
\commentout{
\fixme{Most of the causal inference studies rely on Stable Unit Treatment Value Assumption (SUTVA), the assumption
that one unit’s outcomes are unaffected by another unit’s treatment assignment. However, this assumption is violated in the presence of interference where a unit's response can be affected by the treatment it receives and by the treatments received by its neighbors.} } 
%traditional causal inference methods rely on a critical assumption called the “stable unit treatment value assumption” (SUTVA)

%and $h$ selects one of the activated peers to activate the unactivated ego according to the activation probability vector $P$. %In both cases, there is only one 
%but this assumption is not necessary.
%Different activation functions (e.g., Linear Threshold Model \citep{}) can be employed

%activated neighbor has an activation probability $p$ which is defined as the probability of influencing 

   We define \textit{Individual Contagion Effects (ICE)} as the difference between the outcome of node $v_i$ under two different values for the neighbors' activation $h(\mathbf{Y}_{ngb,t-1})$:
%all peers are activated ($\mathbf{Y}_{ngb,t-1}=1$) and when they are not
%($\textbf{Y}_{ngb,t-1}=0$):
%peer effect exists ($T=1$) and when it does not ($T=0$). 
%The potential outcome of i under treatment T is defined as the value of $Y_{i,t}$ would have taken if the treatment of node i had been set to T. We define Individual Peer Effect (IPE) as the difference between the outcome of node i when peer effect exists and when it does not.
\begin{align}
%p_i= Y_{i,t}(Y_{j,t-1}=1)-Y_{i,t}(Y_{j,t-1}=0)
\tau_i= Y_{i,t}(h(\mathbf{Y}_{ngb,t-1})=1)-Y_{i,t}(h(\mathbf{Y}_{ngb,t-1})=0)
\end{align}

%vi.y(1) and vi.y(0) denote the potential outcomes of vi.y if unit vi were assigned to

Our objective is to estimate ACE, which represents the average of ICE over all nodes.
%and is measured as follows:
%the average of ITE over the instances as
\commentout{
\begin{align}
ACE(V)= \frac{1}{|V|}\sum_{i=1}^{|V|}{\tau_i}
%\E_{i\in V}[Y_i|T_i=t] 
%-\E_{i\in V}[Y_i|T_i=t].
\end{align}}
In observational data, estimating ICE is challenging because we can never simultaneously observe the factual and counterfactual outcomes of a unit. 
%A more common approach is to fit a model with factual outcomes and employ this model to predict counterfactual outcomes.

\begin{figure}

    \centering
        \includegraphics[width=0.3\textwidth]{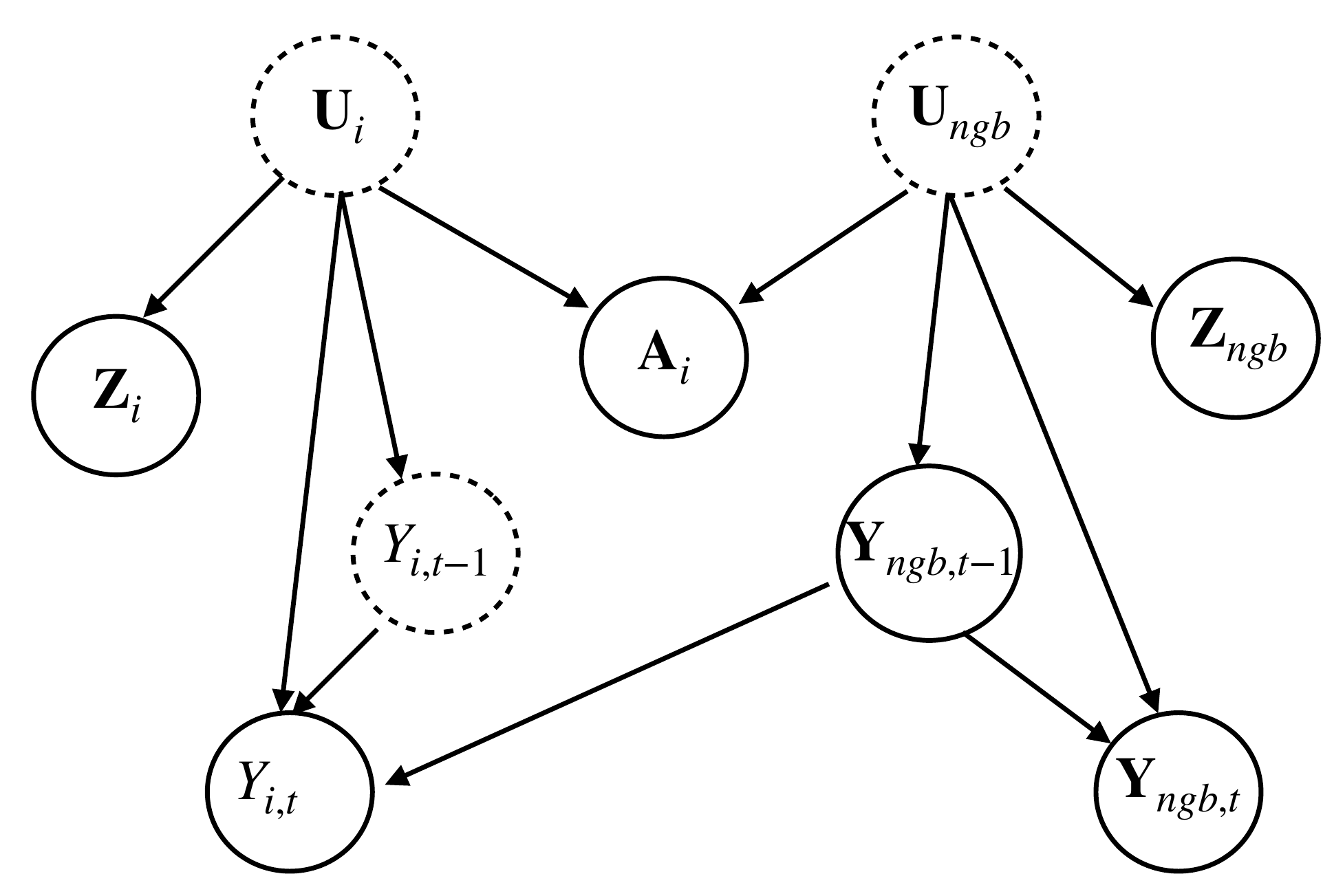}
        %\vspace{-10pt}
        \caption{The causal model for the ego-network of ego $v_i$: $\mathbf{Z}_i$ and $\mathbf{Z}_{ngb}$ are proxies of the hidden confounders. $\mathbf{U}_i$ and $\mathbf{U}_{ngb}$ are unobserved homophilic attributes. Dashed circles show unobserved variables.
        }
    \label{fig:causalmodel}
\end{figure}

%where C is the confounders causing treatment and outcome of nodes.
%One of the main approaches to estimating the causal effect of interest is adjusting for observed covariates in a linear reg
One of the main assumptions in inferring causal effects from observational data is strong ignorability or no unmeasured confounding. According to this condition, the potential outcomes of a node are independent of its treatment assignment given its observed attributes \citep{rosenbaum-biometrika83}.
%We formalize this assumption by using the standard strong ignorability condition
%Structural Causal Models (SCMs) are graphical representations of cause-effect relationships between variables that allow for the identification of the effect of interest \citep{pearl-book09}. 
 %identify backdoor paths and adjust for variables that block these paths between treatment and outcome of interest. 
Under the strong ignorability assumption, adjusting for variables that block the back-door path allows for the identification of the causal effect  \citep{pearl-book09}. A back-door path refers to a path from treatment to outcome in the structural causal model that has an arrow into the treatment variable. When unblocked back-door paths exist, there are two sources of association between treatment and outcome: one is causal (the causal effect of treatment on the outcome), and the other is non-causal (through the back-door path). In the presence of unblocked back-door paths, it becomes challenging to decide whether the observed association is a result of the causal effect or the spurious association enabled by the back-door path.
In the causal model represented in Fig. \ref{fig:causalmodel}, strong ignorability holds if:
%can be formalized as:
\begin{align}
    Y_{i,t}(1), Y_{i,t}(0) \indep \mathbf{Y}_{ngb,t-1} \mid \mathbf{Z}_i, \mathbf{A}_i.
\end{align}
 However, conditioning on $\mathbf{A}_i$ introduces a dependence association between unobserved variables $\mathbf{U}_i$ and $\mathbf{U}_{ngb}$ where the unblocked backdoor path $Y_{i,t} \leftarrow \mathbf{U}_i \rightarrow \mathbf{A}_i \leftarrow \mathbf{U}_{ngb} \rightarrow \mathbf{Y}_{ngb,t-1}$ violates the ignorability assumption ($Y_{i,t} \not\!\perp\!\!\!\perp \mathbf{Y}_{ngb,t-1}| \mathbf{A}_i, \mathbf{Z}_i$) and makes the contagion effects unidentifiable. We are interested in measuring ACE in the presence of an unobserved confounder, i.e., where the unobserved network confounder is the direct cause of the outcome of an ego and its peers. 

\subsection{Double Negative Control Proxies}
One way to account for latent homophily is by considering proxies for unobserved confounders.
Proxies are measurable variables that are correlated with the unobserved variable, and conditioning on them enables the identification of the causal effect \cite{miao-biometrika18}. 
%Proxies can be classified into three types  \citep{tchetgen-arxiv20}: a) variables which
%are common causes of the treatment and outcome variables, b) potential causes of
%the treatment which is related to the outcome only through an unobserved common cause for which the variable is a proxy, and c) potential causes of the outcome which
%are related to the treatment only through an unobserved common cause for which the variable is a proxy.
Negative controls are two groups of common proxies that make the causal effect identifiable in settings with unobserved confounders: 1) \textit{Negative Control Exposure} (NCE) is a treatment variable that does not causally affect the outcome of interest, and 2) \textit{Negative Control Outcome} (NCO) is a variable that is not causally affected by the treatment of interest. NCE and NCO correspond to proxy types b and c, respectively. 
% While there has been research on using negative controls to identify causal effects in i.i.d. data \citep{singharxiv20,kuroki-biomerika14,miao-arxiv18,tchetgen-arxiv20},
  \citep{egami-zrxiv2021} demonstrate that leveraging these two types of negative control proxies can enable the identification of contagion effects in networked data with unobserved confounders. In the causal model presented in Fig. \ref{fig:causalmodel}, a variable $\mathbf{Z}_i$ is considered as an NCO if:
\begin{align}
\label{eq:NCO}
  \mathbf{Z}_i \independent \mathbf{Y}_{ngb,t-1}| \mathbf{U}_i, \mathbf{U}_{ngb}, \mathbf{A}_i
\end{align}
and variable $\mathbf{Z}_{ngb}$ is considered as an NCE if:
\begin{align}
\label{eq:NCE1}
    &\mathbf{Z}_{ngb} \independent Y_{i,t} | \mathbf{Y}_{ngb,t-1}, \mathbf{U}_i, \mathbf{U}_{ngb}, \mathbf{A}_i, \\
     &\mathbf{Z}_{ngb} \independent \mathbf{Z}_i | \mathbf {Y}_{ngb,t-1}, \mathbf{U}_i, \mathbf{U}_{ngb}, \mathbf{A}_i.
\label{eq:NCE2}
\end{align}
\commentout{
\begin{align}

    Z_j \independent Y_{i,t} | Y_{j,t-1}, U_i, U_j, A_{ij} \enskip \text{and} \enskip
     Z_j \independent Z_i | Y_{j,t-1}, U_i, U_j, A_{ij}
     \label{eq:NCE}
\end{align}
}
%\citet{egami-zrxiv2021} show that those 
% variables which are U-relevant and satisfy assumptions \ref{eq:NCO}-\ref{eq:NCE}, can be considered as valid proxies of the unobserved confounder and make the contagion effects identifiable. 
%In the causal model presented in Figure \ref{fig:causalmodel}, assumptions \ref{eq:NCO}-\ref{eq:NCE} hold for variables $\mathbf{Z}_i$ as an NCE and $\mathbf{Z}_{ngb}$ as an NCO. It is worth mentioning that $Y_{i,t-1}$ can be considered as another NCO proxy, but we assume that $Y_{i,t-1}$ is unobserved.
%Assumption 1.2.1 states that W is an auxiliary variable that is conditionally independent of
%the treatment Y21 given the latent confounder U, observed pre-treatment covariates X, and the
%dyadic type S. Assumption 1.2.2 means that Z is an auxiliary variable that is conditionally
%independent of the outcome Y12 and NCO W given the treatment Y21, the latent confounder %U,
%observed pre-treatment covariates X.
Assumption \ref{eq:NCO} implies that the variable $\mathbf{Z}_i$ serves as an auxiliary variable which is independent of the treatment, given the unobserved confounders and the adjacency vector of the node. Assumptions \ref{eq:NCE1} and \ref{eq:NCE2} consider $\mathbf{Z}_{ngb}$ as an auxiliary variable, independent of the outcome and NCO, given treatment, unobserved confounders, and the adjacency vector. 
%Auxiliary variables are variables that can help to make estimates on incomplete data, while they are not part of the main analysis \citep{collins-pm01}.

%After identifying the negative control proxies, 
Various estimators can be employed to infer the causal effect of interest using proxies. One commonly used approach is the \textit{Two-stage Least Squares estimator (TSLS)}. 
%TSLS is a method commonly applied in linear models with instrumental variables, which are variables correlated with the predictor variable but uncorrelated with the outcome variable \citep{angrist-asa95}.
TSLS  consists of two stages \citep{angrist-asa95}. First, a new variable is constructed using the instrumental variables. This variable serves as a proxy for the unobserved confounders. Then,  the estimated values obtained from the first stage are utilized in place of the unobserved confounders, and an \textit{Ordinary Least Squares regression (OLS)} is performed to estimate the causal effect. 
%OLS is a technique used for estimating the coefficients in linear regression models.
~\citep{egami-zrxiv2021} employ the TSLS estimator to quantify contagion effects by leveraging the NCE and NCO proxies as: 
%Specifically, the NCE and NCO proxies are utilized as instrumental variables in the TSLS framework to estimate the causal effects.
%There are two distinct stages in TSLS: 1) 
%They employ Two-stage Least Squares estimator to measure the causal effect of interest. 
%Two-stage least squares (TSLS) is a special case of instrumental variables regression. As the name suggests, there are two distinct stages in two-stage least squares. In the first stage, TSLS finds the portions of the endogenous and exogenous variables that can be attributed to the instruments. This stage involves estimating an OLS regression of each variable in the model on the set of instruments. The second stage is a regression of the original equation, with all of the variables replaced by the fitted values from the first-stage regressions. The coefficients of this regression are the TSLS estimates.
%In the first stage, a new variable is created using the instrument variable.
%In the second stage, the model-estimated values from stage one are then used in place of the actual values of the problematic predictors to compute an OLS model for the response of interest.
\begin{equation}
    Y_{i,t} \sim \mathbf{Y}_{ngb,t-1}+\mathbf{Z}_i|\mathbf{Z}_{ngb}+\mathbf{Y}_{ngb,t-1}
    \label{eq:tsls}
\end{equation}
where the coefficient of $\mathbf{Y}_{ngb,t-1}$ shows the estimated ACE.
%where $Z_{ngb}$ indicates the matrix of neighbors' attributes for all nodes, $Y_{ngb,t-1}$ denotes the matrix of neighbors' outcome at time $t-1$ and the coefficient of $Y_{ngb,t-1}$ shows the estimated peer contagion effect.
%Two-stage least squares are used to compute the estimand.

%\fixme{advantage of T-learner}

\subsection{Issues with high-dimensional proxies}
%Dimensionality refers to the number of attributes or features present in a dataset. 
High-dimensional datasets are characterized by a large number of attributes, such as customers' extensive purchase history in recommender systems.
%, customers' extensive purchase history in recommender systems, or health data of patients. 
In the presence of high-dimensional data, the number of model parameters $p$ exceeds the number of data samples $n$, a problem known as the “Large $p$ Small $n$” issue in causal effect estimation using regression models \citep{bernardo-bs03}. Estimating contagion effects using control proxies can be problematic when the NCO and NCE proxies are high-dimensional because the matrix of model parameters becomes sparse and exhibits a low-rank structure \citep{deaner-arxiv21}. A matrix is considered low-rank if the number of linearly independent variables is significantly smaller than the total number of variables. Including correlated variables in the estimation process increases the variance of the causal estimand \citep{abadie-eco06,de-oup11}, which adversely affects the performance of the estimator \citep{chao-eco05,hansen-bes08}.
This issue becomes even more prominent in TSLS estimation, where the computational burden increases with the number of instruments or predictors. 
%Consequently, scalability issues can arise, making the TSLS approach computationally challenging when dealing with a large number of instruments or predictors.
%This situation can be even worse in TSLS where by increasing the number of instruments or predictors, the computational burden of TSLS can become significant, leading to scalability issues.

%As the number of instruments or predictors increases, the computational burden of TSLS can become significant, leading to scalability issues.
Besides high dimensionality, selection bias can be another source of high variance for the causal effect estimate. This bias arises due to a mismatch in the distribution of attributes between the treatment and control groups. In the presence of selection bias, the observed differences between the outcome of treatment and control nodes may be due to the difference between their attributes rather than the true treatment effect.

%In the presence of high-dimensional proxies, the matrix of model parameters is sparse and has a low-rank structure \citep{deaner-arxiv21}. A matrix is low-rank if the number of linearly independent variables is significantly lower than the total number of variables. Including correlated variables in the estimation increases the variance of the causal estimand \citep{abadie-eco06,de-oup11} and damages the performance of the estimator \citep{chao-eco05,hansen-bes08}.
%Previous studies show that the TSLS estimator is consistent only when the number of instruments increases slowly with the sample size \citep{chao-eco05,hansen-bes08}.

The goal of this paper is to solve the following problem:

\begin{problem}[Contagion Effect Estimation with High-dimensional Proxies]
\label{problem-def}
Let $G = (\bm{V},\bm{E})$ be a graph evolved by latent homophily with high-dimensional double negative control proxies, associated with nodes.
%and let $\hat{\theta}$ be an estimate of ACE.
Our goal is to find an estimate of the average contagion effect (ACE) $\hat{\theta}$ that minimizes the expected error between $\hat{\theta}$ and the true value of ACE $\theta$.

%minimize the expected error between the estimate ($\hat{\theta}$) and the true ($\theta$) values of average contagion effects (ACE) is minimized.
%estimate Average Peer Effects ($\hat{APE}$) such that the estimation error is minimized:
\commentout{
\begin{equation}
\label{objective}
\begin{aligned}
%& \underset{}{\text{maximize}} 
& \underset{}{\mathrm{argmin}}
%&& \sum_{i=1}^{g} \sum_{j=i+1}^{g} (a_{ij} \times \sum_{k=1}^{|c_i|}\sum_{l=1}^{|c_j|} r^{ij}_{kl}) \\
& E|\hat{\theta}-\theta|.
\end{aligned}
\end{equation}}
\end{problem}

\section{Proximal Embedding Framework for Contagion Effect Estimation}
\label{solution}

\begin{figure*}[t]
    \centering
        \includegraphics[height=0.34\textwidth]{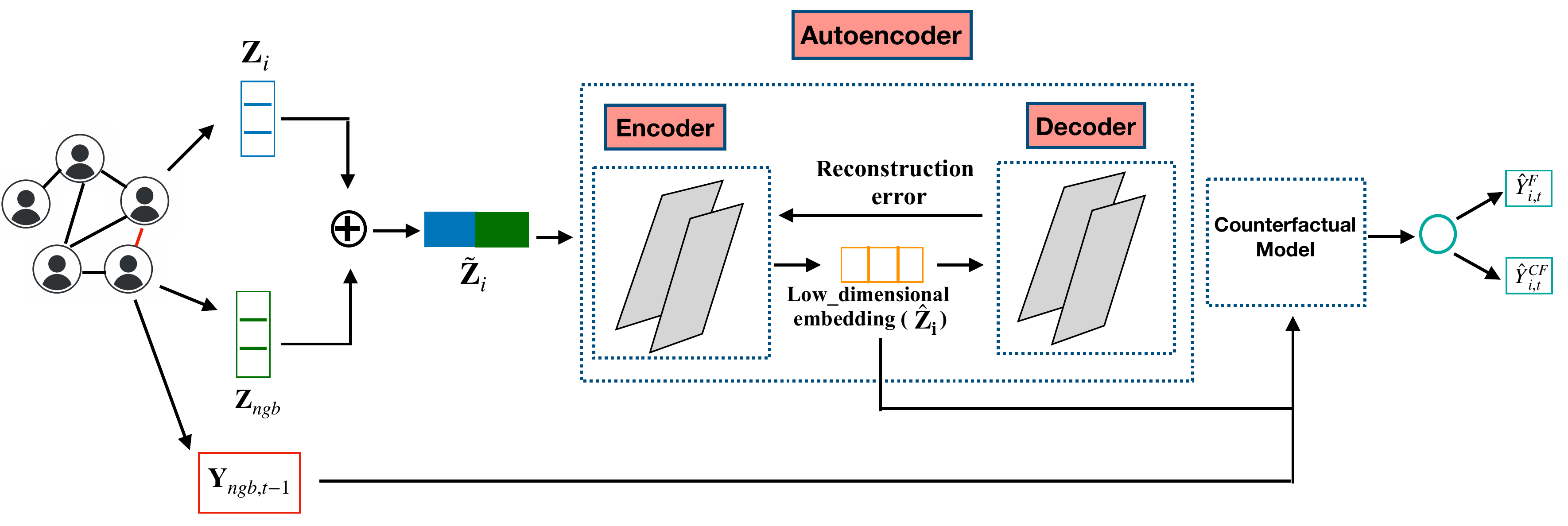}
        \caption{Illustration of the ProEmb framework with three components: 1) embedding learning where VAEs are used to generate low-dimensional embeddings of the proxies, 2) representation balancing where adversarial networks are used to improve the proxy representation mismatch between treatment groups,  and 3) counterfactual model where an estimator is trained to predict the counterfactual outcomes.
        }
    \label{fig:AE}
\end{figure*}

To address high dimensionality and selection bias in contagion effect estimation, % from network data with latent homophily, 
we introduce the \textit{\textbf{Pro}ximal \textbf{Emb}eddings (ProEmb)} framework. ProEmb has three main components, shown in Fig. \ref{fig:AE}. 
\commentout{The first component deals with sparsity and high dimensionality by reducing the number of the initial dependent variables to a smaller number of uncorrelated variables \citep{wang-cvpr14}. One prominent method for dimensionality reduction is Variational Autoencoders (VAEs) \cite{kingma-stat14,rezende-icml14}, a type of autoencoder that represents inputs as probability distributions in the latent space and enforces regularization by ensuring that the distributions generated by the encoder closely resemble a standard Gaussian distribution. 
Such methods combine highly correlated variables into a set of uncorrelated variables to reduce the variance and thereby enhance the optimality of the estimators \citep{de-oup11}.}
The first component tackles issues of sparsity and high dimensionality by reducing dependent variables to uncorrelated ones, thereby improving estimator optimality \citep{wang-cvpr14,de-oup11}. A key technique for this is variational autoencoders (VAEs) \cite{kingma-stat14, rezende-icml14}, which represent inputs as distributions in a latent space and enforce regularization by ensuring encoder-generated distributions closely resemble a Gaussian distribution. 
%These approaches combine correlated variables into uncorrelated sets, reducing the variance and enhancing the estimator optimality \citep{de-oup11}.

 However, simply applying VAEs to causal inference is not sufficient because embeddings generated by VAEs can vary across different treatment groups and it can lead to confounding biases in estimating causal effects. To address this problem, the second component of ProEmb integrates adversarial networks to update the generated representation by VAEs
 and improve the distribution shift between the representation of treatment and control node proxies. This updated representation is then passed on to the third component which consists of a counterfactual learning module that measures counterfactual outcomes using meta-learners \cite{kunzel-nas19}.
To the best of our knowledge, ProEmb is the first method that integrates VAEs, adversarial networks, and meta-learners to improve causal effect estimation more generally, and more specifically contagion effect estimation in networks with unobserved confounders. Next, we describe each component in more detail.

 %We propose a novel framework to improves contagion effect estimation accuracy in settings with unobserved confounders and high-dimensional proxy variables.
 \commentout{
 A common approach for dealing with sparsity and high-dimensionality is reducing the number of the initial dependent variables to a smaller number of uncorrelated variables through dimensionality reduction techniques \citep{wang-cvpr14}. 
 Such methods combine highly correlated variables into a set of uncorrelated variables to reduce the variance and thereby enhance the optimality of the estimators \citep{de-oup11}. Our framework leverages such techniques to reduce the dimensionality of proxy variables and increase the contagion effect estimation accuracy.}
 
 %that is, data that contain high-dimensional proxies. 
% \subsection{Proximal Embedding Framework}
\commentout{
 We introduce \textit{\textbf{Pro}ximal \textbf{Emb}eddings (ProEmb)}, a framework for estimating contagion effects in the presence of latent homophily and high-dimensional proxies. We assume that the experimenter has classified observed variables into NCO and NCE proxies based on assumptions \ref{eq:NCO}-\ref{eq:NCE2}.
  Our framework represented in Fig. \ref{fig:AE} consists of three main components: 1) embedding learning which reduces the dimensionality of proxy variables, 2) representation balancing which improves the distribution shift between the representation of  treatment and control nodes proxies, and 3) counterfactual learning which measures counterfactual outcomes.}

 \subsection{Embedding learning} The goal of this component is learning a low-dimensional representation of high-dimensional and sparse proxies while preserving the parts of proxies that are predictive of the outcomes. 
 We assume that the experimenter has classified observed variables into NCO and NCE proxies based on assumptions \ref{eq:NCO}-\ref{eq:NCE2}.
 We use VAEs to learn low-dimensional representations for each node's proxies.
 %, utilizing the attributes of the node and its neighboring nodes. 
 VAEs have demonstrated remarkable success in dimensionality reduction due to their ability to both capture the underlying structure of high-dimensional data and regularize the latent space, which helps to prevent overfitting and improve generalization performance ~\cite{gregor-icml15,jimenez-neurips16,pu-neurips16,portillo-aj20}. 
 
In order to adapt VAEs to the problem of contagion estimation with high-dimensional proxies, one has to be careful to consider 1) how to capture latent homophily, 2) how to sample diverse low-dimensional representations from the representation space during training and inference, and 3) how to reconstruct the original high-dimensional proxy vectors, in order to evaluate and improve the performance of the model. ProEmb's variational autoencoder addresses these considerations through each of its three parts:
 \begin{enumerate}
     \item \textit{Probabilistic Encoder}.
     %With the aim of modeling p(z|x), 
     This component transforms high-dimensional proxies into a distribution in the latent space to infer the unobserved confounders. Since $\mathbf{Z}_i$ as an NCO and $\mathbf{Z}_{ngb}$ as an NCE variable are proxies of the unobserved homophilic attributes, we expect to recover latent features by applying a well-trained encoder model to the concatenation of these proxies. Let $\mathbf{\tilde{Z}}_{i}=\{ z_{i,1},...,z_{i,n},z_{ngb,1},\allowbreak...,z_{ngb, n} \} $ denote the concatenated vector of proxies $\mathbf{Z}_i=\{ z_{i,1},...,z_{i,n}\}$ and $\mathbf{Z}_{ngb}=\{z_{ngb,1},...,z_{ngb, n}\}$ with dimension $n$. We use the encoder layer with $L$ fully-connected
layers to map proxies $\mathbf{\tilde{Z}}_{i}$ to low-dimensional latent vector $\mathbf{Z'}_{i}$ as: %$\mathbf{\hat{Z}}_{i}=\{\hat{z}_1,...,\hat{z}_m\}$ 
     %with dimension $m$ where $m < n$. 
     %The encoder uses $L$ fully-connected
%layers to map $\mathbf{\tilde{Z}}_{i}$ to $\mathbf{Z'_{i}}$, i.e.,
 \begin{align}
   \mathbf{Z'}_{i} = g(\mathbf{W}_l ... g(\mathbf{W}_1\mathbf{\tilde{Z}}_i)),
 \end{align}
where $g$ indicates the activation function (e.g., Relu) and $\{\mathbf{W}_l\}, l\in \{1,..., L\}$ represents the weight matrices of the fully connected layers of the encoder.

\item \textit{Sampler}. 
 The sampler plays a crucial role in generating latent vectors from the learned distribution in the latent space. These vectors are randomly sampled from the distribution $p(\mathbf{\hat{Z}}_i| \mathbf{\tilde{Z}}_i)$, utilizing the mean and log-variance values obtained from the encoder's output.
The latent layer is represented by two sets of neurons: one set representing the means of the latent space, and the other set representing the log-variances, 
%referred to as the latent means and latent log-variances and 
measured as:
\begin{align}
    \mu &= \mathbf{W_{\mu}\mathbf{Z'}_i} + \mathbf{b^{\mu}}, \\
    ln \, \delta^2 &= \mathbf{W_{\delta}\mathbf{Z'}_i} + \mathbf{b^{\delta}},
\end{align}
 where $\mathbf{b^{\mu}}$ and $\mathbf{b^{\delta}}$ are vectors of biases. A proxy representation is sampled from the latent space as:  
\begin{align}
    \mathbf{\hat{Z}}_i \sim p(\mathbf{\hat{Z}}_{i}|\mathbf{\tilde{Z}}_i) = \mathcal{N}(\mu,exp(ln \ \ \delta^2)).
\end{align}
$\mathbf{\hat{Z}}_{i}$ contains the low-dimensional representation of the proxies which is later used by the counterfactual learning component for contagion effect estimation.
     
 \item \textit{Probabilistic Decoder}.
 %The decoder samples a point in latent space from the Gaussian distribution and then decodes it as the reconstruction.
 The decoder attempts to reconstruct the original proxy vector $\mathbf{\tilde{Z}}_{i}$ from the proxy representation $\mathbf{\hat{Z}_{i}}$. The decoder uses $\hat{L}$ fully-connected
layers to map $\mathbf{\hat{Z}_{i}}$ to $\mathbf{\tilde{Z}_i}$, i.e.,
     \begin{align}
   \mathbf{Z}_i^{\prime\prime} = f(\mathbf{\hat{W}}_l ... f(\mathbf{\hat{W}}_1\mathbf{\hat{Z}}_i)),
 \end{align}
    where $\mathbf{Z}_i^{\prime\prime}$ shows the reconstructed representation, $f$ indicates the activation function, and $\{\mathbf{\hat{W}}_l\}, l\in {1,...,\hat{L}}$ denotes the weight matrix of the fully connected layers.
 \end{enumerate}
 
 %The objective of the VAEs component is to capture the relevant information from proxies while reducing the dimensionality. 
 The loss function of VAEs consists of two main parts: 1) the reconstruction loss  which measures the dissimilarity between the original data and the data reconstructed by the VAEs, and
 2) the Kullback–Leibler (KL) divergence \cite{kullback-ams51} which is a regularizer and  
 quantifies the discrepancy between the inferred distribution $p(\mathbf{\hat{Z}}|\mathbf{\tilde{Z}})$ and the known distribution $p(\mathbf{\tilde{Z}})$.
 %The KL divergence encourages the latent space to adhere to a prior distribution, typically a standard Gaussian distribution. 
 The loss function of VAEs is defined as:
 \begin{equation}
 %\begin{aligned}
   \mathcal{L}_{vae} = \frac{1}{|\mathbf{V}|} \sum_{i=1}^{|\mathbf{V}|}{|{z^\prime_i}-z_i|^2} + KL(p(\mathbf{\hat{Z}}_i|\mathbf{\tilde{Z}}_i)|p(\mathbf{\tilde{Z}}_i))_{i=1}^{|\mathbf{V}|}.
   % \end{aligned}
\end{equation}

\subsection{Representation balancing} 
%\fixme{hasbeenusedtoachieve balancebetweentreatmentgroupdistributions,seeking representationsthatarebothpredictiveofpotentialoutcomes,andbalancedacrosstreatmentgroups
%is optimized by maximizing the probability of distinguishing between real and generated data
%}
Since the embedding learning models are trained on the factual outcomes and used to predict the counterfactual outcomes, minimizing the error in factual outcomes $Y^F_{i,t}$ does not guarantee the simultaneous error reduction in counterfactual outcomes $Y^{CF}_{i,t}$. Therefore, it is crucial to develop a model that tackles the distribution mismatch between treatment and control nodes in terms of their proxy representations.
In this particular component, our focus is on enhancing proxy representation to achieve similarity between the induced distributions for treated and control nodes.
 Inspired by \cite{jiang-cikm22}, we employ the discriminator component of Generative Adversarial Networks \cite{goodfellow-neurips14} to address the imbalance proxy representations generated by VAEs.
 
 Let $\mathcal{D}:  \mathbf{\hat{Z}}_i  \rightarrow \{0,1 \}$ denote the discriminator function, mapping the latent representation $\mathbf{\hat{Z}}_i$ to $Y_{i,t-1}$.
 Initially, we train the discriminator to maximize the probability of accurately predicting $Y_{i,t-1}$ from the latent representation. This is achieved by optimizing the  discriminator loss function: 
\begin{align}
\mathcal{L}_{\mathcal{D}}&= \frac{1}{|\mathbf{V}|}\sum_{i=1}^{|\mathbf{V}|}(Y_{i,t-1} \  log \ \mathcal{D}(\hat{\mathbf{Z}}_i)\\ \nonumber
    &+(1-Y_{i,t-1}) \ log(1-\mathcal{D}(\hat{\mathbf{Z}}_i)).
    %\vspace{-10pt}
\end{align}
\commentout{
\begin{align}
\mathcal{L}_{\mathcal{D}}= \frac{1}{|\mathbf{V}|}\sum_{i=1}^{|\mathbf{V}|}(Y_{i,t-1} \  log \ \mathcal{D}(\hat{\mathbf{Z} }_i)
    +(1-Y_{i,t-1}) \ log(1-\mathcal{D}(\hat{\mathbf{Z} }_i)).
\end{align}}
 We update the latent representation $\mathbf{\hat{Z}}_i$ such that the distribution $p(Y_{i,t-1}|\mathbf{\hat{Z}}_i)$ becomes uniform. Considering that $Y_{ngb,t-1}$ is binary, a uniform distribution implies that $p(Y_{ngb,t-1}=1|\mathbf{\hat{Z}}_i) = p(Y_{ngb,t-1}=0|\mathbf{\hat{Z}}_i) = 0.5$. %To achieve this regularization, we define the following loss term:
The $p(Y_{ngb,t-1}|\mathbf{\hat{Z}}_i)$ regularization loss is defined as:
\begin{align}
     \mathcal{L}_{rb}= \frac{1}{|V|}\sum_{i=1}^{|V|}{(\mathcal{D}(\mathbf{\hat{Z}}_i)-0.5)^2}.
 \end{align} 
The regularization loss $\mathcal{L}_{rb}$ is then backpropagated to the encoding part of the VAEs, enabling the update of the latent representation $\mathbf{\hat{Z}}_i$ such that the discriminator $\mathcal{D}$ cannot accurately predict $Y_{ngb,t-1}$. This leads to a more balanced and unbiased latent representation for proxies.

\subsection{Counterfactual learning}
This component focuses on training a model to infer the counterfactual outcomes from low-dimensional embeddings of proxies  $\mathbf{\hat{Z}_{i}} \in R^m$ as well as the peers' outcome $\mathbf{Y}_{ngb, t-1}$. The factual outcomes are used to train the model.
The objective function of this component during training is to minimize the error of the inferred factual outcomes defined as $\frac{1}{n}\sum_{i=1}^n{(\hat{Y}_{i,t}- Y_{i,t})}^2$ where $\hat{Y}_{i,t}$ indicates the predicted factual outcome by ProEmb. 
 %A causal meta-learner uses the outputs of base learners (e.g., Linear Regression) for estimating causal effects ~\citep{kunzel-nas19}.
 %We utilize T-Learner to estimate counterfactual outcomes.
 %it uses so-called base learners to estimate the conditional expectations of the outcomes separately for units under control and those under treatment.

%A vast majority of studies uses Heterogeneous Treatment Effect (i.e., the effect of treatment of different individuals) estimators to measure counterfactuals \citep{hill-cgs11,athey-sta15,johansson-icml16}. T-learner meta-learning algorithm is an example of such estimators and is used to measure Conditional Average Treatment Effect (CATE). 
%Meta-learners are examples of such estimators.    
 %\fixme{Add references to different heterogeneous treatment effect estimators to measure counterfactuals}
%Our framework can leverage advanced methods for estimating heterogeneous treatment effects that rely on machine learning techniques \citep{hill-cgs11, athey-sta15, johansson-icml16}. These methods are designed to estimate treatment effects in scenarios with heterogeneous treatment responses. 
We assume that the proxy representation is sufficiently informative to satisfy the strong ignorability assumption ($Y_{i,t} \!\perp\!\!\!\perp \mathbf{Y}_{ngb,t-1}| \mathbf{\hat{Z}_i}$), which allows us to make counterfactual predictions. To make this process more concrete, we demonstrate how our framework would use a common Heterogeneous Treatment Effect (HTE) estimation algorithm, the T-learner. However, our framework could leverage other HTE estimation algorithms as well.
 T-learner meta-learning algorithm is an example of such estimators and is used to measure Conditional Average Treatment Effect (CATE).
 A meta-learner is a framework to estimate the Individual Treatment Effects (ITE) using any supervised machine learning estimators known as base-learners ~\citep{kunzel-nas19}.
 %A causal meta-learner uses the outputs of base learners (e.g., Linear Regression) for estimating causal effects ~\citep{kunzel-nas19}.
 %We utilize T-Learner to estimate counterfactual outcomes.
 %it uses so-called base learners to estimate the conditional expectations of the outcomes separately for units under control and those under treatment.
 In T-learner, two  base-learners are trained with treatment ($\mu_t$) and control nodes ($\mu_c$) to estimate the conditional expectations of the outcomes given observed attributes.
 \commentout{
\begin{align}
    \mu_t(\mathbf{z})=\mathop{\mathbb{E}}[Y_t(y=1)|\mathbf{Z}=\mathbf{z}],\\
    \mu_c(\mathbf{z})=\mathop{\mathbb{E}}[Y_t(y=0)|\mathbf{Z}=\mathbf{z}].
\end{align}
}
 $\mu_t$ is employed to
predict the counterfactual outcomes of control nodes, and 
$\mu_c$ is used to predict the counterfactual outcomes
of treatment nodes. The difference between the predicted outcomes by treatment and control models show ITE. 
\commentout{
is measured as:
 \begin{align}
    \hat{\tau}(\mathbf{z})= \hat{\mu}_t(\mathbf{z}) - \hat{\mu}_c(\mathbf{z})
 \end{align}
 where $\hat{\mu}_t(\mathbf{z})$ and $\hat{\mu}_c(\mathbf{z})$ are the predictions. 
}
 A T-learner has the advantage of simplicity and adaptability, as it utilizes any machine learning model as a base-learner.

\section{Experiments}
\label{experiments}
In this section, we evaluate the performance of different methods for contagion effect estimation on semi-synthetic datasets. We also demonstrate the applicability of our approach for detecting contagion effects in a Twitter dataset about the 2017 French presidential election \cite{burghardt-icwsm23} and a Peer Smoking dataset \cite{mermelstein-ntr09}.
\subsection{Semi-synthetic data generation}
As ground truth for causal effects is unavailable in real-world datasets, researchers commonly rely on synthetic and semi-synthetic datasets to evaluate causal inference methods. In this section, we describe the semi-synthetic datasets we generated for our experiments. It is important to note that the generation doesn't consider embeddings and is therefore not biased towards an embedding-based solution. 
We utilize four real-world datasets: 1) \textit{Hateful Users}, which is a sample of 5,000 hateful and normal tweets \citep{ribeiro-icwsm18}, 2) \textit{Stay-at-Home (SAH)}, which is a sample of 30,000 tweets reflecting users' attitudes toward stay-at-home orders during the COVID-19 pandemic  \citep{fatemi-dsaa22}, 3) \textit{BlogCatalog}, which is a sample of 5,196 bloggers from an online community where users
post blogs \cite{guo-wsdm20}, and 4) \textit{Flickr} which is a sample of 7,575 users who share photos on Flickr online social media platform \cite{guo-wsdm20}.
%We initially preprocess.
%the data by removing URLs from tweet texts, converting all words to lowercase, and eliminating punctuation and stopwords. We also perform lemmatization and stemming on the words. 
In the first two datasets, each tweet exhibits a unique distribution over several topics, reflecting the hidden semantic structure of the tweet. We consider the topic distribution of each tweet as the unobserved confounder $\mathbf{U}_i$.
To extract the topic distribution of each tweet, we employ Latent Dirichlet Allocation (LDA) \citep{blei-jmlr} and measure the coherence score
%. We utilize the elbow curve, which plots the number of topics against the coherence score,
to determine the optimal number of topics. We obtain 20 topics for SAH and 50 topics for the Hateful Users dataset. For the BlogCatalog and Flickr datasets, we follow \cite{guo-wsdm20} and learn 50 topics.
%The coherence score evaluates the interpretability of topics by humans. 
More dataset details are in the Appendix.

\textbf{Ego-networks model}.
Since our causal model relies on the assumption that ties form between nodes by latent homophily, we generate the networks synthetically.
We consider data on both networks and dyads and generate ties between nodes based on latent homophily. The advantage of considering both dyadic and network data is that it allows us to examine scenarios where a node is influenced by either a single activated neighbor or multiple activated neighbors. By considering dyadic data, we can focus on the interactions between pairs of nodes and gain insights into how one node's activation affects its immediate neighbor. This analysis provides valuable information about the dynamics at the micro-level. On the other hand, analyzing network data allows us to capture the broader influence of multiple activated neighbors on a node. 

 In the dyadic model, each node in the graph is connected to only one other node. The probability of an edge forming between node $v_i$ and $v_j$ is determined by the cosine similarity of their latent attribute vectors $\mathbf{U}_i$ and $\mathbf{U}_j$. 
This means that individuals with similar latent attributes are more likely to be connected. 

In the network model, we aim to generate networks growing based on latent homophily and preferential attachment. We start with $m_0=3$ fully connected seed nodes. At each time step, a new node $v_j$ connects to $m=3$ existing nodes, selected randomly with a probability proportional to the node's degree (~\citep{piva-jcn21}):
\begin{align}
\pi(k_i|j)=\frac{\cos(\mathbf{U}_i,\mathbf{U}_j)k_i}{\sum_n \cos(\mathbf{U}_i,\mathbf{U}_n)k_n}.
\end{align}
%We consider data on both networks and dyads and generate ties between nodes based on homophily. The connection model and experimental results for dyadic data are described in the appendix.
%In the dyadic model, each node in the graph can only be connected to one other node. The probability of an edge forming between node $v_i$ and $v_j$ is determined by the cosine similarity of their latent attribute vectors $\mathbf{U}_i$ and $\mathbf{U}_j$.
%This means that individuals with similar latent attributes are more likely to be connected. 
%We follow ~\citep{piva-jcn21} to generate networks growing based on homophily and preferential attachment (included in Supplement).
%Each alter can influence the ego's outcome with the probability of $0.3$ if the ego has not be in
%Each unaffected ego gets the influence from alters with the probability of $0.3$.
In our network model, we assume that activated neighbors may activate an inactivated ego with the probability of $0.3$. In the Supplement, we also show results for a different model that focuses on dyads and networks with mean() for h.
%and only one peer can activate the ego.
%. The peer who activates the ego is considered the only peer who affects the ego.

%This means that each ego can be activated by just one neighbor and 

\textbf{Counterfactual model}.
We generate the outcome of each node in two consecutive time steps. $Y_{i,t-1}$ is generated as:
\begin{align}
    Y_{i,t-1} = Bernoulli(sigmoid(\alpha_u \textbf{U}_i  +\epsilon))
\end{align}
where $\epsilon \sim \mathcal{N}(0,\,1)$, and $\alpha_u$ is the unobserved confounder coefficient vector with the size of $\mathbf{U}_i$. We generate the factual and counterfactual outcomes of each node at time $t$ as:
\begin{align}
\label{eq:foutcome1}
    &Y_{i,t}^F=\beta_u \textbf{U}_i + \beta_y Y_{i,t-1}+ \tau h(\mathbf{Y}_{ngb,t-1}) +\epsilon
   % \label{eq:fcoutcome1}
\end{align}
\begin{equation}
\label{eq:fcoutcome2}
    Y_{i,t}^{CF}=\beta_u \mathbf{U}_i + \beta_y Y_{i,t-1}+ \tau (1-h(\mathbf{Y}_{ngb,t-1})) +\epsilon
\end{equation}
where $\beta_u$ is the unobserved confounder coefficient vector. 
%We use max() and mean() for h() in the experiments. 
In our experiments, we utilize both the max() and mean() functions for h(). Specifically, when employing the mean() function to measure $\mathbf{Y}_{ngb,t-1}$, we transform the score to 1 if it surpasses 0.5, and if it falls below this threshold, we assign it a value of 0.

%\fixme{Among multiple peers that can activate an ego, one of them is selected to activate the node at the end. This contrasts with the assumption made in \cite{egami-zrxiv2021} where the function h represents a mean aggregation function, leading to a distinct causal model compared to the one considered in this paper.}

%$\epsilon \sim \mathcal{N}(\mu,\,\sigma^{2})$.

\commentout{
\begin{figure*}[ht]
\begin{subfigure}{.24\textwidth}
  \centering
  % include second image
  \includegraphics[width=\textwidth]{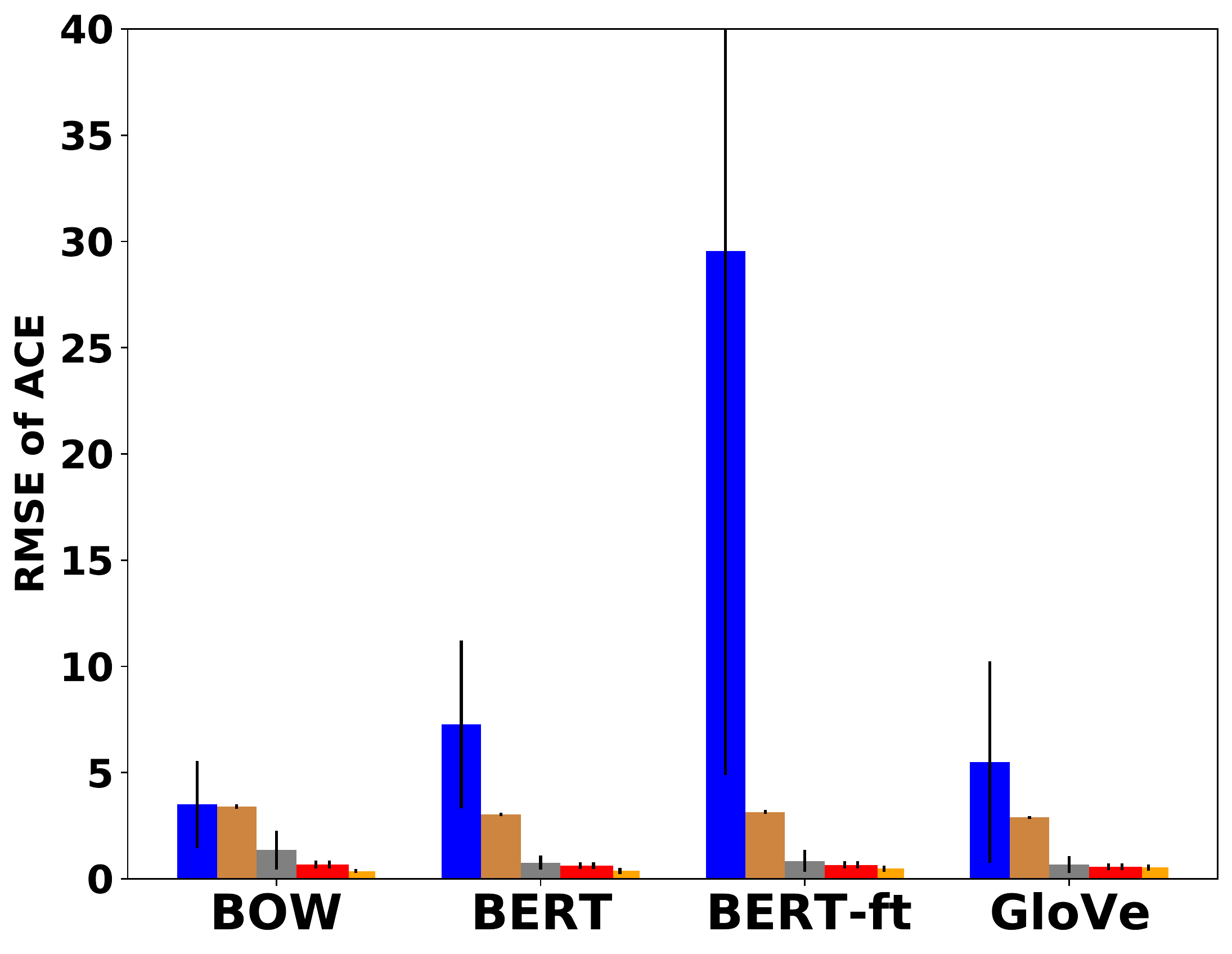}  
  \caption{SAH, $\beta_u \sim \mathcal{N}(0,3)$}
  \label{fig:sub-second}
\end{subfigure}
\begin{subfigure}{.24\textwidth}
  \centering
  % include second image
  \includegraphics[width=\textwidth]{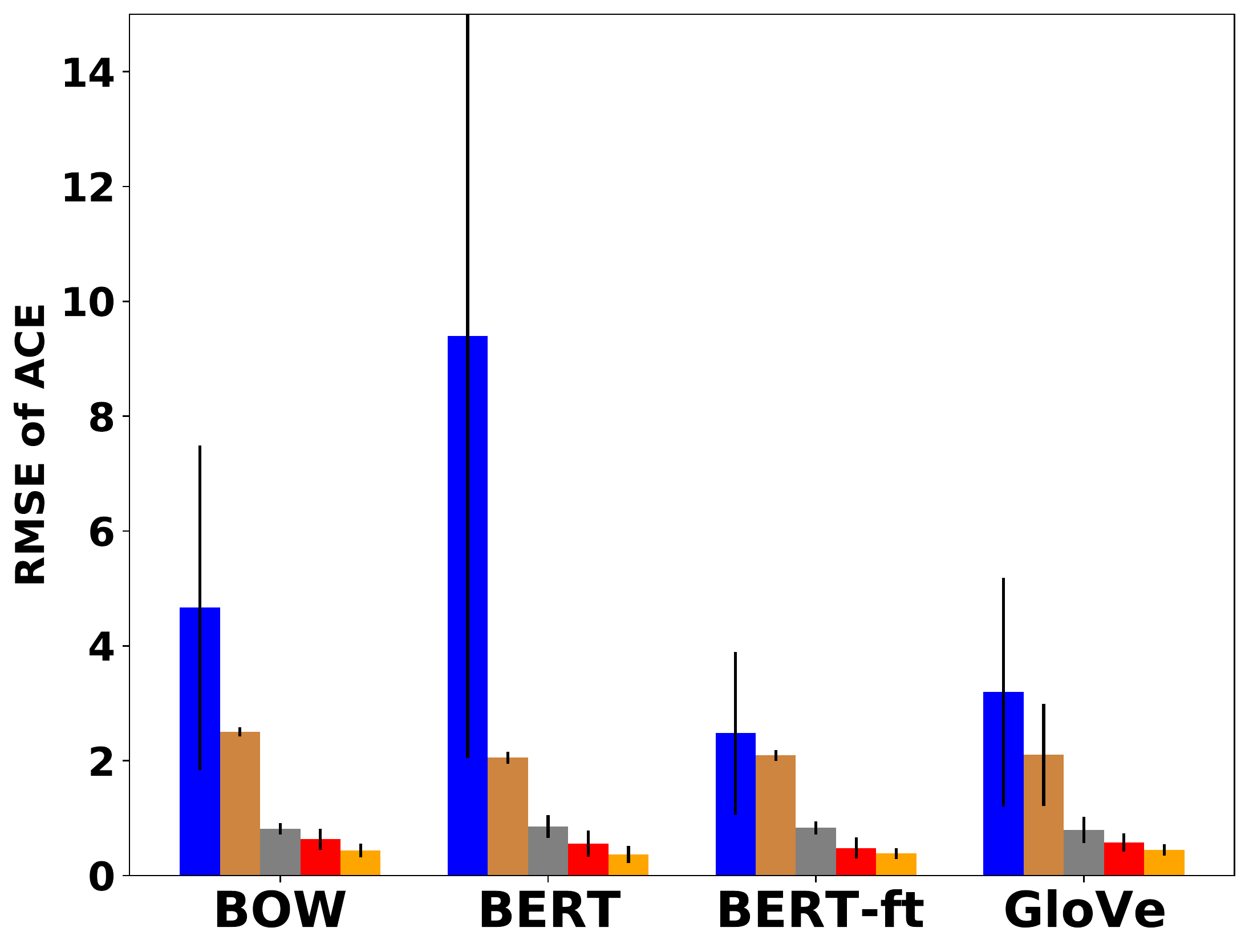}  
  \caption{SAH, $\beta_u \sim \mathcal{N}(5,2)$}
  \label{fig:sub-second}
\end{subfigure}
\begin{subfigure}{.24\textwidth}
  \centering
  % include second image
  \includegraphics[width=\textwidth]{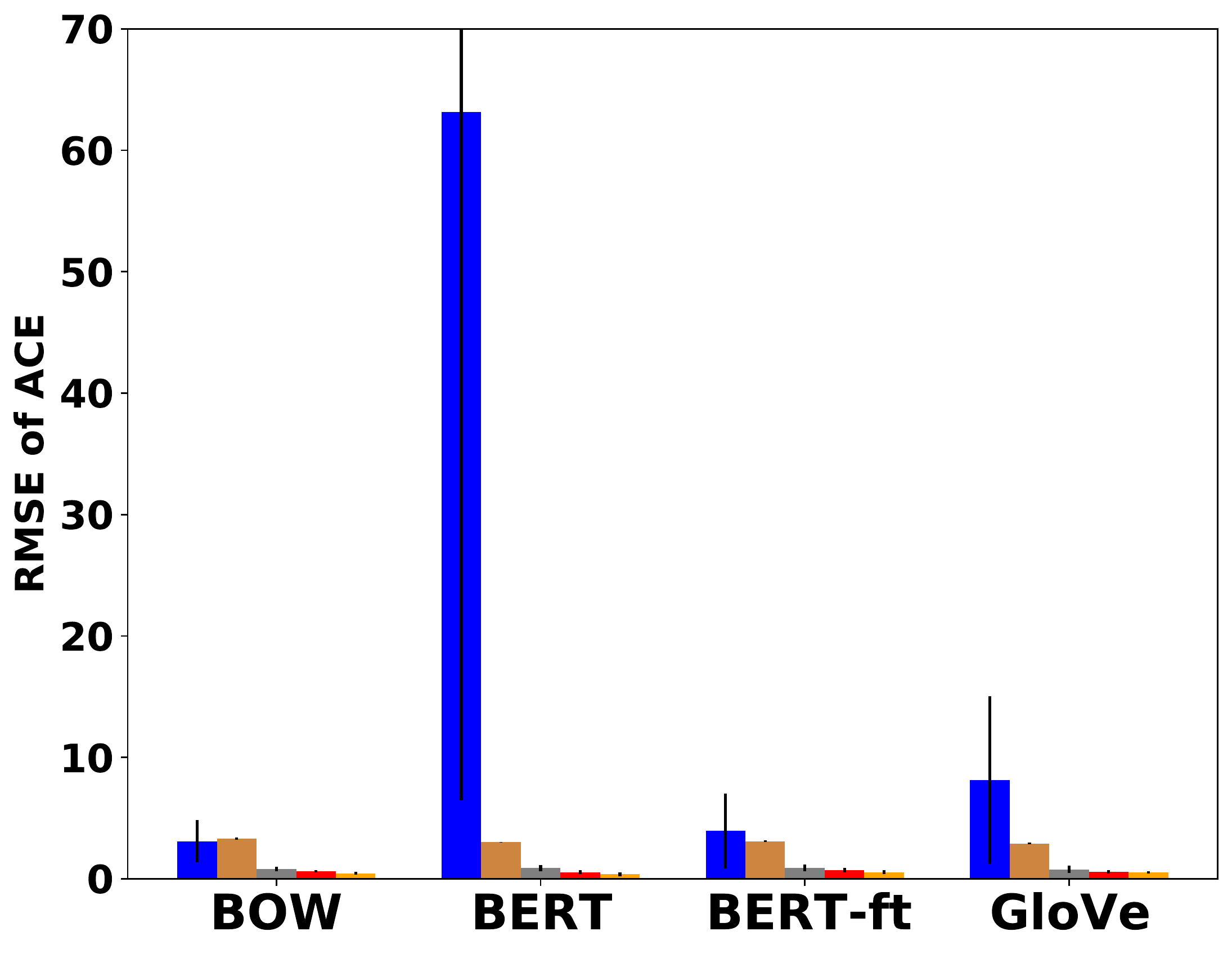}  
  \caption{Hateful, $\beta_u \sim \mathcal{N}(0,3)$}
  \label{fig:sub-second}
\end{subfigure}
\begin{subfigure}{.24\textwidth}
  \centering
  % include second image
  \includegraphics[width=\textwidth]{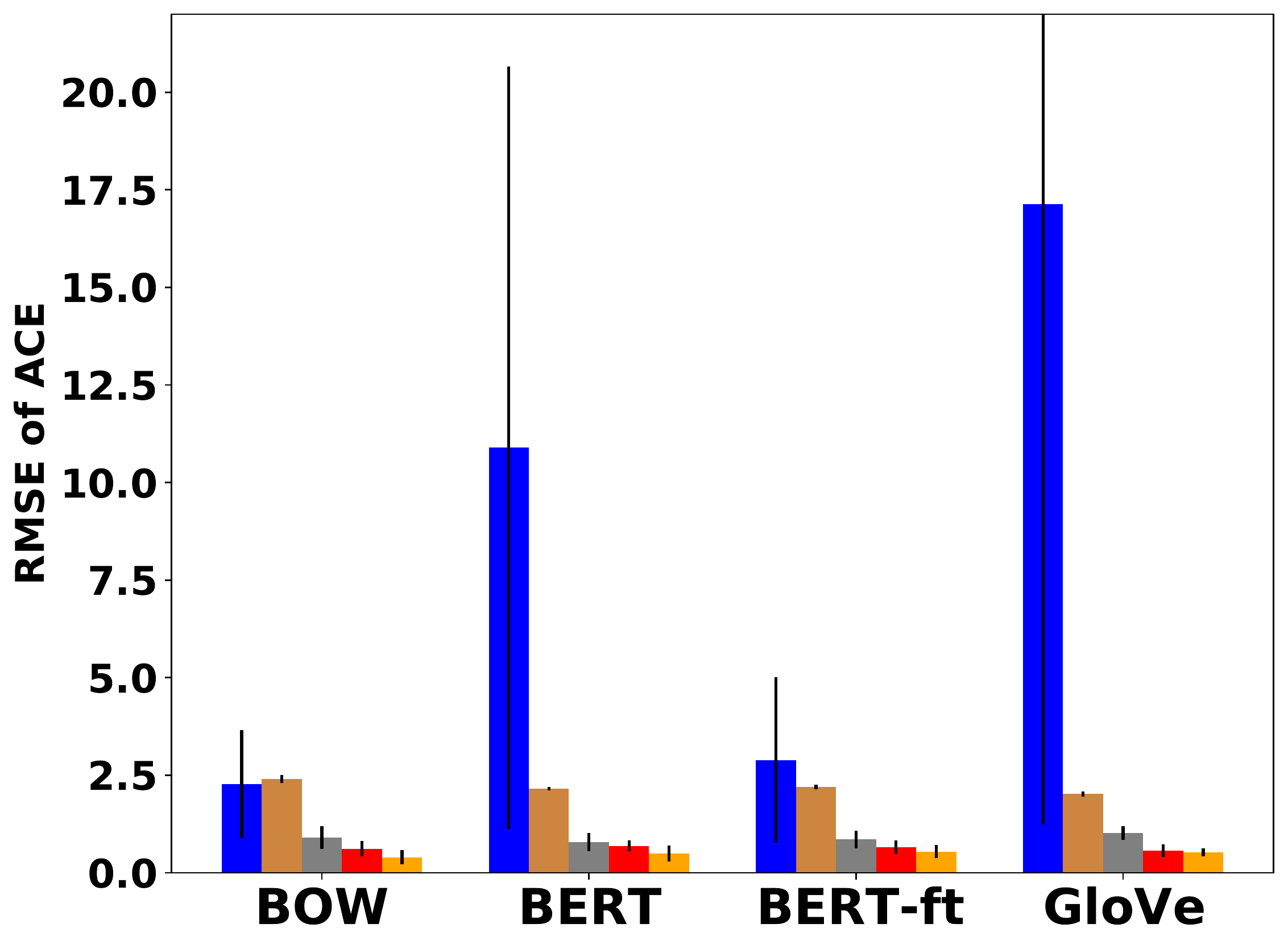}  
  \caption{Hateful, $\beta_u \sim \mathcal{N}(5,2)$}
  \label{fig:sub-second}
\end{subfigure}
\caption{Comparison of RMSE of ACE using various baseline methods in dyadic data. Error bars represent the standard deviation of the estimated effects. TSLS has significantly higher estimation bias and variance than PE-LR in all datasets.}
\label{fig:TSLE_dyad}
\end{figure*}
 }

\subsection{Experimental setup}
We consider two types of attributes $\mathbf{Z}_i$. We use
    %\item Bag-of-words (BoW): 
      bag-of-words (BoW) to represent documents as vectors (vector size of 4,939 for SAH, 13,146 for Hateful Users, 8,189 for BlogCatalg, and 12,047 for Flickr). %We utilize the \textit{CountVectorizer()} function from the \textit{Sklearn} library to obtain the BoW vectors. 
     %For both the SAH and Hateful Users datasets, this function generates BoW vectors with dimensions of $4,939$ and $13,146$, respectively.
    %\item Word embeddings: 
    To understand the value of adding VAEs to our framework, we also experiment with simple embeddings derived from BoW:
    %Word embeddings are numerical representations of words that capture semantic and syntactic information, as well as the contextual relationships between words in a document. We calculate the word embedding of each tweet by averaging the word embeddings of all tokens in the tweet. We employ two models to obtain the word embeddings. 
 %Global Vectors for Word Representation (GloVe) is an unsupervised algorithm that maps words to a vector space, in a way that semantic similarity between words is reflected by vector distances \citep{pennington-emnlp14}. We choose the 
 GloVe-200d model \citep{pennington-emnlp14} and
%trained on a dataset of $2B$ tweets, $27B$ tokens, and a vocabulary of $1.2M$ words.
 \textit{Bidirectional Encoder Representations from Transformers (BERT)} \citep{devlin-acl19}. 
%We use the pre-trained BERT-base model %containing $12$ transformer layers, $12$ attention heads in each layer, and $110M$ parameters in total.
 We also further train the BERT model for $1,000$ steps with SAH and Hateful Users datasets to get new embeddings corresponding to the context of each dataset. We refer to this model as \textit{BERT-ft}. 
%We use the $AdamW$ optimizer with a learning rate of $2e-5$, max-seq-length of $128$, and batch sizes of $32$. We refer to this model as \textit{BERT-ft}.

%\end{itemize}
%We leverage \textit{Auto-encoder} architecture for dimension reduction and obtain the latent embedding of 
 %high-dimensional proxies. An Auto-encoder is a neural network architecture that learns a compressed representation of the input. An Auto-encoder architecture consists of three modules: 1) Encoder which compresses the input data to a new representation, 2) bottleneck which contains compressed knowledge representations, and 3) Decoder which reconstructs the original representation from a new representation. The low-dimensional embedding vector generated by the trained encoder component is considered a proxy variable. The objective function of an Autoencoder is aligned with our embedding component of the ProEmb framework. 

\begin{table*}[h]
\centering
\caption{Comparison of RMSE of ACE with various baseline methods, employing the max() and mean() activation functions and BoW feature representation. Numbers following $\pm$ indicate the standard deviation of the estimates.}
\resizebox{2\columnwidth}{!}{
    \begin{tabular}{|c|c|c|c|c|c|c|c|c|c|c|}
        \cline{2-11}
        \multicolumn{1}{c|}{} & \multicolumn{5}{c|}{Max()} & \multicolumn{5}{c|}{Mean()} \\
        \hline
        Dataset & TSLS & CEVAE & NetD & T-GB & PE-GB & TSLS & CEVAE & NetD & T-GB & PE-GB \\
        \hline
        SAH & 2.75 $\pm$ 1.35 & 2.27 $\pm$ 0.09 & 0.87 $\pm$ 0.7 & 0.61 $\pm$ 0.13 & \textbf{0.4 $\pm$ 0.1} & 5.42 $\pm$ 3.6 & 2.42 $\pm$ 0.09 & 0.85 $\pm$ 0.45 & 0.65 $\pm$ 0.21 & \textbf{0.47 $\pm$ 0.24} \\
        \hline
        Hateful Users & 3.28 $\pm$ 1.96 & 2.6 $\pm$ 0.08 & 0.88 $\pm$ 0.16 & 0.58 $\pm$ 0.07 & \textbf{0.41 $\pm$ 0.08} & 4.6 $\pm$ 2.8 & 2.51 $\pm$ 0.06 & 0.66 $\pm$ 0.16 & 0.62 $\pm$ 0.11 & \textbf{0.47 $\pm$ 0.15} \\
        \hline
        BlogCatalog & 207 $\pm$ 109 & 1.83 $\pm$ 0.12 & 0.38 $\pm$ 0.13 & 0.27 $\pm$ 0.06 & \textbf{0.09 $\pm$ 0.03} & 620 $\pm$ 481 & 3.41 $\pm$ 0.25 & 0.23 $\pm$ 0.12 & 0.19 $\pm$ 0.09 & \textbf{0.11 $\pm$ 0.06} \\
        \hline
        Flickr & 128 $\pm$ 105 & 2.12 $\pm$ 0.13 & 0.46 $\pm$ 0.27 & 0.35 $\pm$ 0.11 & \textbf{0.12 $\pm$ 0.04} & 160 $\pm$ 120 & 2.76 $\pm$ 0.18 & 0.36 $\pm$ 0.21 & 0.28 $\pm$ 0.12 & \textbf{0.13 $\pm$ 0.07} \\
        \hline
    \end{tabular}
}
\label{tb:bow_bs}
\end{table*}

 \begin{figure*}[ht]
\begin{subfigure}{.24\textwidth}
  \centering
  % include second image
  \includegraphics[width=\textwidth]{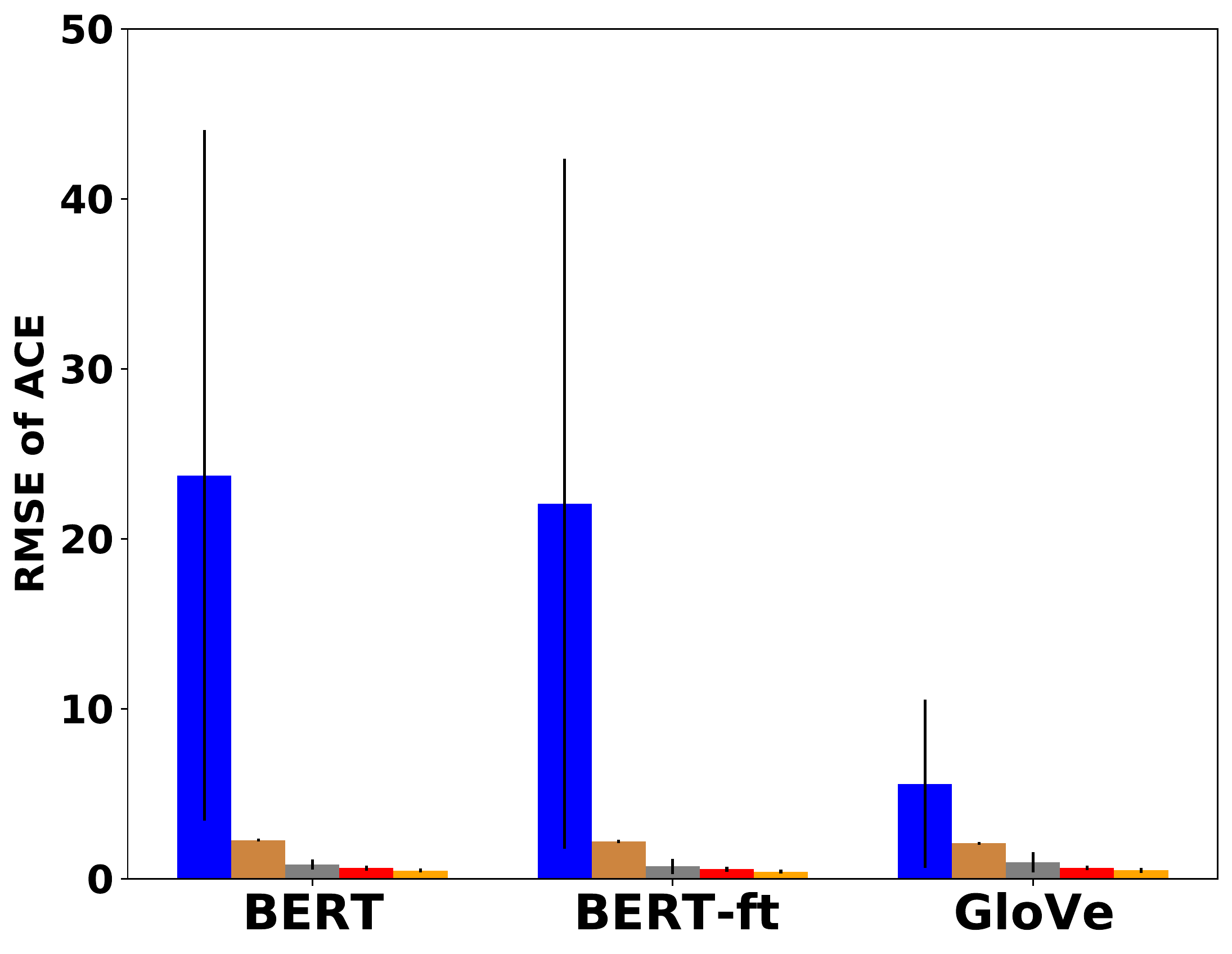}  
 % \vspace{-15pt}
  \caption{SAH, $\beta_u \sim \mathcal{N}(0,3)$}
  \label{fig:sub-second}
\end{subfigure}
\begin{subfigure}{.24\textwidth}
  \centering
  % include second image
  \includegraphics[width=\textwidth]{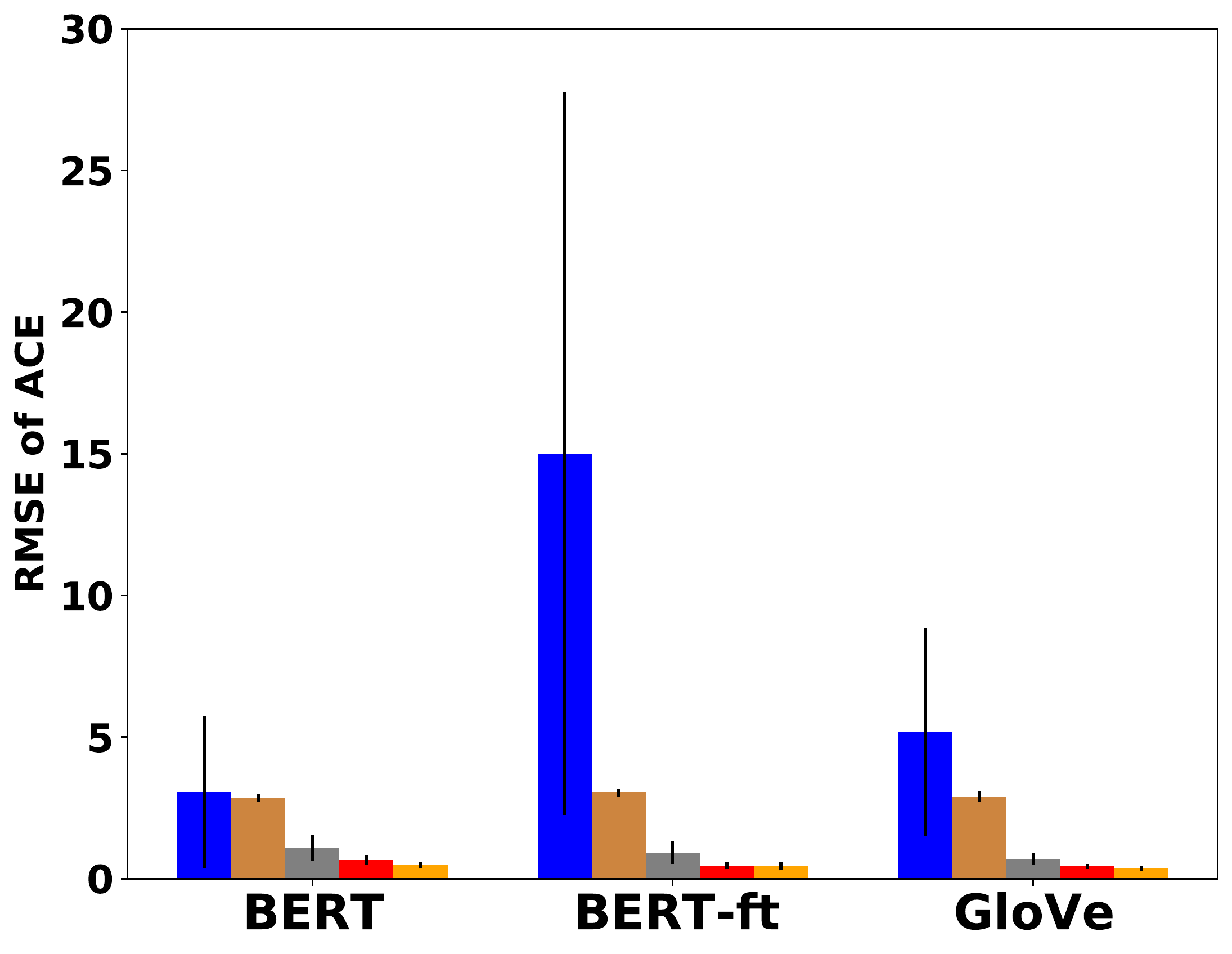}  
  %\vspace{-15pt}
  \caption{SAH, $\beta_u \sim \mathcal{N}(5,2)$}
  \label{fig:sub-second}
\end{subfigure}
\begin{subfigure}{.24\textwidth}
  \centering
  % include second image
  \includegraphics[width=\textwidth]{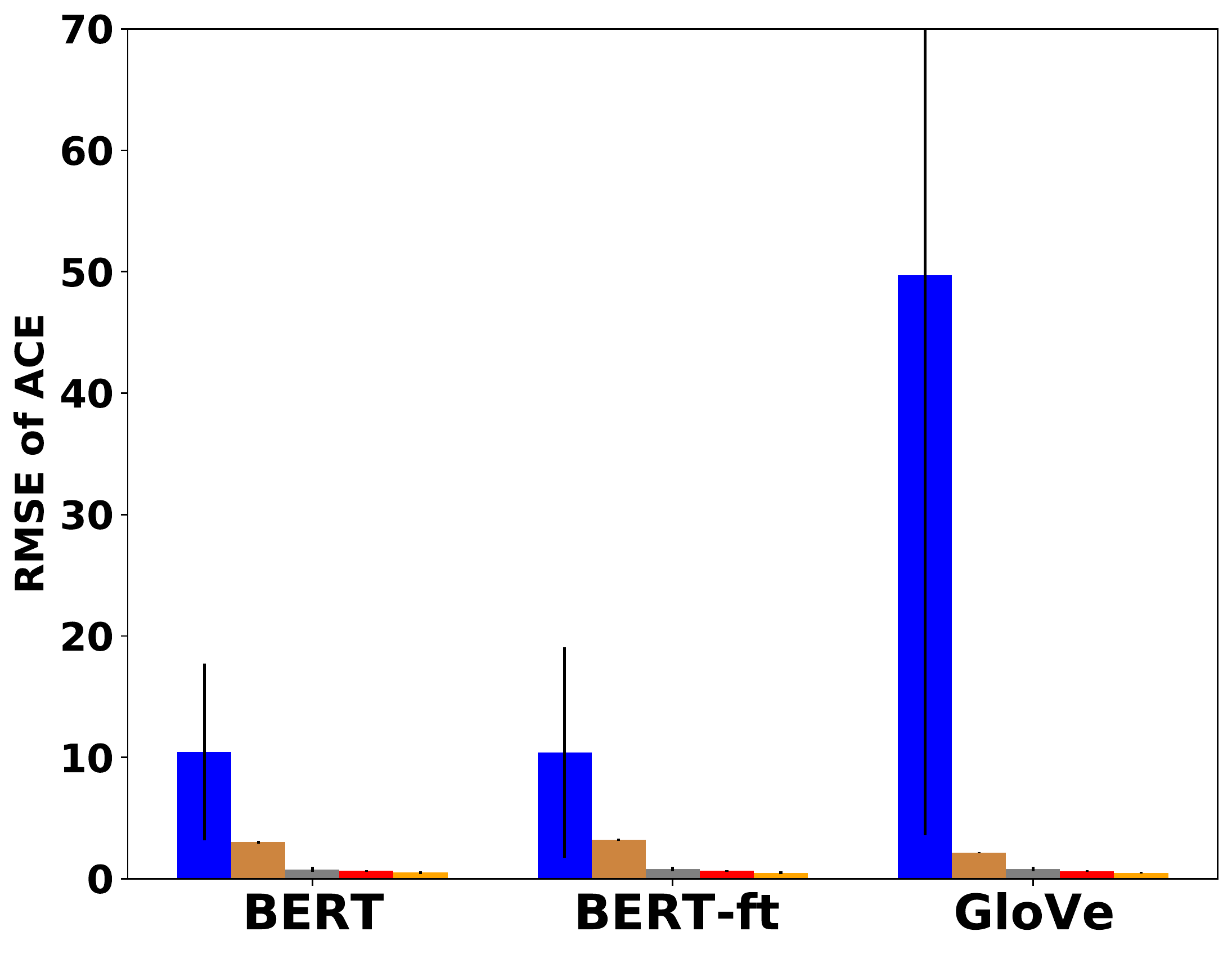} 
 % \vspace{-15pt}
  \caption{Hateful, $\beta_u \sim \mathcal{N}(0,3)$}
  \label{fig:sub-second}
\end{subfigure}
\begin{subfigure}{.24\textwidth}
  \centering
  % include second image
  \includegraphics[width=\textwidth]{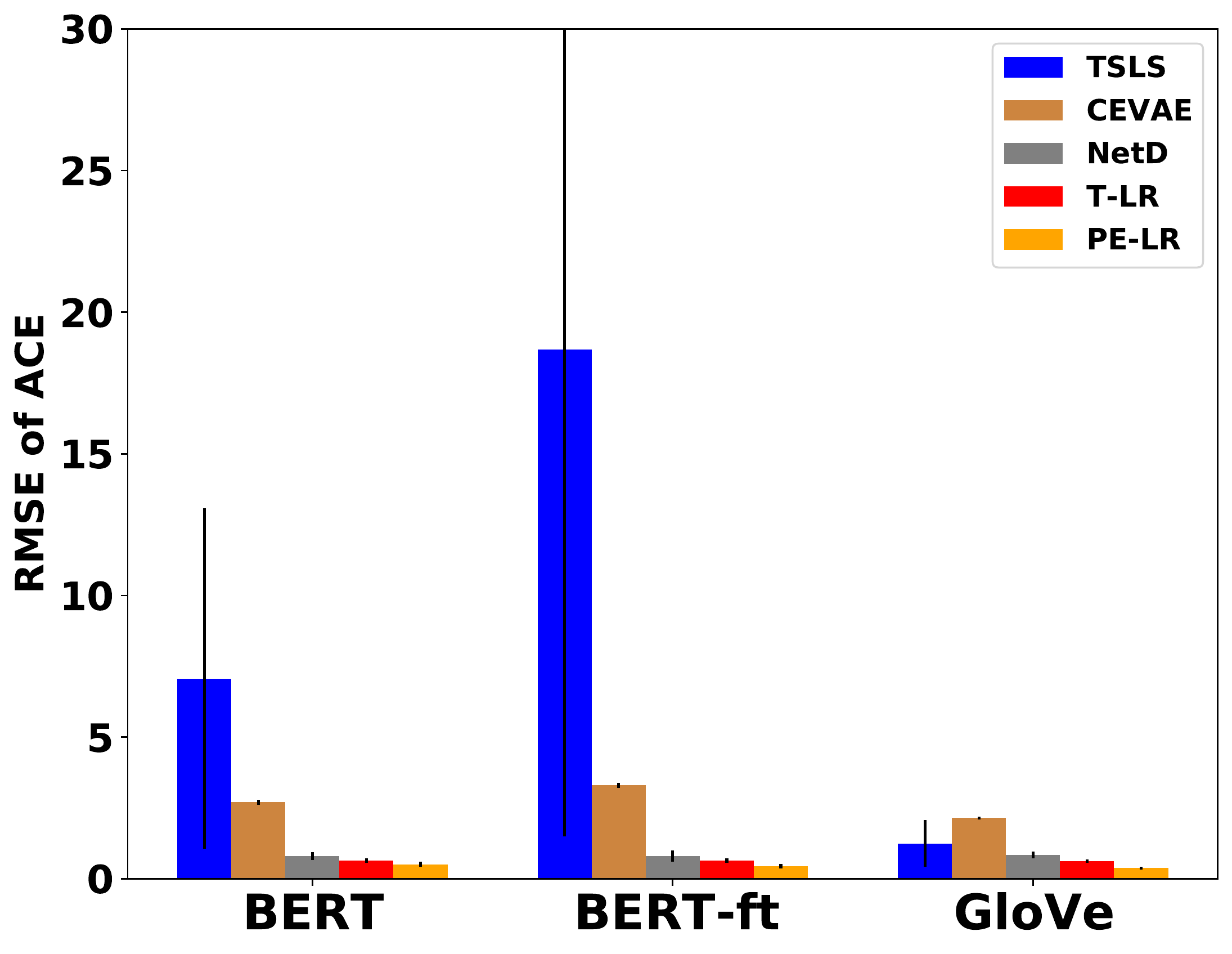}  
 % \vspace{-15pt}
  \caption{Hateful, $\beta_u \sim \mathcal{N}(5,2)$}
  \label{fig:sub-second}
\end{subfigure}
\caption{Comparison of RMSE of ACE with various baseline methods, employing the max() activation function and embeddings derived from BoW.  
%Error bars represent the standard deviation of the estimated effects.
%TSLS has significantly higher estimation bias and variance than PE-LR in all datasets.
}
%\vspace{-15pt}
\label{fig:TSLE}
\end{figure*}

 %\textbf{Counterfactual model:} 
%\begin{itemize}
%\item S-learner: In S-learner, the node's treatment is considered as a feature similar to all of the other features of the node in learning the outcome model and estimating potential outcomes.
%\item T-learner: 
%\end{itemize}

To understand the value of choosing different base-learners in our framework, we employ three types of base-learners for the T-Learner estimator: 1) Linear Regression (LR), 2) Gradient Boosted Trees (GB), and 3) Neural Network (NN) with \textit{Multi-layer Perceptrons (MLP)}. In ProEmb with LR (\textit{PE-LR}), ProEmb with GB (\textit{PE-GB}), and ProEmb with NN (\textit{PE-NN}) which are our proposed models, first, the integration of the VAEs and the discriminator is trained and applied to reduce the dimensionality of the proxies, and then T-Learner with different base-learners are deployed to predict the counterfactuals. 
We set the embedding dimension of the VAEs as the dimension of the unobserved confounder variable in each dataset (20 in SAH and 50 in the Hateful Users, BlogCatalog, and Flickr datasets).  Refer to the Appendix for more information on hyperparameter tuning.

\commentout{To train the VAEs, discriminators, and MLP models, we conduct a hyperparameter search for the learning rate and the number of epochs. The learning rate is searched within the set $\{0.1, 0.01, 0.001, 0.0001\}$, while the number of epochs is searched within $\{10, 30, 50, 70, 100\}$. The best results are achieved with a learning rate of 0.001 and 50 epochs for both models.
For the VAEs, we search the number of hidden units of the hidden layers in $\{100, 200, 300\}$ and the number of encoder and decoder layers in ${1, 2, 3, 4}$. We select a network with 100 hidden nodes, a 3-layer encoder, and a 3-layer decoder with a ReLU activation function.
In the discriminator component, after hyperparameter search, we determine that four hidden layers, with linear activation functions, produce the best performance. The output layer utilizes a Sigmoid function.
Regarding the MLP, we search for the number of hidden units and the number of fully connected layers. Ultimately, we train an MLP model with two fully connected layers, each containing 125 hidden units. We set the embedding dimension of the VAEs as the dimension of the unobserved confounder variable in each dataset (20 in SAH and 50 in the Hateful Users dataset).}

To report the estimation error of different models, we measure the \textit{Root Mean Squared Error (RMSE)} of contagion effects over 10 runs.
\commentout{
as:
\begin{equation*}
   RMSE= \sqrt{\frac{1}{S}\sum_{s=1}^{S} (\hat{ACE}_s-ACE_s)^2}
\end{equation*}
where $S$ is the number of runs, 
ACE and $\hat{ACE}$ are the true and estimated contagion effect in run $s$, respectively. In all experiments, we set $S$ to 10 and $\tau_i$ to 1 for all nodes.}
We consider the BoW or word embedding vector of each user's tweet as an NCO proxy and the BoW or word embedding vector of the peer's tweet as an NCE proxy of the hidden topic distributions. 
%As the baseline methods, we assess the performance of TSLE applied on the high-dimensional proxies and the ProEmb framework without the AE component.
%we compare the performance of ProEmb with methods without embedding components and the TSLS estimator. 
%considered as the proxies of unobserved confounder. 
\commentout{
\begin{equation}
    ivreg(Y_t \sim Y_{ngb,t-1}+Z_{ngb}|Z,Y_{ngb,t-1} , data)
\end{equation}
where $Z_{ngb}$ is the matrix of neighbors' attributes for all nodes and $Y_{ngb,t-1}$ is the matrix of neighbors' outcome at time $t-1$.}
Following \cite{egami-zrxiv2021}, we set $\beta_y=0.2$ in Eq. \ref{eq:foutcome1} and Eq.\ref{eq:fcoutcome2}. In addition, 
we vary the strength of unobserved confounding coefficient vector $\beta_u$ with two different distributions  $\beta_u \sim \mathcal{N}(5,2)$ and $\beta_u \sim \mathcal{N}(0,3)$ and $\alpha_u \sim \mathcal{N}(0,1)$. 
%To get BoW vectors with different sizes, we remove words with specific frequencies by setting $min\_df \in \{1,2,3,5,10,15,20\}$ and $max\_df\in{25,50}$ parameters of CountVectorizer function of sklearn. 

%\vspace{-30pt}
\commentout{
\begin{figure*}[ht]
\begin{subfigure}{.24\textwidth}
  %\centering
  % include second image
  \includegraphics[width=\textwidth]{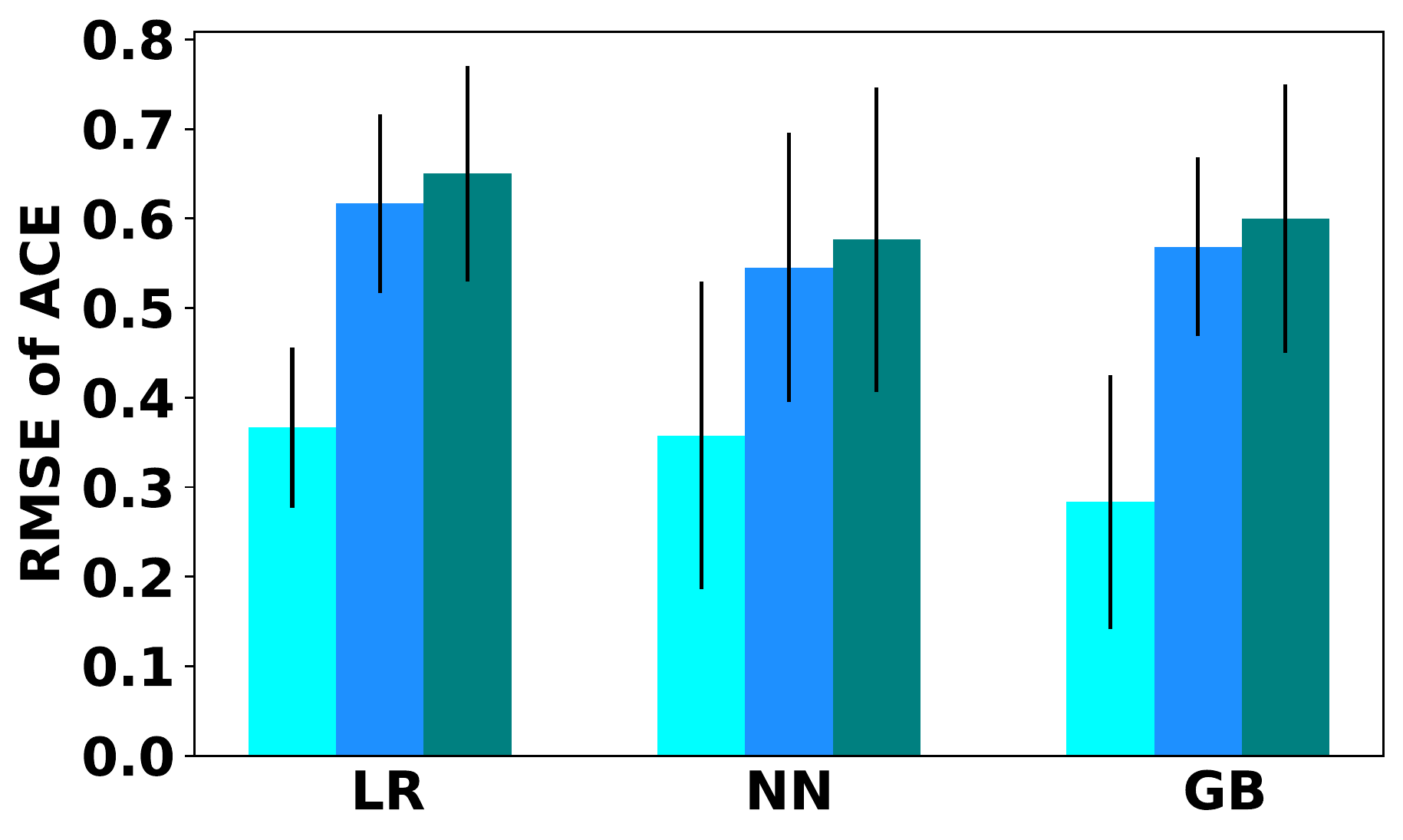}  
  \caption{SAH, $\beta_u \sim \mathcal{N}(0,3)$ }
  \label{fig:sub-second}
\end{subfigure}
\begin{subfigure}{.24\textwidth}
  %\centering
  % include second image
  \includegraphics[width=\textwidth]{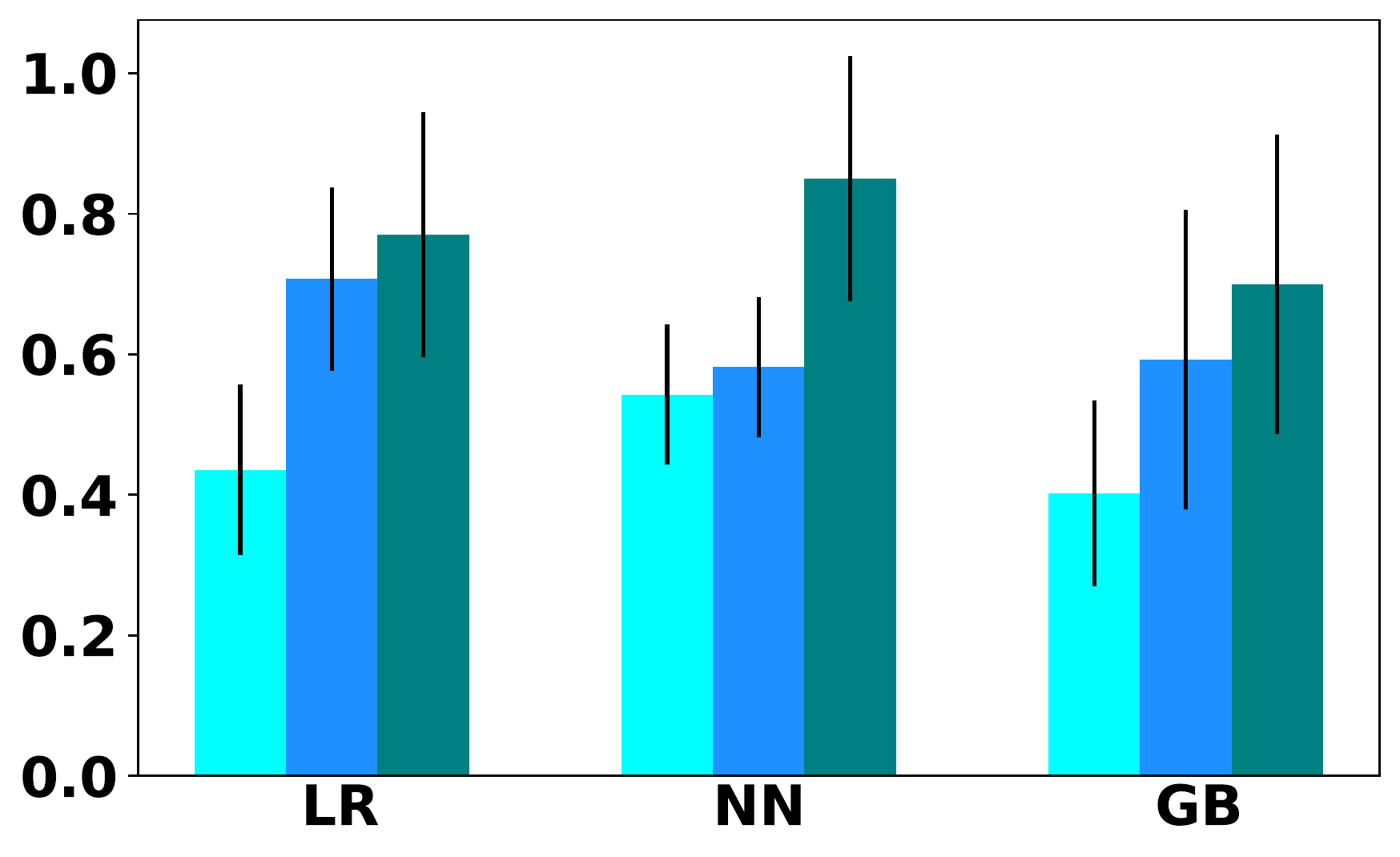}  
  \caption{SAH, $\beta_u \sim \mathcal{N}(5,2)$}
  \label{fig:sub-second}
\end{subfigure}
\begin{subfigure}{.24\textwidth}
  %\centering
  % include second image
  \includegraphics[width=\textwidth]{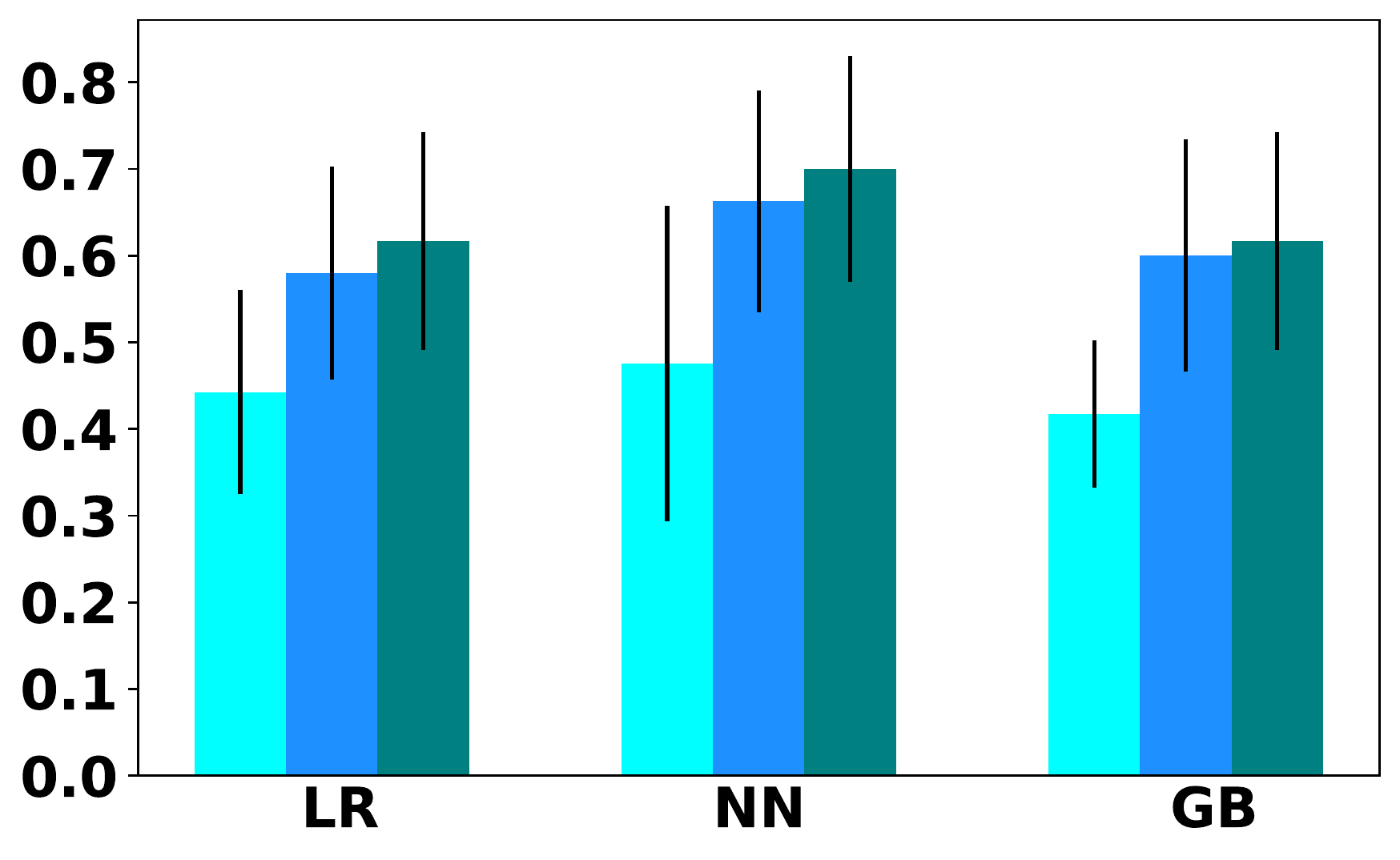}  
  \caption{Hateful, $\beta_u \sim \mathcal{N}(0,3)$}
  \label{fig:sub-second}
\end{subfigure}
\begin{subfigure}{.24\textwidth}
  %\centering
  % include second image
  \includegraphics[width=\textwidth]{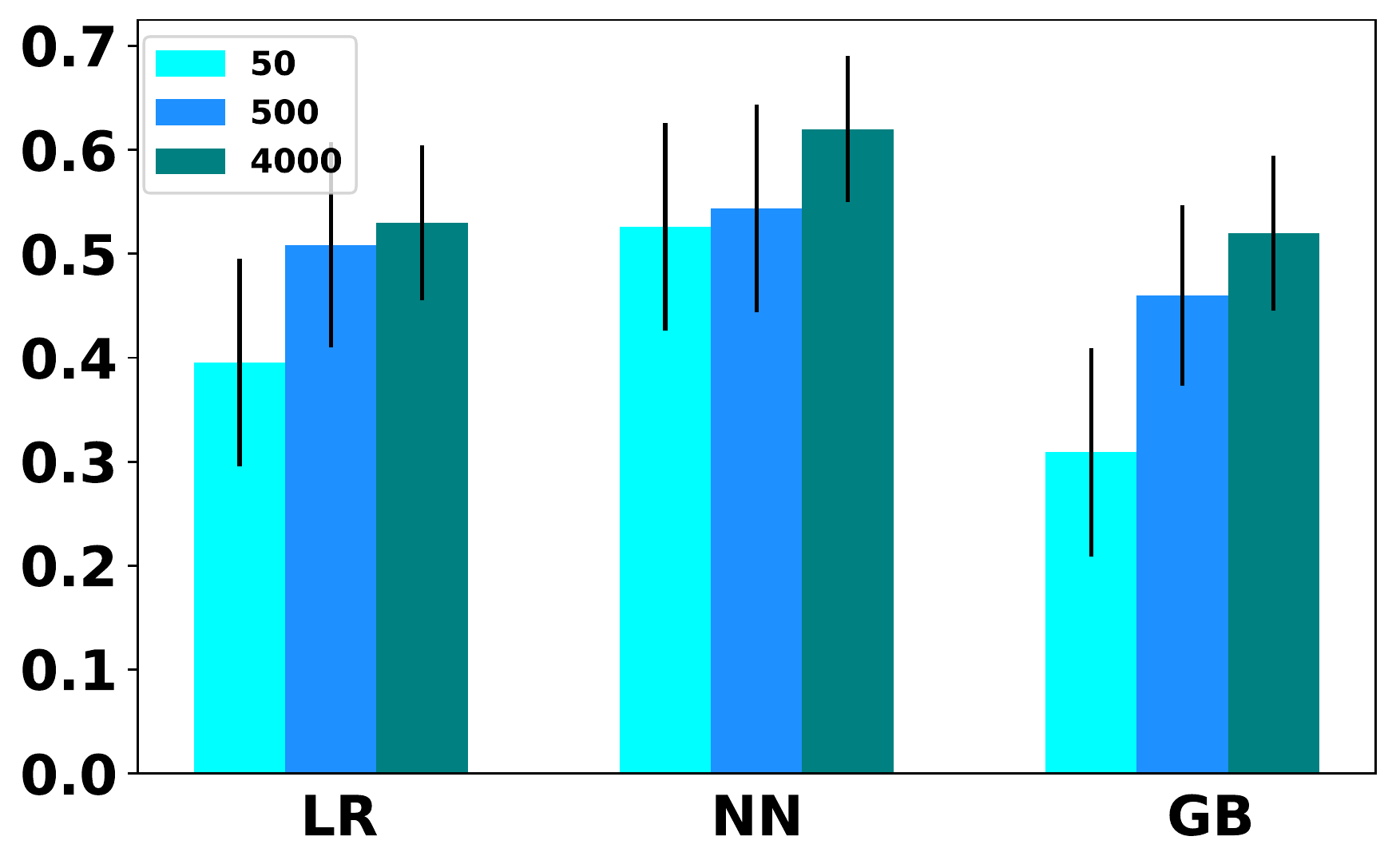}  
  \caption{Hateful, $\beta_u \sim \mathcal{N}(5,2)$}
  \label{fig:sub-second}
\end{subfigure}
\caption{RMSE of ACE in ProEmb with varying embedding vector dimensions in SAH and Hateful Users dataset with dyadic data. The x-axis represents different types of base-learners used in the counterfactual component of the ProEmb framework.}
\label{fig:AE_dim_dyad}
\end{figure*}}
\textbf{Baselines:} We compare the performance of ProEmb variants against four different baselines.
%\begin{itemize}
     TSLS is the only existing and state-of-the-art method that makes contagion effects identifiable in network data with unobserved confounders using negative control proxies \citep{egami-zrxiv2021}. 
     %We use \textit{ivreg} function from $R$ software to fit a TSLS estimator and infer the coefficient of $Y_{{ngb},t-1}$ showing the contagion effects.
    %As shown in Eq. \ref{eq:tsls}, the coefficient of $\textbf{Y}_{ngb,t-1}$ represents the peer effects of interest \citep{egami-zrxiv2021}.
    \textit{Causal Effect Variational Autoencoder (CEVAE)} is a VAEs-based model for inferring ITE with unobserved confounders \citep{louizos-neurips17}. Although this model is primarily intended for non-network datasets, we adapt it to network data by concatenating available proxies for the unobserved confounders ($\mathbf{Z}_i$ and $\mathbf{Z}_{ngb}$) as the noisy proxy vector for each node.
      \textit{Network Deconfounder (NetD)} 
      %utilizes network information to infer latent confounders and estimate ITE \citep{guo-wsdm20}. 
       exploits \textit{Graph Convolutional Networks (GCNs)} to learn representations of hidden confounders by mapping features and network structure into a shared representation space \cite{guo-wsdm20}. 
      %An output function is then learned to infer the potential outcome of a node based on the treatment and the representation of hidden confounders.
      We also consider only a T-Learner with Linear Regression (\textit{T-LR}), Gradient Boosted Tree (\textit{T-GB}), and neural network (\textit{T-NN}) as the base-learners.
%\end{itemize}

%\fixme{\textbf{Real world demonstration}: In the Appendix, we demonstrate the applicability of our approach for detecting contagion effects in the 2017 French presidential election \cite{burghardt-icwsm23}.}

\subsection{Results}
\subsubsection{Comparison to all baselines.}
%We evaluate the performance of different estimators to measure contagion effects.
In Table \ref{tb:bow_bs}, we provide a comprehensive comparison of our method, PE-GB, with all baseline models (\textit{TSLS}, \textit{CEVAE}, \textit{NetD}, and \textit{T-GB}), assessing their performance in estimating ACE using BoW features as proxy variables across all datasets. We employ both the max() and mean() activation functions to evaluate the models. The values following \stackanchor{+}{-} represent the standard deviation of the estimate results.
%In Fig. \ref{fig:TSLE}, we present a comparison between all baselines (\textit{TSLS}, \textit{CEVAE}, \textit{NetD}, and \textit{T-LR}) and our method PE-LR  in estimating ACE using various features as proxy variables in both datasets with network data.
The results show that in all datasets \textit{TSLS} consistently achieves significantly higher error and variance compared to the other models, especially our proposed method \textit{PE-GB}. This was one of the most surprising results in our study since \textit{TSLS} is a well-established estimation method in causal inference.
Additionally, it's worth noting that CEVAE, a method that utilizes VAEs for causal effect inference in non-network data, demonstrates relatively lower performance when contrasted with our approach, \textit{PE-GB}. This observation highlights the significance of mitigating the representation mismatch between treatment and control when estimating contagion effects.
%\fixme{This is more significant in BlogCatalog and Flickr datasets, probably because of the }
%For instance, in the SAH dataset with $\beta_u \sim \mathcal{N}(0,3)$, we observe errors of $2.71$ versus $0.46$ with BoW, $23.73$ vs. $0.50$ with BERT, $22.07$ vs. $0.44$ with BERT-ft, and $5.58$ vs. $0.55$ with GloVe. 
Among the four baselines, the \textit{T-GB} model significantly reduces estimation errors which is why we used T-Learner as a baseline in the third experiment. We obtain consistent results in data on dyads (Fig. \ref{fig:TSLE_dyad}).
%By conducting this evaluation, we provide robust evidence supporting the superior performance of T\_LR in estimating contagion effects, highlighting their effectiveness across different datasets and features.

\subsubsection{Sensitivity to word embedding methods.}
In this experiment, we evaluate the performance of baseline methods using various word embedding techniques. Due to the unavailability of the original text from the BlogCatalog and Flickr datasets, we present the results exclusively for the SAH and Hateful Users datasets. As depicted in Fig. \ref{fig:TSLE}, our observations consistently align with those obtained using the BoW method.
It is evident that TSLS exhibits the highest levels of bias and variance when estimating contagion effects across different embedding techniques for both datasets with different unobserved confounding coefficients. However, while Netd outperforms both TSLS and CAVAE, it doesn't quite reach the level of effectiveness demonstrated by our proposed method, \textit{PE-GB}.

\begin{figure*}[ht]
 %\vspace{-10pt}
\begin{subfigure}{.47\textwidth}
  \centering
  % include second image
  \includegraphics[width=\textwidth]{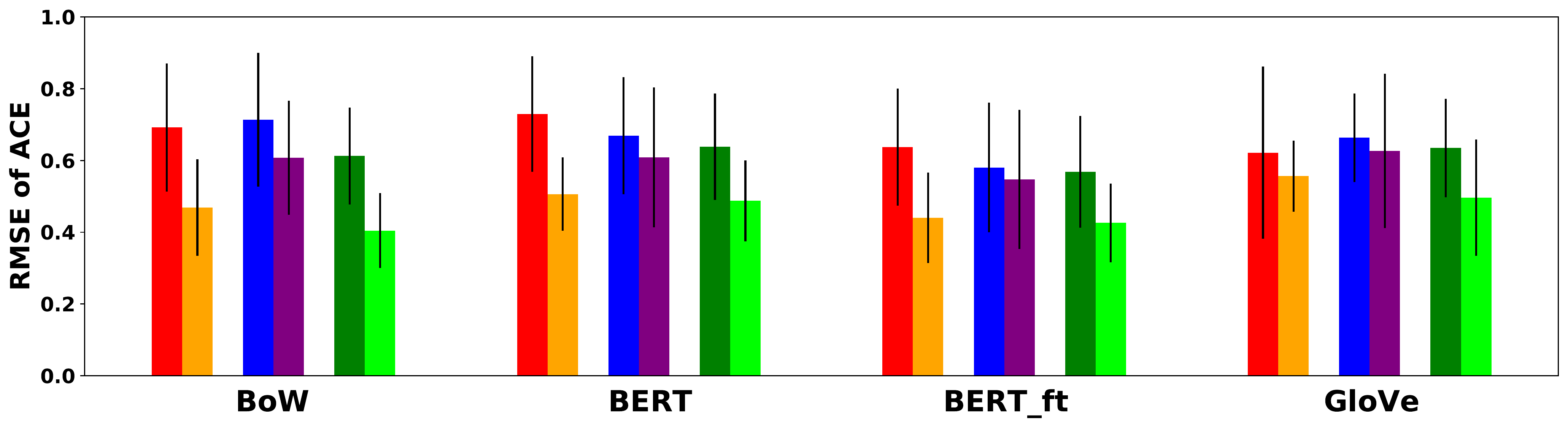} 
  %\caption{SAH}
  \label{fig:sub-second}
\end{subfigure}
\begin{subfigure}{.51\textwidth}
  \centering
  % include second image
  \includegraphics[width=\textwidth]{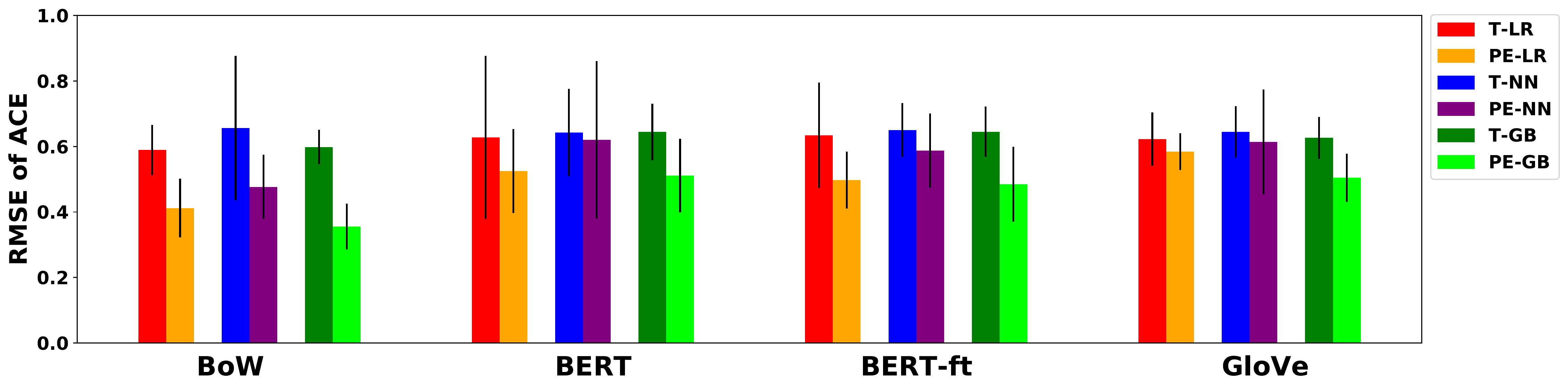}  
  %\caption{Hateful Users}
  \label{fig:sub-second}
\end{subfigure}
%\begin{subfigure}{\textwidth}
  %\centering
  % include second image
  %\includegraphics[width=\textwidth]{figs/SAH_mu5_std2.pdf}  
  %\caption{ $\beta_u \sim \mathcal{N}(\mu=5,\,\sigma^{2}=2)$}
  %\label{fig:sub-second}
%\end{subfigure}
\caption{RMSE of ACE in SAH (left) and Hateful Users dataset (right), considering network data and utilizing the max() activation function, with $\beta_u \sim \mathcal{N}(0,3)$.
%Bars show the standard deviation of the estimated effect. 
}
\label{fig:SAH}
\end{figure*}
\begin{figure*}[ht]
\begin{subfigure}{.24\textwidth}
  %\centering
  % include second image
  \includegraphics[width=\textwidth]{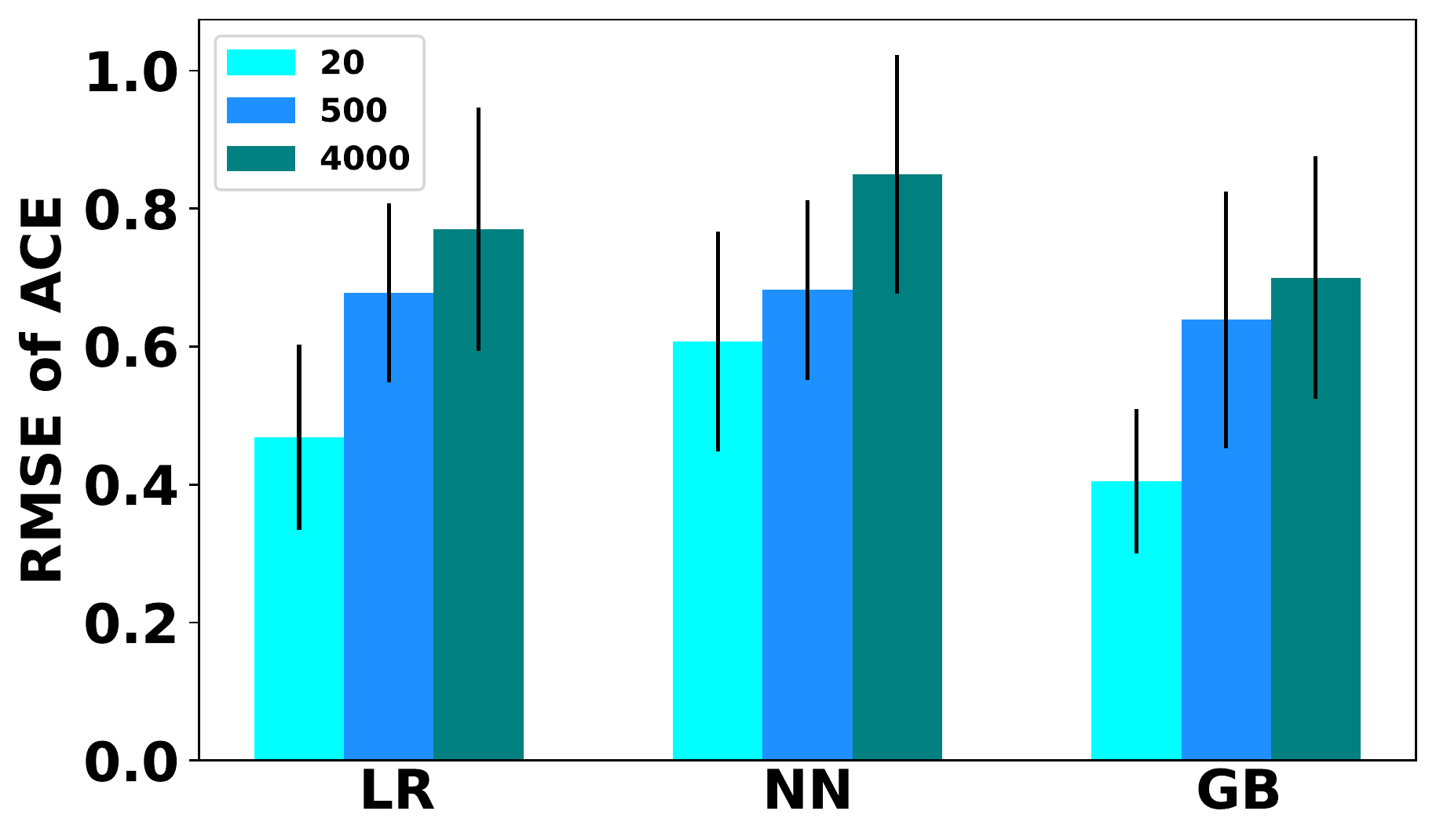} 
  %\vspace{-15pt}
  \caption{SAH}
  \label{fig:sub-second}
\end{subfigure}
\begin{subfigure}{.24\textwidth}
  %\centering
  % include second image
  \includegraphics[width=\textwidth]{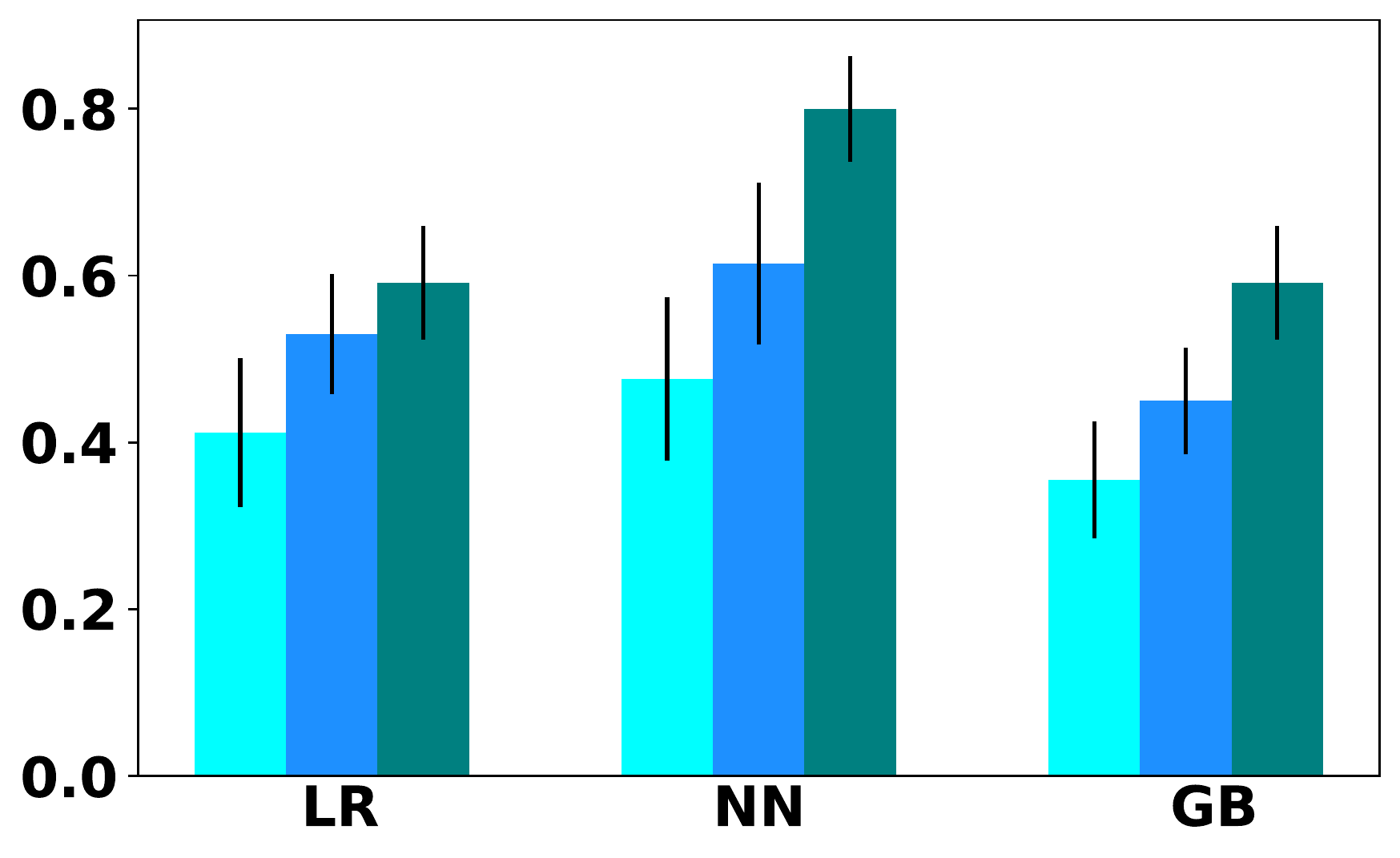} 
  %\vspace{-15pt}
  \caption{Hateful Users}
  \label{fig:sub-second}
\end{subfigure}
\begin{subfigure}{.24\textwidth}
  %\centering
  % include second image
  \includegraphics[width=\textwidth]{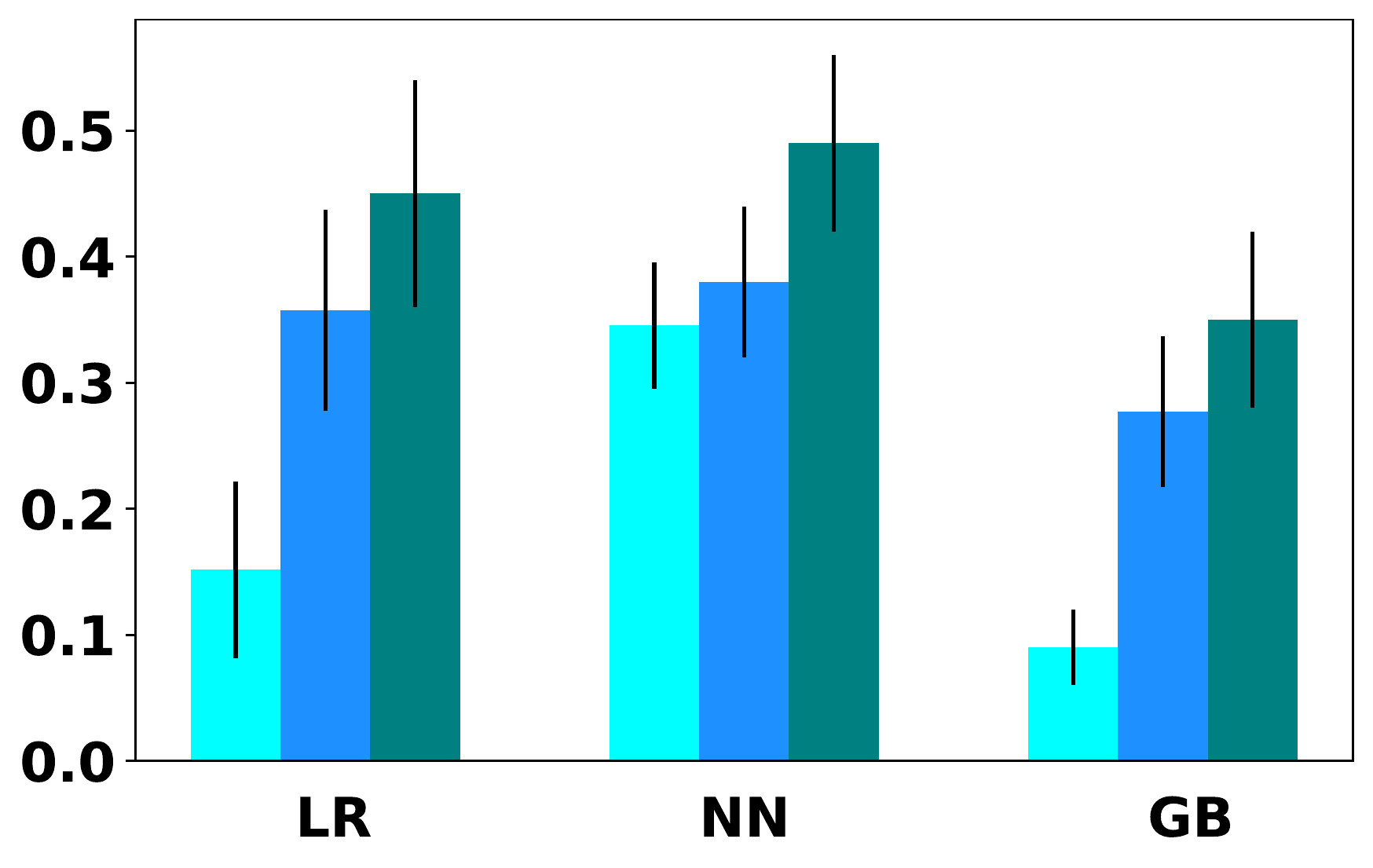} 
  %\vspace{-15pt}
  \caption{BlogCatalog}
  \label{fig:sub-second}
\end{subfigure}
\begin{subfigure}{.24\textwidth}
  %\centering
  % include second image
  \includegraphics[width=\textwidth]{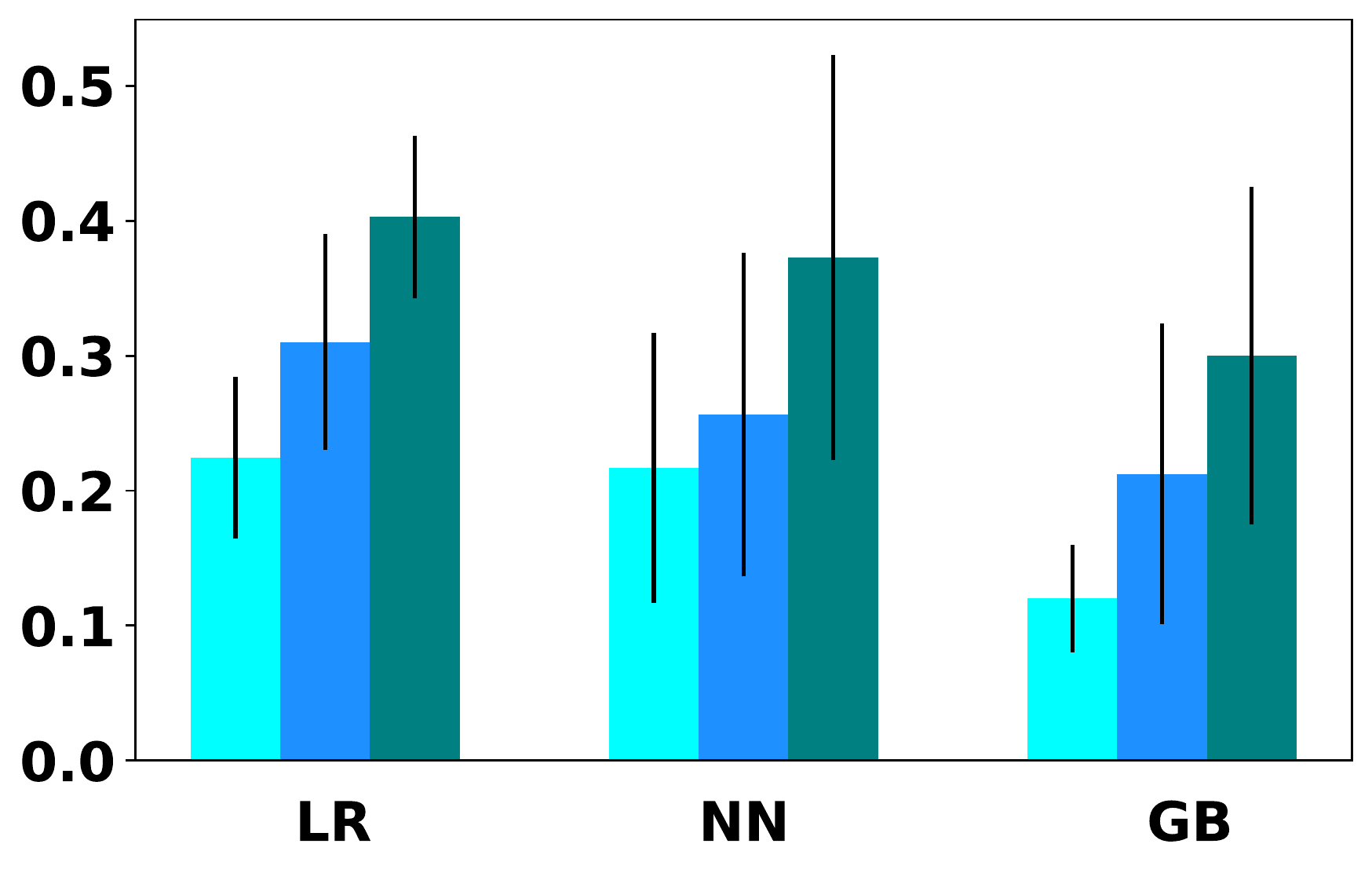}  
  %\vspace{-15pt}
  \caption{Flickr}
  \label{fig:sub-second}
\end{subfigure}
\caption{RMSE of ACE in ProEmb with varying embedding vector dimensions in four datasets and $\beta_u \sim \mathcal{N}(0,3)$ employing max() activation function. The x-axis represents different types of base-learners used in the counterfactual component of ProEmb.}
\label{fig:AE_dim}
\end{figure*}

\subsubsection{Comparison between ProEmb variants.}
%To assess the performance of variants of the ProEmb in reducing contagion effect estimation error, 
We compare the RMSE of estimated contagion effects using ProEmb and the best-performing baseline, T-Learner, with different base-learners. 
As Fig. \ref{fig:SAH} shows, \textit{PE-LR} and \textit{PE-GB} outperform other methods in both datasets. 
%For example, PE-LR with BoW features obtains $33.3\%$ and $30.8\%$ estimation error reduction compared to T-LR in SAH. 
%Results are consistent in the Hateful Users dataset and graphs on dyads. %PE-LR with BoW features reduces the T-LR estimation error by $29.3\%$ and $16.3\%$  with $\beta_u \sim \mathcal{N}(0,3)$ and $\beta_u \sim \mathcal{N}(5,2)$, respectively.
Results show that BoW is a more effective feature representation method. One possible explanation is that the topic distribution is derived from BoW vectors, which could serve as a more suitable representation of the latent topics.
The RMSE for PE is consistently lower for different base-learners and different features even if they do not always appear to be statistically significant due to overlapping variance.
%The results for dyadic data are consistent with findings from datasets with network data.
%Due to space constraints, we omit the results showing the comparison between T-Learner and ProEmb variants with $\mathcal{N}(5,2)$ and dyadic data.
%Plots for SAH and Hateful Users datasets with $\beta_u \sim \mathcal{N}(\mu=5,\,\sigma^{2}=2)$ and dyadic data can be seen in Appendix. 
%These results are consistent with findings from datasets with network data. ProEmb variants achieve better contagion effect estimation compared to T-Learner variants.
%In the datasets with dyadic model, we observe that T\_LR with BoW vectors as the proxy variables gets the least error compared to other features 
%(0.14 in SAH and 0.17 in Hateful Users dataset with $\beta_u \sim \mathcal{N}(\mu=0,\,\sigma^{2}=3)$, $0.2$ in SAH and $0.18$ in Hateful Users dataset with $\beta_u \sim \mathcal{N}(\mu=5,\,\sigma^{2}=2)$). One potential reason is that topic distribution is extracted from BoW vectors and they can be a more appropriate representative of the hidden topics. %Results on the network data are consistent with dyadic data where AE\_T\_LR and AE\_T\_GB with BoW outperform other methods. 
%due to two reasons: \fixme{1) the outcome model is linear and NN models perform well in non-linear regression, and 2) Adding} 
Results across datasets show that \textit{PE-GB} obtains better performance compared to other variants in most cases.
%, especially with GloVe features. 

\commentout{T\_LR with BoW vectors as the proxy variables gets the least error compared to other features (0.35 in SAH and 0.41 in Hateful Users dataset with $\beta_u \sim \mathcal{N}(0,3)$, 0.47 in SAH and 0.43 in Hateful Users dataset with $\beta_u \sim \mathcal{N}(5,2)$). One potential reason is that topic distribution is extracted from BoW vectors and they can be a more appropriate representative of the hidden topics. Results on the network data are consistent with dyadic data where AE\_T\_LR and AE\_T\_GB with BoW outperform other methods. 
}
%Details on ACE estimation error of different methods in the SAH dataset with network data can be seen in Appendix.

%Figure \ref{} show consistent results in the Hateful Users dataset. 
\subsubsection{Sensitivity to the dimension of the embedding}
To investigate the impact of the embedding vector dimension on the estimation error of ProEmb variants, we train the ProEmb models with BoW features and different numbers of embedding dimensions from 20 to 4000. Due to the dimensionality and the superior performance of various methods when using BoW feature representation, we employ BoW feature representation in this experiment.
%the performance of the ProEmb framework with various base-learners in the SAH and Hateful Users datasets using network data.
% ProEmb variants with a small number of dimensions, specifically 20 in the SAH dataset and 50 in the Hateful Users dataset, yield the lowest estimation error across both datasets. 
As the number of dimensions increases (Fig. \ref{fig:AE_dim}), the estimation error also increases for all ProEmb variants, with \textit{PE-GB} achieving
%Among the variants, the estimation error of ProEmb_T_GB exhibits more stability as the number of dimensions increases compared to the other variants. However, the overall trend is an increase in estimation error with higher dimensionality.
%For instance, in the SAH dataset, increasing the number of dimensions from 20 to 4,000 leads to an increase in the estimation error of the PE-GB variant from 0.4 to 0.7 for graphs with $\beta_u \sim \mathcal{N}(0,3)$, and from 0.43 to 0.52 for graphs with $\beta_u \sim \mathcal{N}(5,2)$.
%Similarly, in the Hateful Users dataset, increasing the number of dimensions from 50 to 4,000 results in an increase in the estimation error of the PE-LR variant from 0.41 to 0.59 for graphs with $\beta_u \sim \mathcal{N}(0,3)$, and from 0.46 to 0.6 for graphs with $\beta_u \sim \mathcal{N}(5,2)$.
%These results indicate that PE-GB with a dimension of 20 in the SAH dataset and 50 in the Hateful Users dataset achieves 
the lowest error among all ProEmb variants. 
%The results for the network model with the mean() activation function and dyadic interactions can be found in the appendix.
The results are consistent across different datasets, with network and dyads ego-network, and when utilizing mean() activation functions, as shown in the Appendix (Fig. \ref{fig:AE_dim_dyad} and Fig. \ref{fig:AE_dim_mean}).
%Similar findings are observed with experiments on dyadic data.

\subsection{Real world demonstration}
One of the main challenges in social studies is measuring the effect of friends on their peers and the strength of such effects in different domains. As a demonstration of the applicability of our approach to detecting contagion effects in real-world scenarios, we analyze two datasets: 1) French Election, and 2) Peer Smoking. 
French Election is a Twitter dataset about the 2017 French presidential election \cite{burghardt-icwsm23}. This dataset comprises of 5.3M tweets related to the election, encompassing attitudes, concerns, and emotions expressed in each tweet. Our objective is to measure the extent to which a friend's tweet with a specific emotion or attitude influences a user's decision to post a tweet with the same emotion or attitude.
We consider four different outcomes in our analysis: 1) vote against which represents the author’s attitude toward voting against a candidate, 2) anger emotion, 3) love emotion, and 4) religious concern. The preprocessing details about this dataset are available in the Appendix.
Overall, we find:
\begin{itemize}
    \item Friends' tweets about voting against a candidate have a small negative effect, meaning that they are less likely to tweet about it themselves ($\hat{\theta}_{PE-GB} = -0.013$, p-value=0.001).
    \item Our method does not reveal a significant contagion effect between users regarding concerns related to religion ($\hat{\theta}_{PE-GB} = -0.002$, p-value = 0) or love emotion ($\hat{\theta}_{PE-GB} = 0.007$, p-value=0.019).
    \item The anger emotions expressed by peers in their tweets have a small negative impact on the emotional tone of users who retweet those posts, leading to a tendency for opposite emotions to be reflected in their retweets ($\hat{\theta}_{PE-GB}  = -0.016$, p-value=0).
\end{itemize}

The Peer Smoking dataset comprises 1,263 9th and 10th graders from 16 high schools in the Chicago area, observed across three distinct waves \cite{mermelstein-ntr09}. Our primary objective with this dataset is to assess the influence of peer smoking behaviors during Wave I on an individual's smoking habits during Wave II. We filter the dataset for youth who do not smoke cigarettes in Wave I.
We investigate two scenarios: 1) to what extent does an individual's boyfriend or girlfriend's smoking behavior affect their own smoking habits?, and 2) How does the smoking behavior of the group of friends an individual hang out with influence their own smoking habits? We examine cigarette smoking habits as outcomes. In the first scenario, we have 14 features for the individual and 14 features for their neighbors. In the second scenario, there are 16 features for the individual and 13 features for their neighbors. To prepare the dataset for analysis, we apply one-hot encoding to convert categorical features into numerical representations. In the first scenario, this results in 51 features for the central node and 84 features for their neighbors. In the second scenario, the individual is characterized by 57 features, while their neighbors have 78 features each.
In summary, our findings reveal the following:
\begin{itemize}
    \item The cigarette smoking habits of boyfriends or girlfriends have a positive effect on the individual's cigarette smoking behavior ($\hat{\theta}_{PE-GB}=0.112$, p-value= 0.0005). TSLS estimates a similar contagion effect in this scenario; however, it lacks statistical significance ($\hat{\theta}_{TSLS}=0.167$, p-value= 0.366).
    %TSLS: 0.167, p-value=0.366
    %boyfriend->smoking (0.105)
    %\item Our analysis does not uncover a statistically significant relationship between the cigarette smoking habits of boyfriends or girlfriends and the individual's marijuana smoking behavior. This may be attributed to the distinct nature of these two behaviors ($\hat{\theta}_{PE-GB} =0.031$, P-value=0.183).
    %boyfrined -> marijuna 0.022
    \item The cigarette smoking habits of the individual's circle of close friends, have a lower but also positive effect on the cigarette behavior of the individual ($\hat{\theta}_{PE-GB} =0.061$, P-value=0.0001).  While TSLS identifies a somewhat stronger contagion effect, this effect is not statistically significant ($\hat{\theta}_{TSLS}=0.127$, p-value= 0.408).
    %TSLS: 0.127, p-value=0.408
    %\item Friends->mari smoking behavior (PE-GB=0.086)
\end{itemize}

\section{Conclusion}
\label{conclusion}
In this paper, we introduce the Proximal Embeddings (ProEmb) framework for increasing the accuracy of contagion effect estimation in observational network data affected by latent homophily and selection bias. Our framework comprises three key components: 1) embedding learning, which utilizes Variational Autoencoders to map high-dimensional proxies to low-dimensional representations and capture latent homophily, 2) representation balancing, which leverages adversarial networks to address the representation mismatch between treatment groups' proxy representations, and 3) counterfactual learning, which employs meta-learners to estimate counterfactual outcomes.
Our results demonstrate the superiority of the ProEmb framework compared to the baselines in reducing the contagion effect estimation error. 
We observe that using our embedding framework with Bag-of-words as the input representation yields better results compared to off-the-shelf word embedding methods.
A potential future direction is developing a framework to account for multi-hop contagion effects in observational data with unobserved confounders.

%% The file named.bst is a bibliography style file for BibTeX 0.99c
\bibliographystyle{ACM-Reference-Format}
\bibliography{0-main}

\pagebreak 

\appendix
\section{Appendix}
%\appendixpage
%\addappheadtotoc

%\begin{center}
 %   \Large
%    \textbf{Supplement to Contagion Effect Estimation Using Proximal Embeddings}
        
%\end{center}

%\title{Supplement to Contagion Effect Estimation Using Proximal Embeddings}
%\maketitle
\subsection{Experiments}

\subsubsection{Datasets}
\textbf{Hateful Users} dataset is a sample of Twitter's retweet graph in which users are annotated as hateful or non-hateful \citep{ribeiro-icwsm18}. 
%and consists of $100,386$ users annotated as hateful or non-hateful, with $200$ most recent tweets for each user \citep{ribeiro-icwsm18}. 
In this dataset, we are interested in measuring the causal effect of 5,000 users' hatefulness on the time their peers spend on the Twitter application/website. 
 %We randomly sample $5,000$ tweets to create the final Hateful Users dataset and 
%\fixme{ We learn 50 LDA topics (selected based on coherence) from $5,000$ tweets to obtain the unobserved confounder vector.}
We extend this dataset by synthetically generating 1) connections representing user $v_i$ retweeting user $v_j$'s tweet, 2) the outcome of each node at time $t-1$, indicating the hatefulness of the user, and 3) the outcome of each node at time $t$, indicating the amount of time the individual spends on the Twitter application/website.
%ElBoW curve shows that with 50 topics, we obtain a reasonable coherence score. \\
%In this dataset, we learn $50$ LDA topics from tweets.\\
 %To obtain $U_i$, 50 LDA topics are learned from the tweets. 
 \\
 \textbf{Stay-at-Home (SAH)} dataset comprises $30,000$ tweets U.S.-based English tweets reflecting users' attitudes toward stay-at-home orders during the COVID-19 pandemic \citep{fatemi-dsaa22}.
 %In our experiments, we learn $20$ LDA topics to capture the unobserved confounder vector.
 Our objective is to investigate the causal effect of having friends who follow social distancing orders
 on the user's health.
 We augment the SAH dataset by synthetically generating: 1) connections indicating that user $v_i$ retweeted user $v_j$'s tweets, 2) the outcome of each node at time $t-1$, representing the stance of the tweet toward stay-at-home mandates, and 3) the outcome of each node at time $t$, indicating the level of user's health. 
 \\
 \textbf{BlogCatalog} dataset comprises communities of bloggers actively contributing content within the online BlogCatalog community. The dataset's features are constructed as bag-of-words representations, capturing the keywords found within bloggers' personal descriptions. The connections within the dataset symbolize the relationships of friendship and interaction between pairs of bloggers.
 \\
 \textbf{French Election} dataset contains 5.3M tweets related to the election. We begin by filtering this dataset to include only tweets and retweets that were posted before the second election date (May 2023), resulting in 4.2M tweets. Then, we construct the retweet network containing 3.1M connections. Following this, we filter the dataset for tweets from users who tweeted at least one tweet after retweeting a tweet. This process yields a total of 13k users with 190k tweets. 
%We use \textit{XLM-RoBERTa-base} to obtain the embedding of each tweet or retweet.
Since a user may have multiple retweets, we consider the average of each user's tweets' Bag of Words (BoW) representation, which has a vector size of 7,573, as the NCO proxy. Additionally, we calculate the average of each user's retweet embeddings and use them as the NCE proxy. We employ the BoW representation because our approach yields the lowest estimation error when it is utilized. We use the mean() activation function in this experiment. We report the estimation of the contagion effect using PE-GB because it achieves the best performance in almost all datasets.
\begin{figure*}[ht]
% \vspace{-10pt}
\begin{subfigure}{.24\textwidth}
  \centering
  % include second image
  \includegraphics[width=\textwidth]{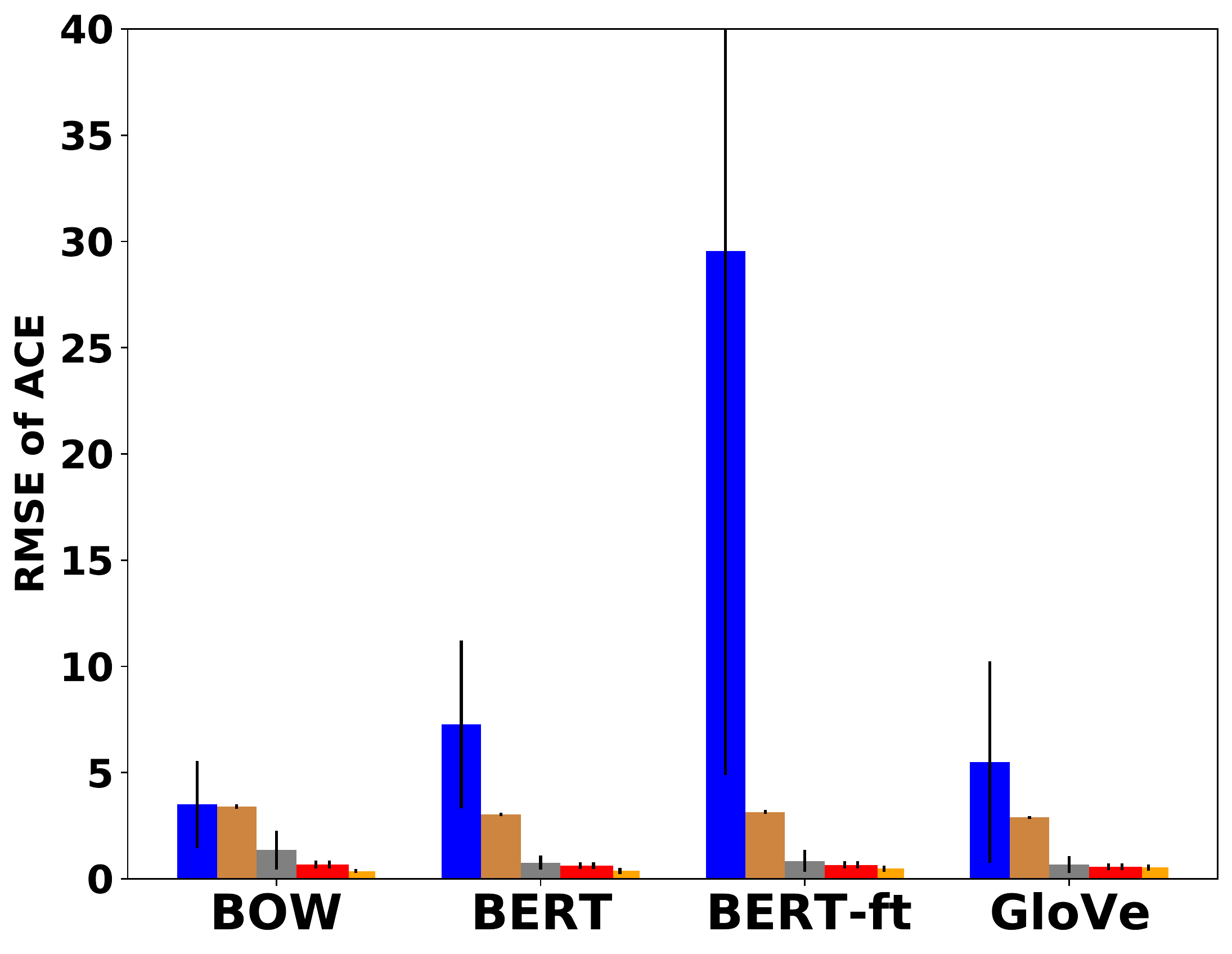}  
  \caption{SAH, $\beta_u \sim \mathcal{N}(0,3)$}
  \label{fig:sub-second}
\end{subfigure}
\begin{subfigure}{.24\textwidth}
  \centering
  % include second image
  \includegraphics[width=\textwidth]{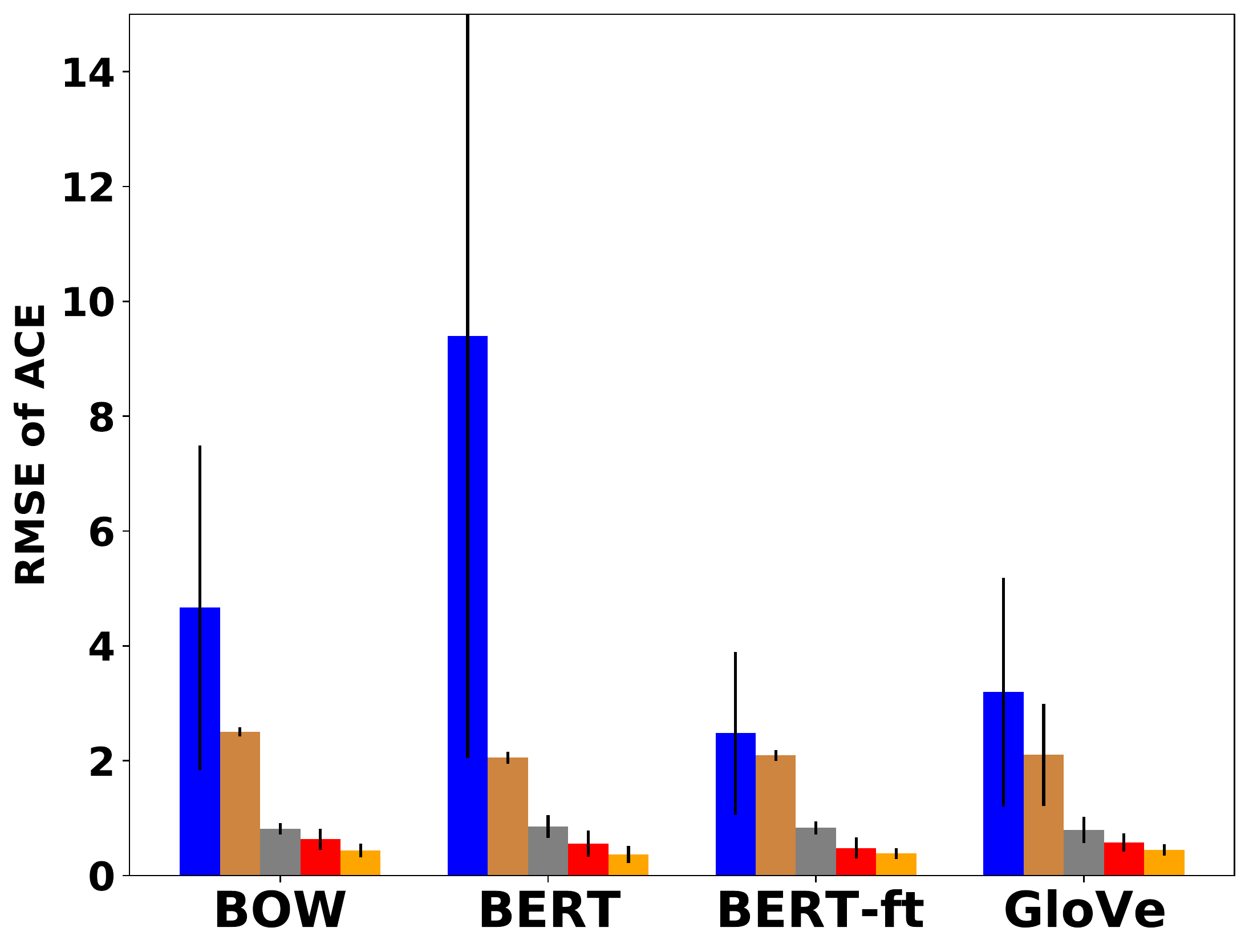}  
  \caption{SAH, $\beta_u \sim \mathcal{N}(5,2)$}
  \label{fig:sub-second}
\end{subfigure}
\begin{subfigure}{.24\textwidth}
  \centering
  % include second image
  \includegraphics[width=\textwidth]{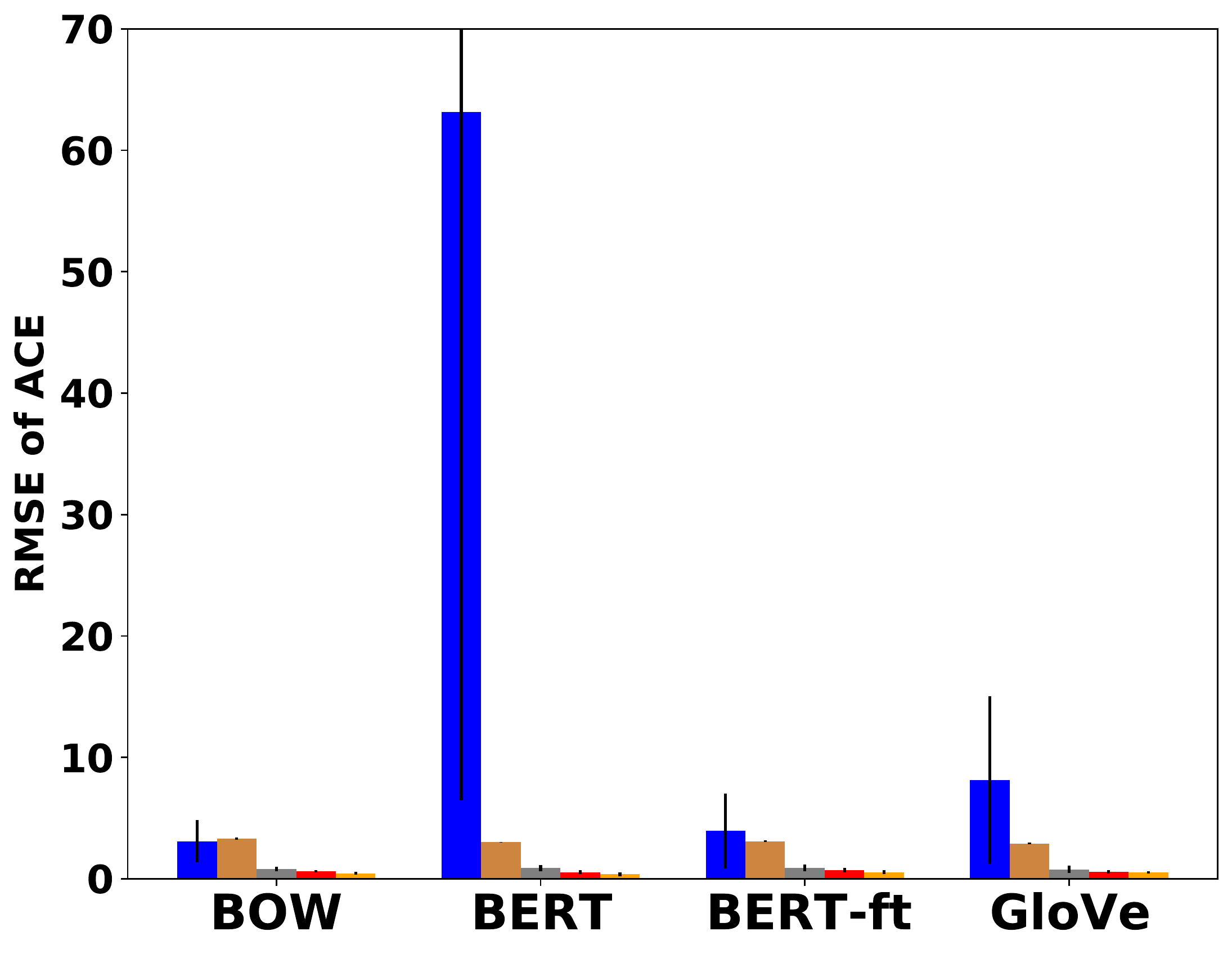}  
  \caption{Hateful, $\beta_u \sim \mathcal{N}(0,3)$}
  \label{fig:sub-second}
\end{subfigure}
\begin{subfigure}{.24\textwidth}
  \centering
  % include second image
  \includegraphics[width=\textwidth]{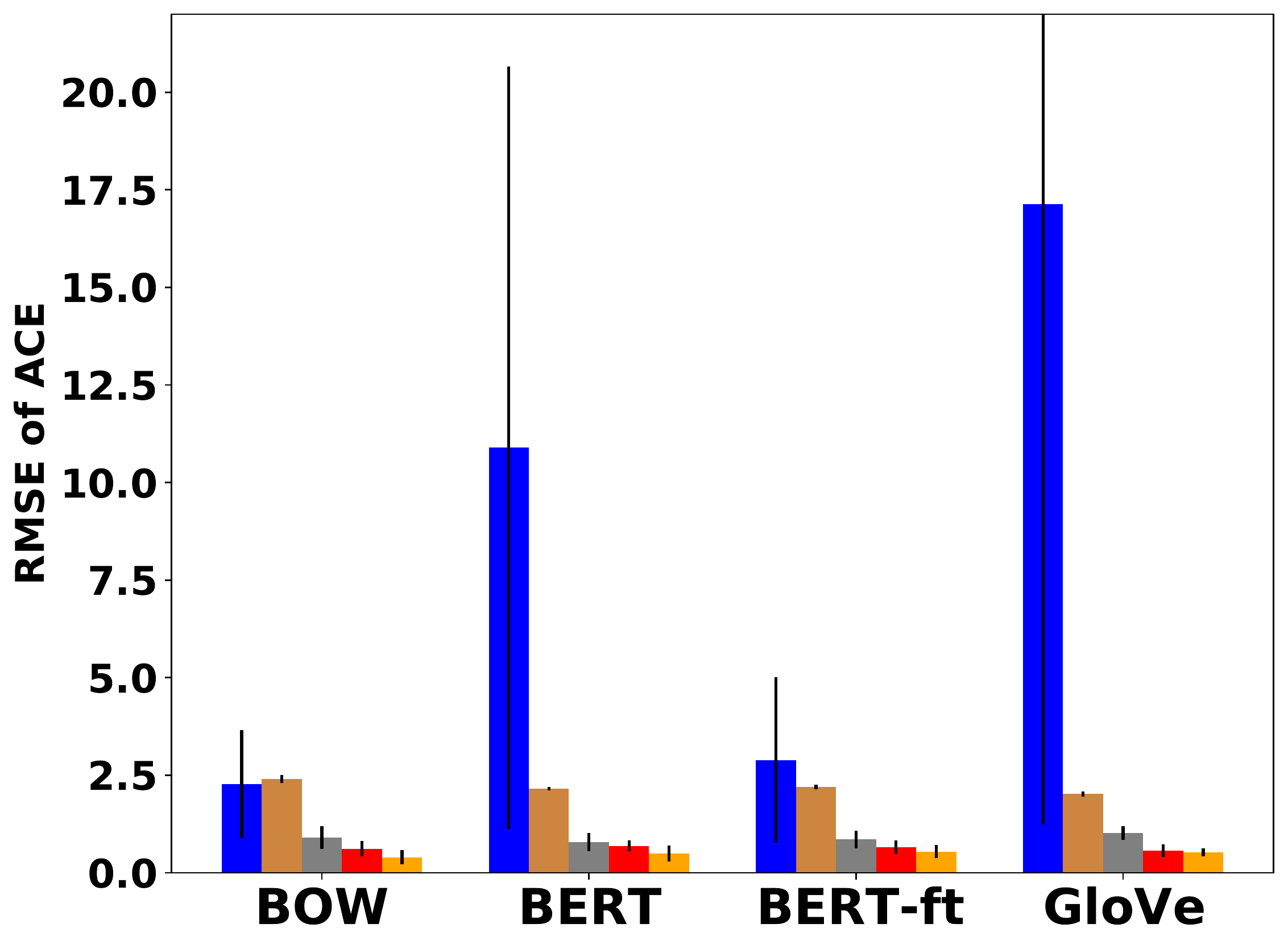}  
  \caption{Hateful, $\beta_u \sim \mathcal{N}(5,2)$}
  \label{fig:sub-second}
\end{subfigure}
%\vspace{-10pt}
\caption{Comparison of RMSE of ACE using various baseline methods in dyadic data. Error bars represent the standard deviation of the estimated effects.}
\label{fig:TSLE_dyad}
\end{figure*}

\begin{figure*}[ht]
 %\vspace{-10pt}
\begin{subfigure}{.46\textwidth}
  \centering
  % include second image
  \includegraphics[width=\textwidth]{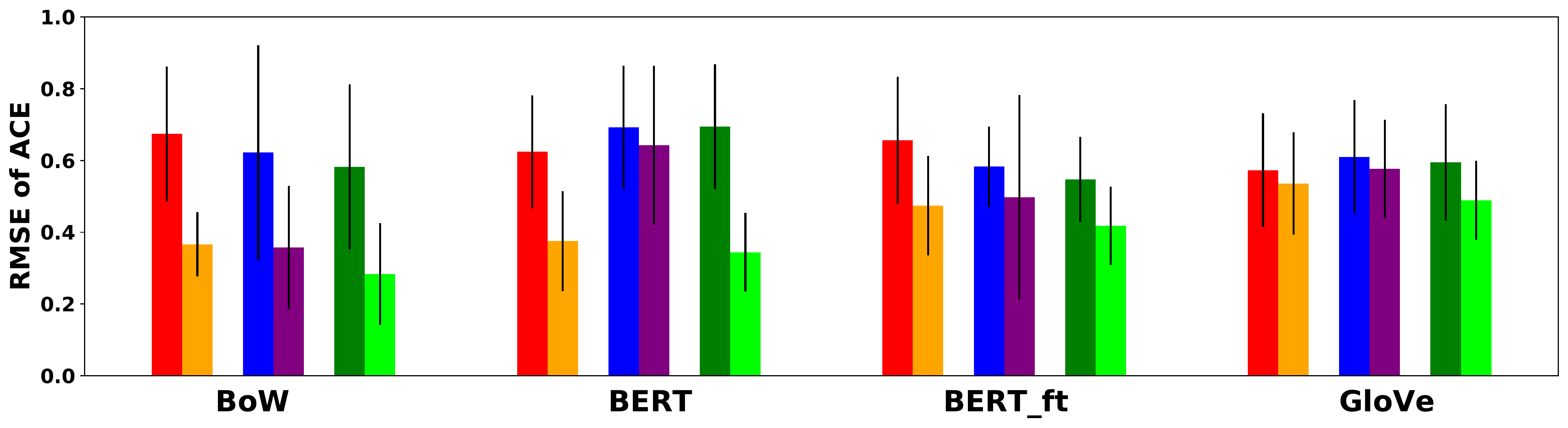} 
  %\caption{SAH}
  \label{fig:sub-second}
\end{subfigure}
\begin{subfigure}{.52\textwidth}
  \centering
  % include second image
  \includegraphics[width=\textwidth]{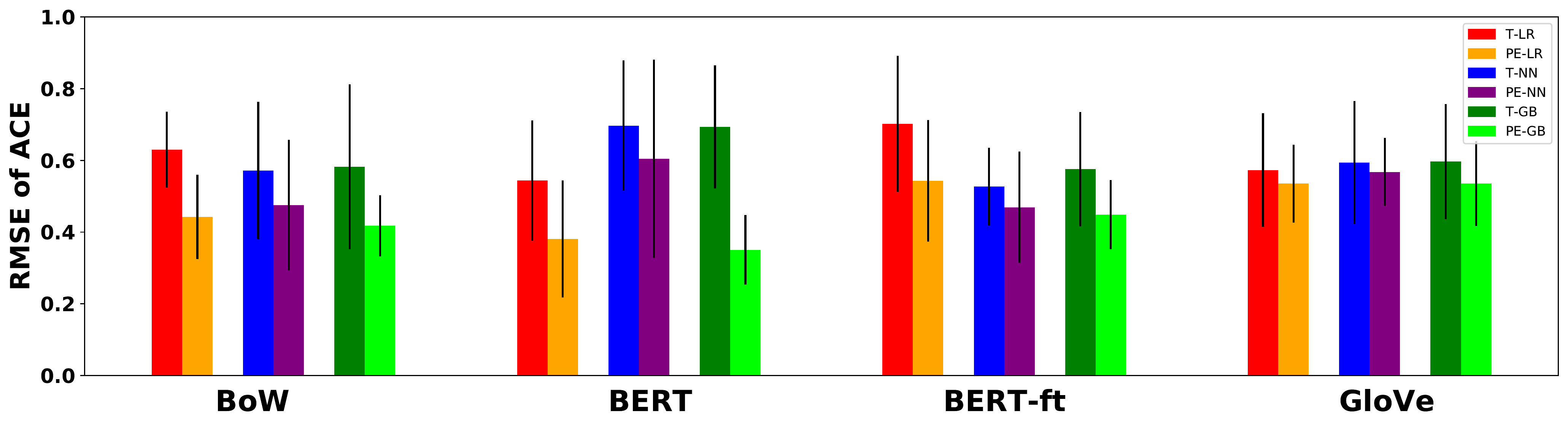}  
  %\caption{Hateful Users}
  \label{fig:sub-second}
\end{subfigure}
\caption{RMSE of ACE in SAH (left) and Hateful Users dataset (right) using dyadic data and $\beta_u \sim \mathcal{N}(0,3)$.
%Bars show the standard deviation of the estimated effect. 
}
\label{fig:SAH_dyad}
%\vspace{-10pt}
\end{figure*}
\begin{figure*}[ht]
%\vspace{-10}
\begin{subfigure}{.24\textwidth}
  %\centering
  % include second image
  \includegraphics[width=\textwidth]{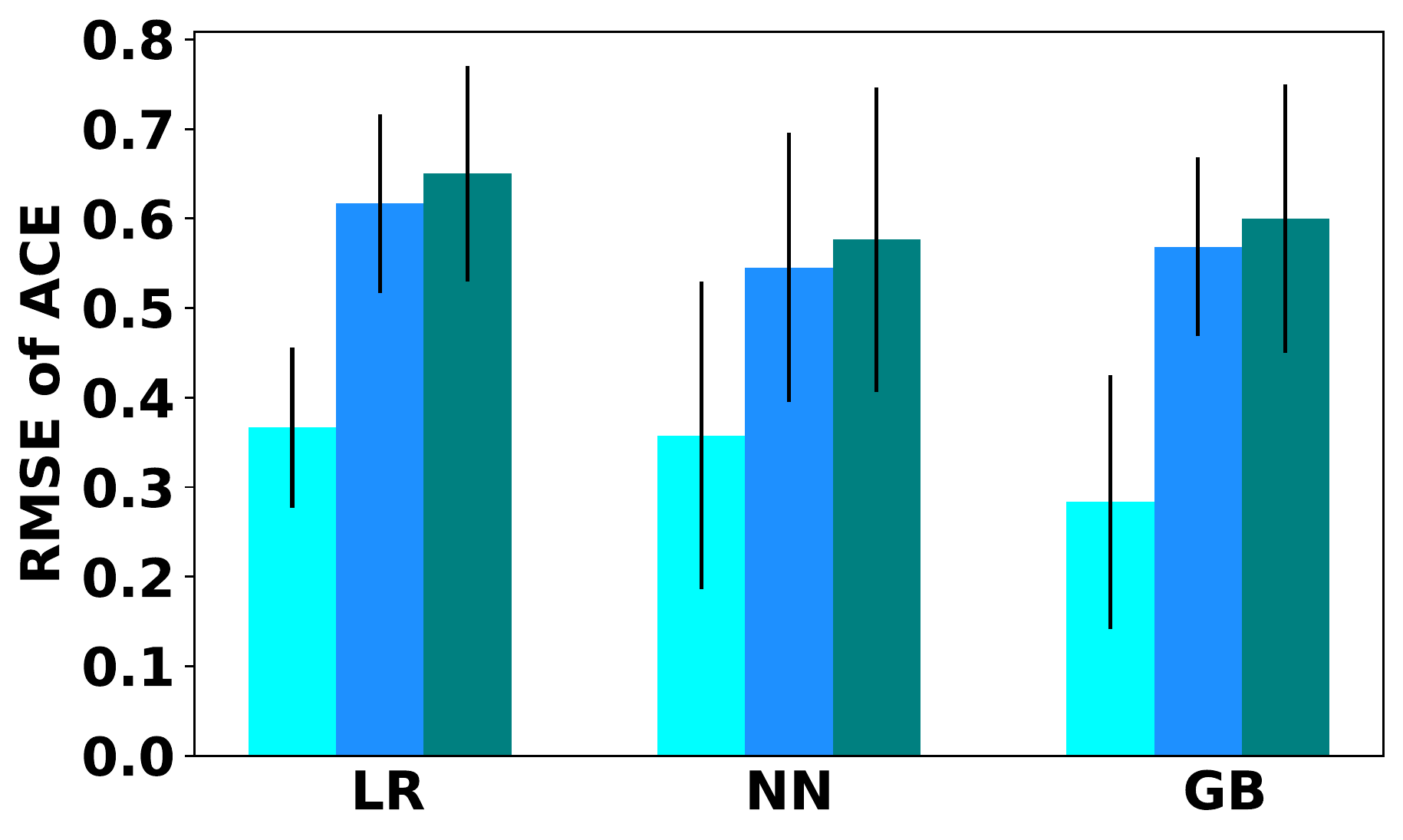}  
  \caption{SAH, $\beta_u \sim \mathcal{N}(0,3)$ }
  \label{fig:sub-second}
\end{subfigure}
\begin{subfigure}{.24\textwidth}
  %\centering
  % include second image
  \includegraphics[width=\textwidth]{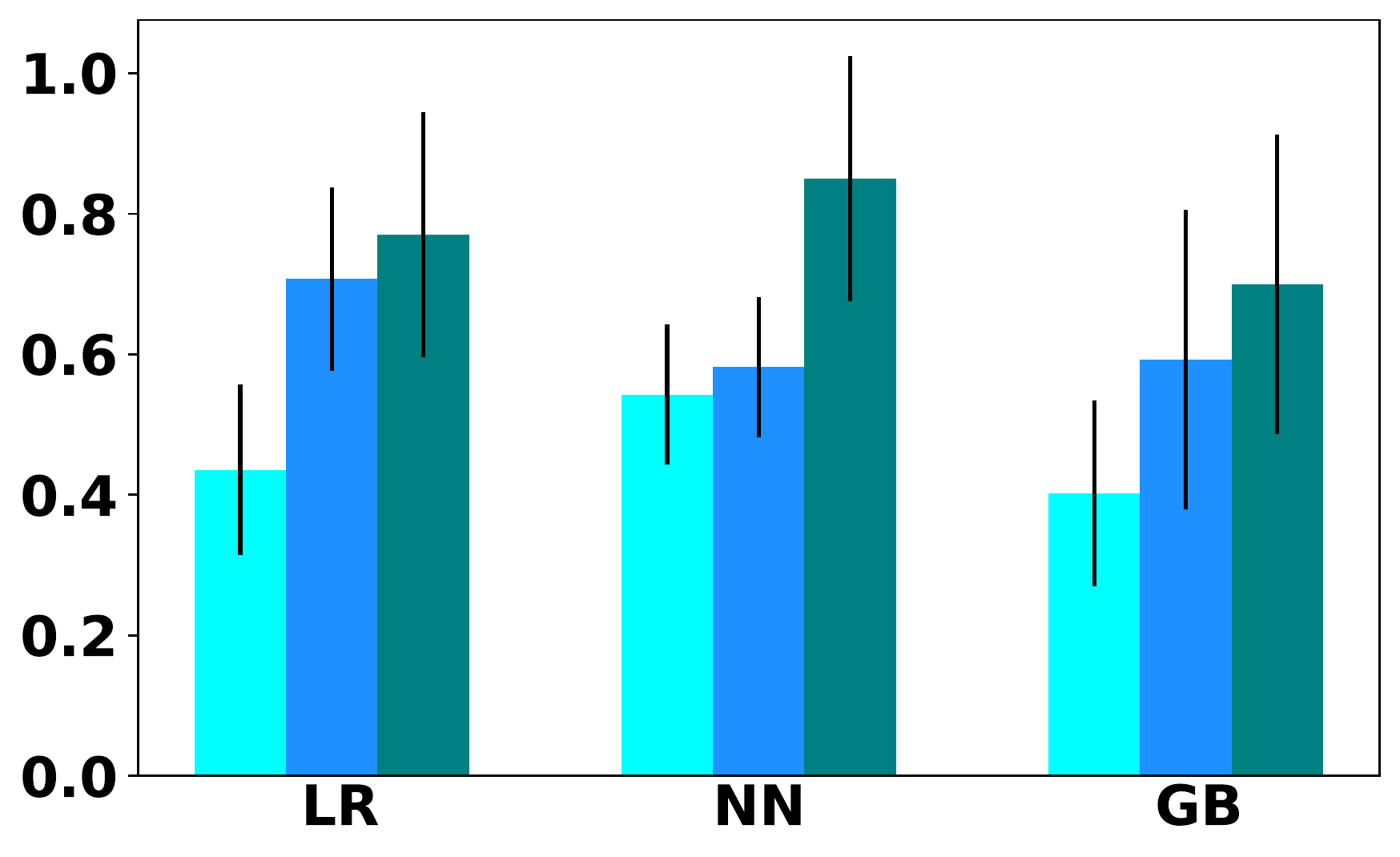}  
  \caption{SAH, $\beta_u \sim \mathcal{N}(5,2)$}
  \label{fig:sub-second}
\end{subfigure}
\begin{subfigure}{.24\textwidth}
  %\centering
  % include second image
  \includegraphics[width=\textwidth]{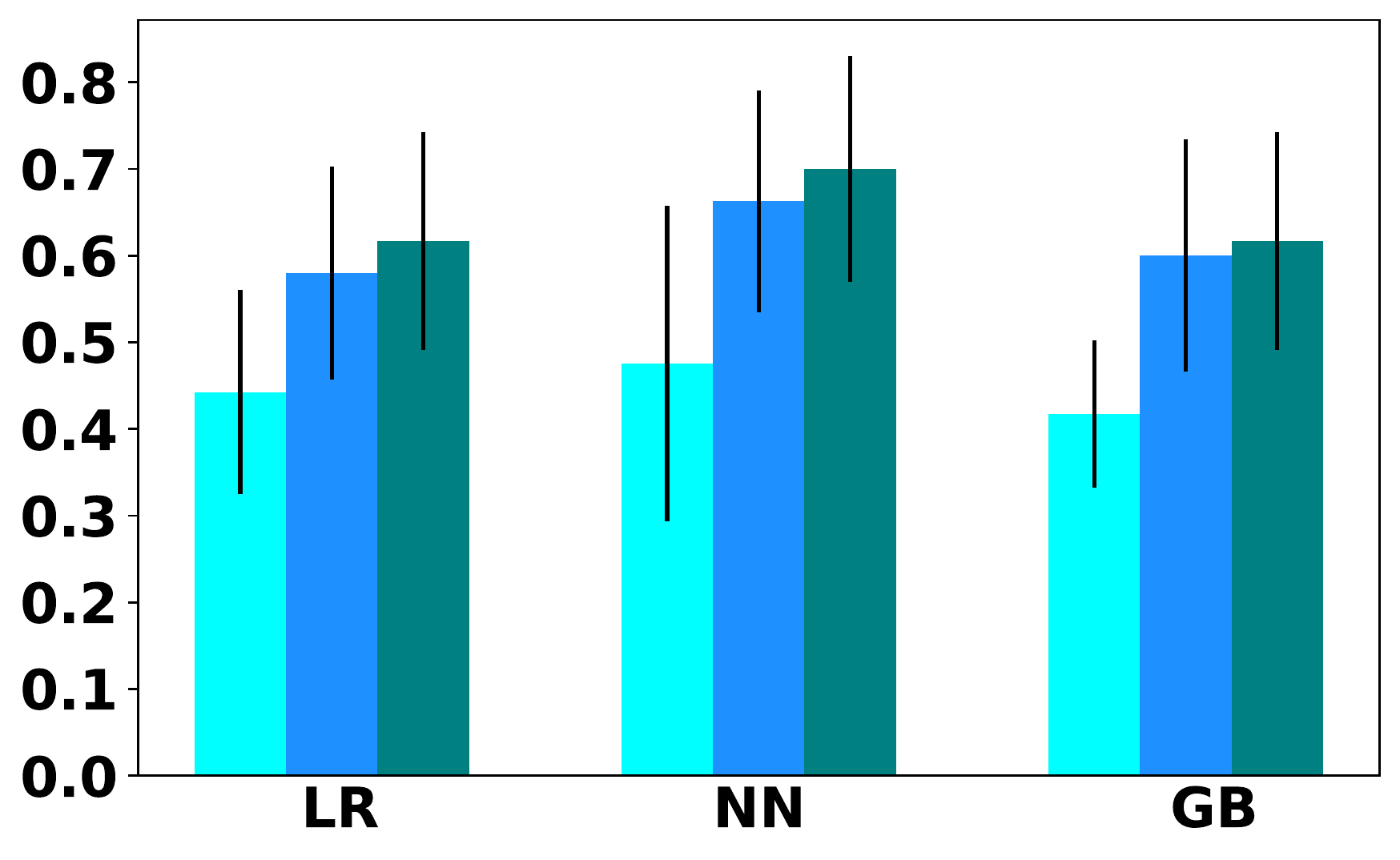}  
  \caption{Hateful, $\beta_u \sim \mathcal{N}(0,3)$}
  \label{fig:sub-second}
\end{subfigure}
\begin{subfigure}{.24\textwidth}
  %\centering
  % include second image
  \includegraphics[width=\textwidth]{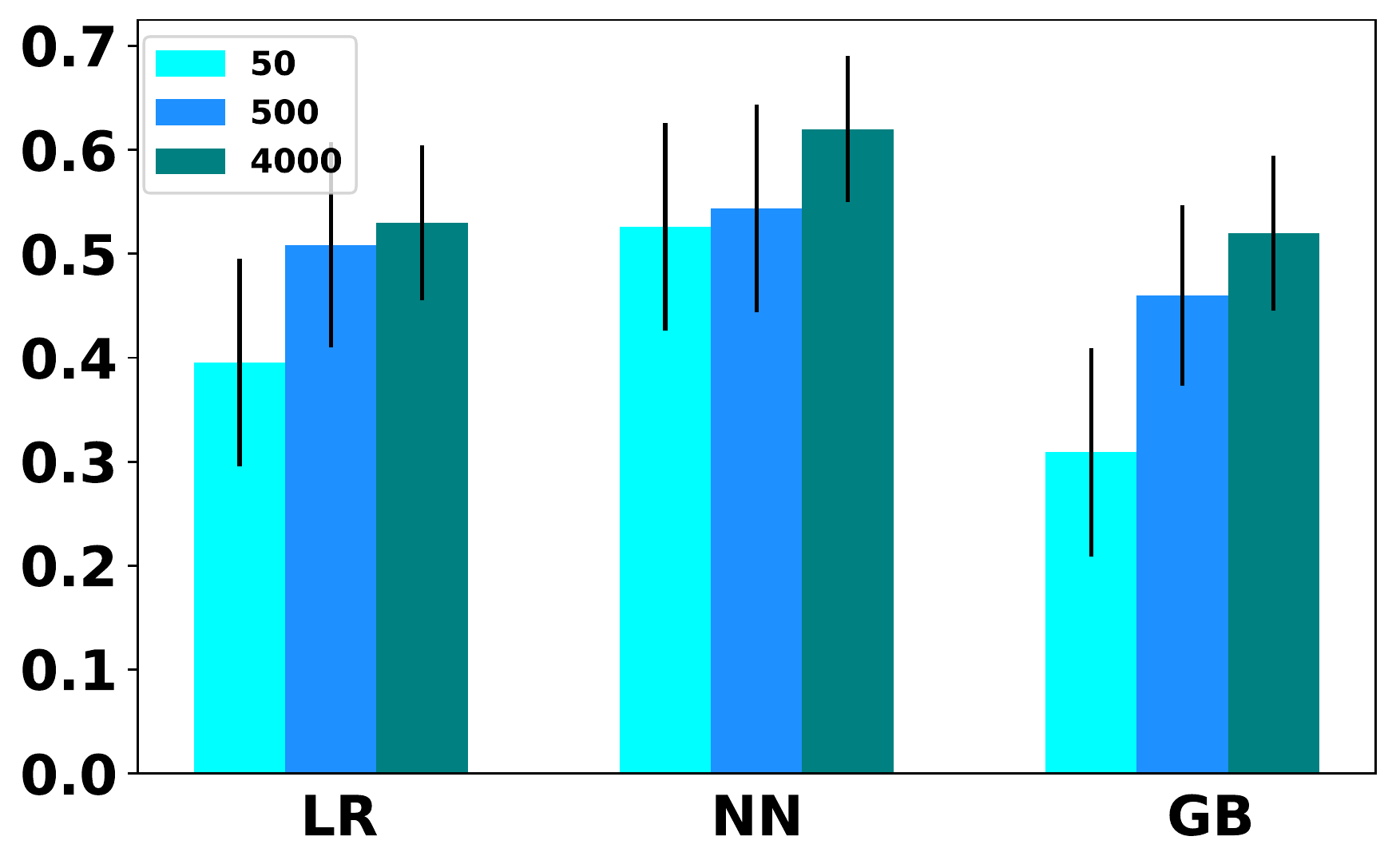}  
  \caption{Hateful, $\beta_u \sim \mathcal{N}(5,2)$}
  \label{fig:sub-second}
\end{subfigure}

%\vspace{-10}
\caption{RMSE of ACE in ProEmb with varying embedding vector dimensions in SAH and Hateful Users dataset with dyadic data. The x-axis represents different types of base-learners used in the counterfactual component of the ProEmb framework.}
\label{fig:AE_dim_dyad}
%\vspace{-10}
\end{figure*}

\subsection{Hyperparameter tuning}
To train the VAEs, discriminators, and MLP models, we conduct a hyperparameter search for the learning rate and the number of epochs. The learning rate is searched within the set $\{0.1, 0.01, 0.001, 0.0001\}$, while the number of epochs is searched within $\{10, 30, 50, 70, 100\}$. The best results are achieved with a learning rate of 0.001 and 50 epochs for both models.
For the VAEs, we search the number of hidden units of the hidden layers in $\{100, 200, 300\}$ and the number of encoder and decoder layers in ${1, 2, 3, 4}$. We select a network with 100 hidden nodes, a 3-layer encoder, and a 3-layer decoder with a ReLU activation function.
In the discriminator component, after hyperparameter search, we determine that four hidden layers, with linear activation functions, produce the best performance. The output layer utilizes a Sigmoid function.
Regarding the MLP, we search for the number of hidden units and the number of fully connected layers. Ultimately, we train an MLP model with two fully connected layers, each containing 125 hidden units. %We set the embedding dimension of the VAEs as the dimension of the unobserved confounder variable in each dataset (20 in SAH and 50 in the Hateful Users dataset). 
In BERT-ft, We use the $AdamW$ optimizer with a learning rate of $2e-5$, max-seq-length of $128$, and batch sizes of $32$.
%Refer to the Supplement for more information on hyperparameter tuning.
 %(i.e., T-learner with LR), T\_GB (i.e., T-learner with GB), and T\_NN (i.e., T-learner with NN) models are deployed to predict the counterfactuals.

\begin{figure*}[ht]
 %\vspace{-10pt}
\begin{subfigure}{.4\textwidth}
  \centering
  % include second image
  \includegraphics[width=\textwidth]{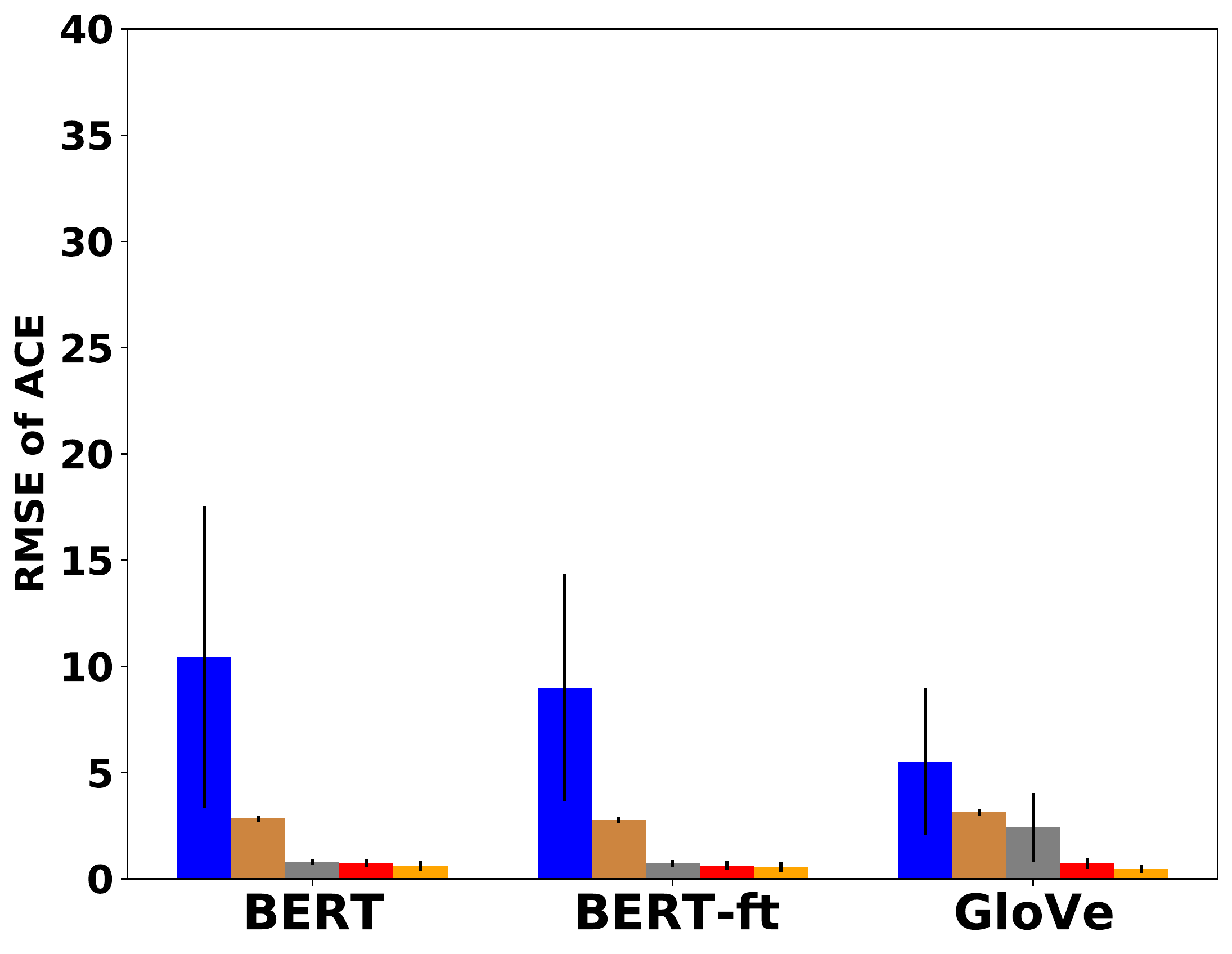} 
  \caption{SAH, $\beta_u \sim \mathcal{N}(0,3)$}
  \label{fig:sub-second}
\end{subfigure}
\begin{subfigure}{.4\textwidth}
  \centering
  % include second image
  \includegraphics[width=\textwidth]{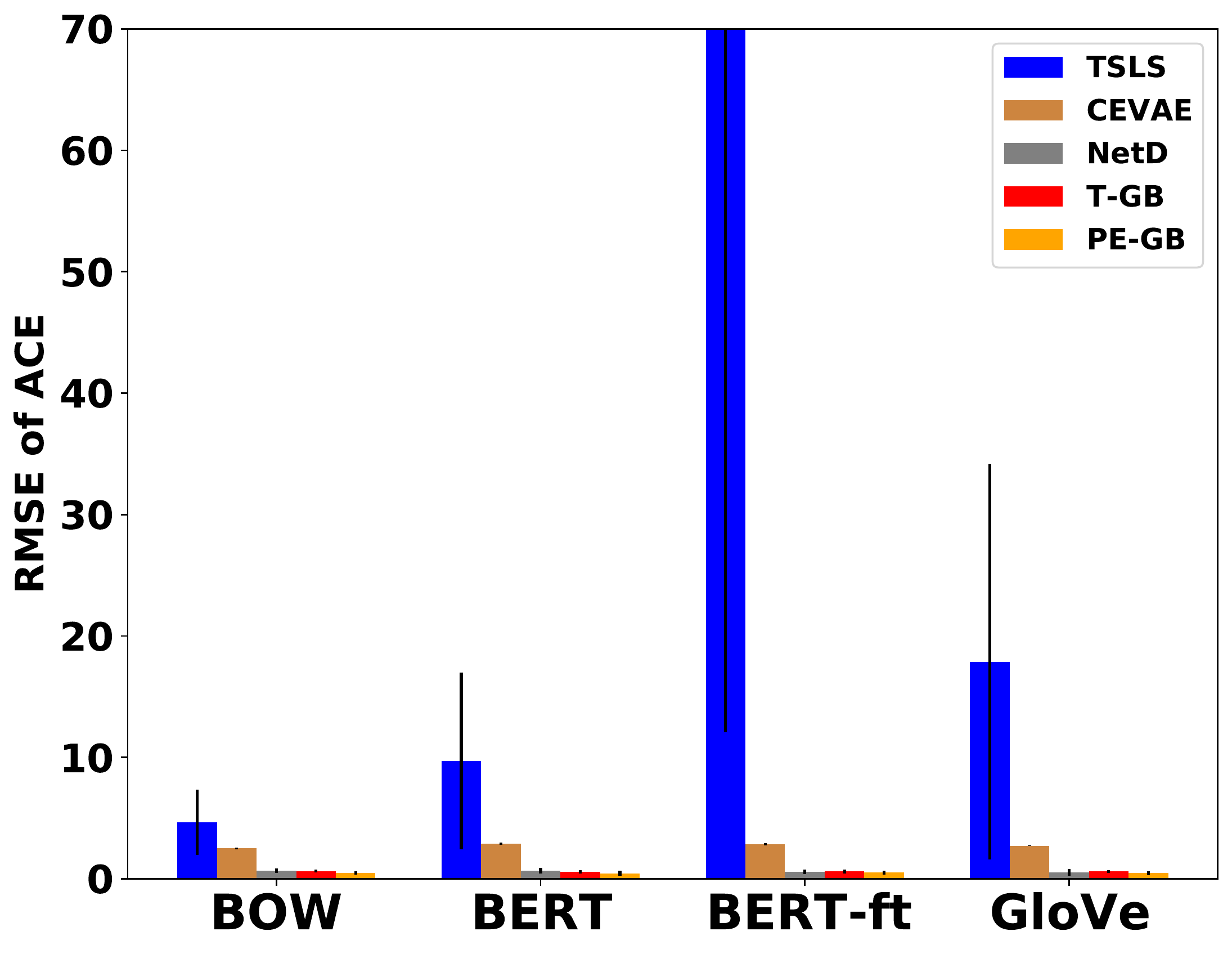}  
  \caption{Hateful Users, $\beta_u \sim \mathcal{N}(0,3)$}
  \label{fig:sub-second}
\end{subfigure}
%\vspace{-20pt}
\caption{RMSE of ACE in SAH (left) and Hateful Users dataset (right) using network data employing mean() activation function and $\beta_u \sim \mathcal{N}(0,3)$.
%Bars show the standard deviation of the estimated effect. 
}
\label{fig:SAH_mean}
%\vspace{-10pt}
\end{figure*}

\begin{figure*}[ht]
% \vspace{-10pt}
\begin{subfigure}{.49\textwidth}
  \centering
  % include second image
  \includegraphics[width=\textwidth]{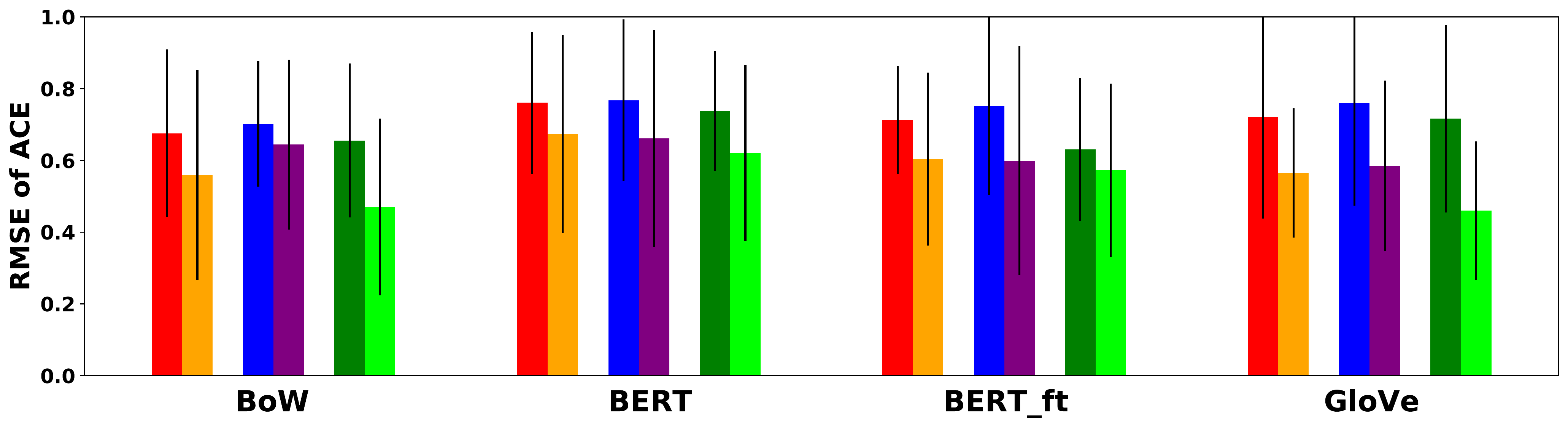} 
  %\caption{SAH}
  \label{fig:sub-second}
\end{subfigure}
\begin{subfigure}{.49\textwidth}
  \centering
  % include second image
  \includegraphics[width=\textwidth]{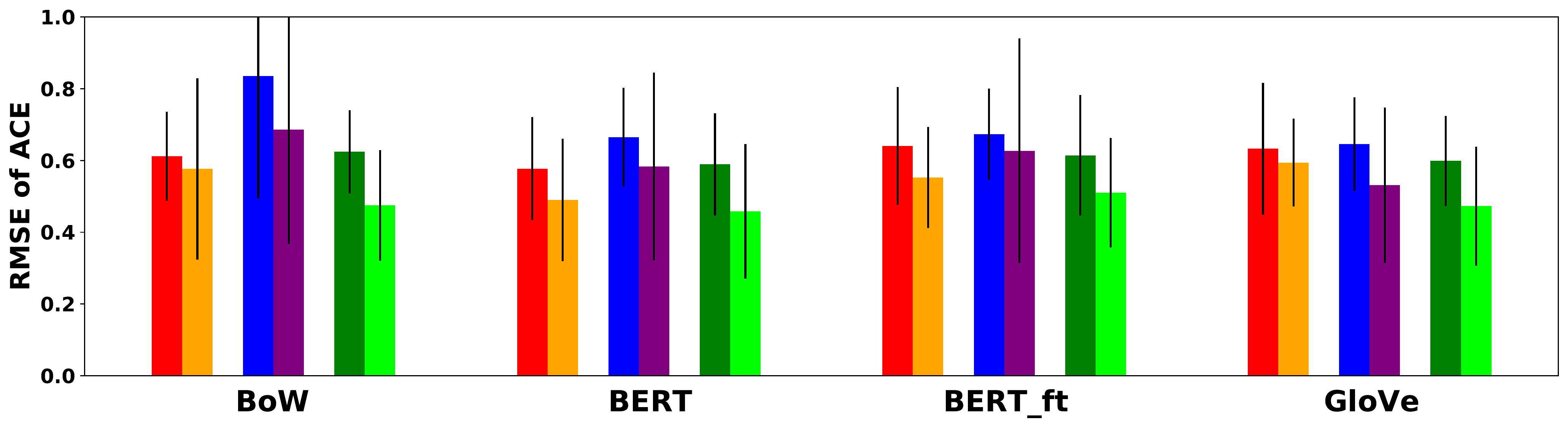}  
  %\caption{Hateful Users}
  \label{fig:sub-second}
\end{subfigure}
%\vspace{-20pt}
\caption{RMSE of ACE in SAH (left) and Hateful Users dataset (right) using network data deploying mean() activation function and $\beta_u \sim \mathcal{N}(0,3)$.
%Bars show the standard deviation of the estimated effect. 
}
\label{fig:SAH_dyad}
%\vspace{-10pt}
\end{figure*}
\begin{figure*}[ht]
\begin{subfigure}{.24\textwidth}
  %\centering
  % include second image
  \includegraphics[width=\textwidth]{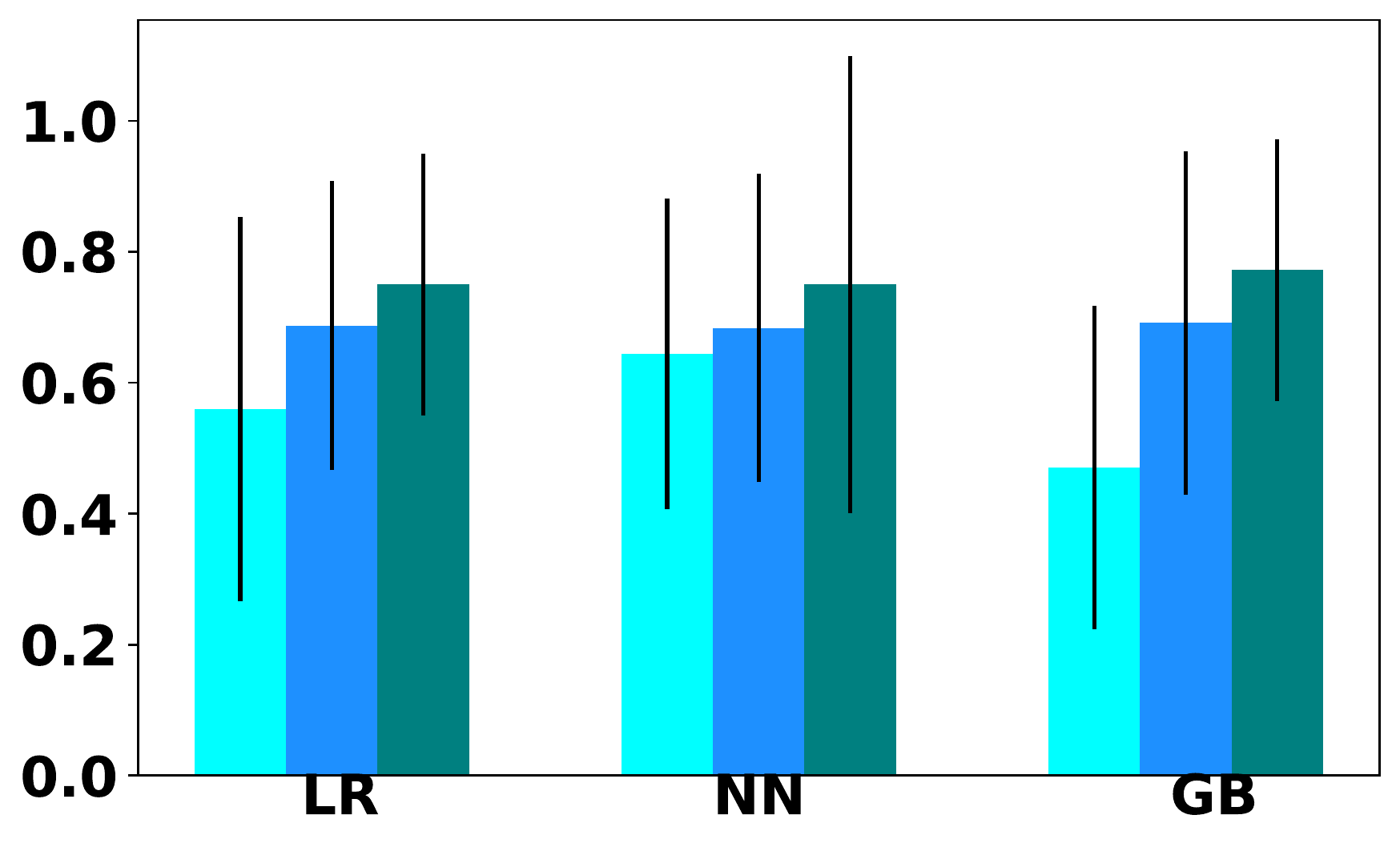} 
  %\vspace{-15pt}
  \caption{SAH}
  \label{fig:sub-second}
\end{subfigure}
\begin{subfigure}{.24\textwidth}
  %\centering
  % include second image
  \includegraphics[width=\textwidth]{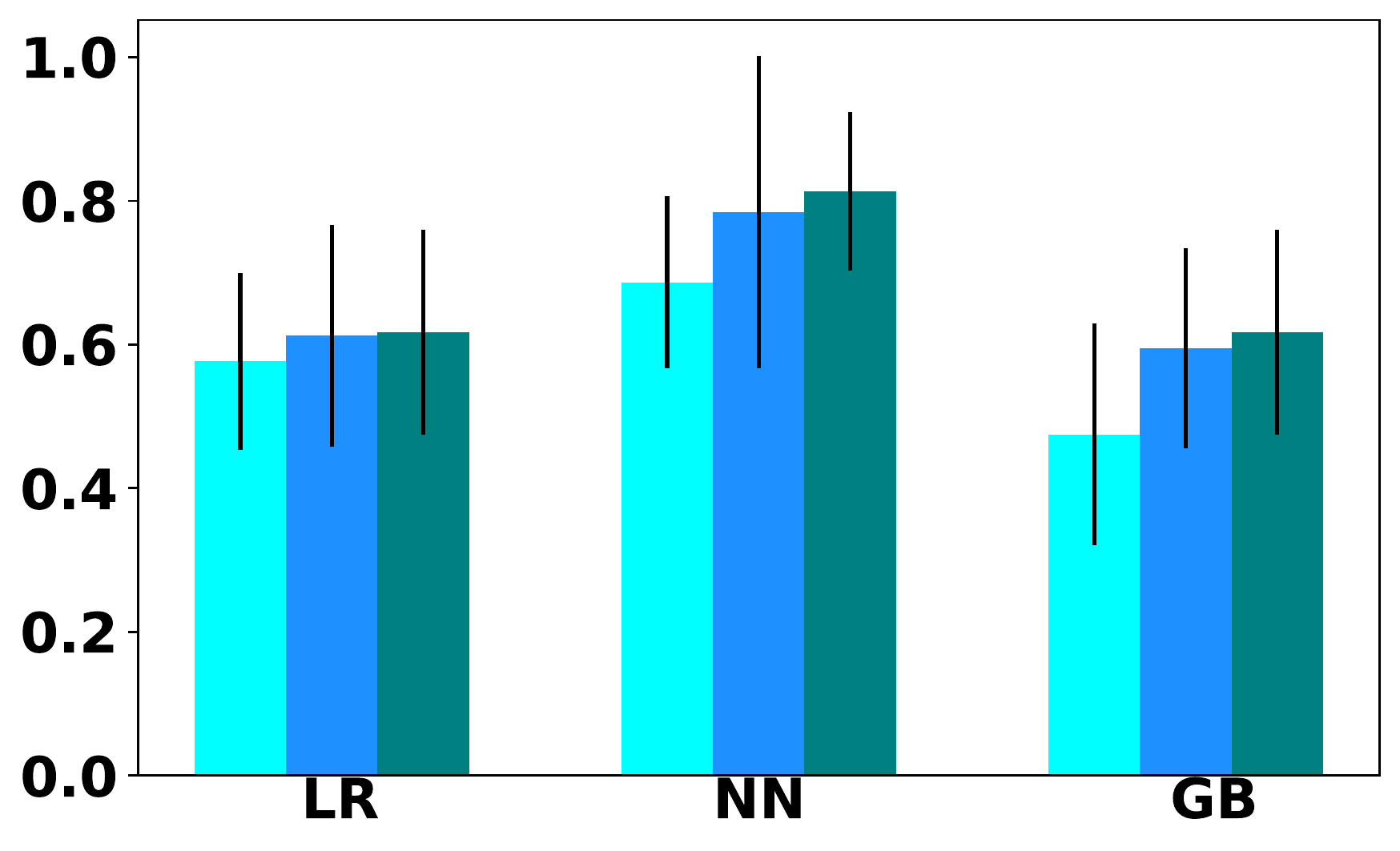} 
  %\vspace{-15pt}
  \caption{Hateful Users}
  \label{fig:sub-second}
\end{subfigure}
\begin{subfigure}{.24\textwidth}
  %\centering
  % include second image
  \includegraphics[width=\textwidth]{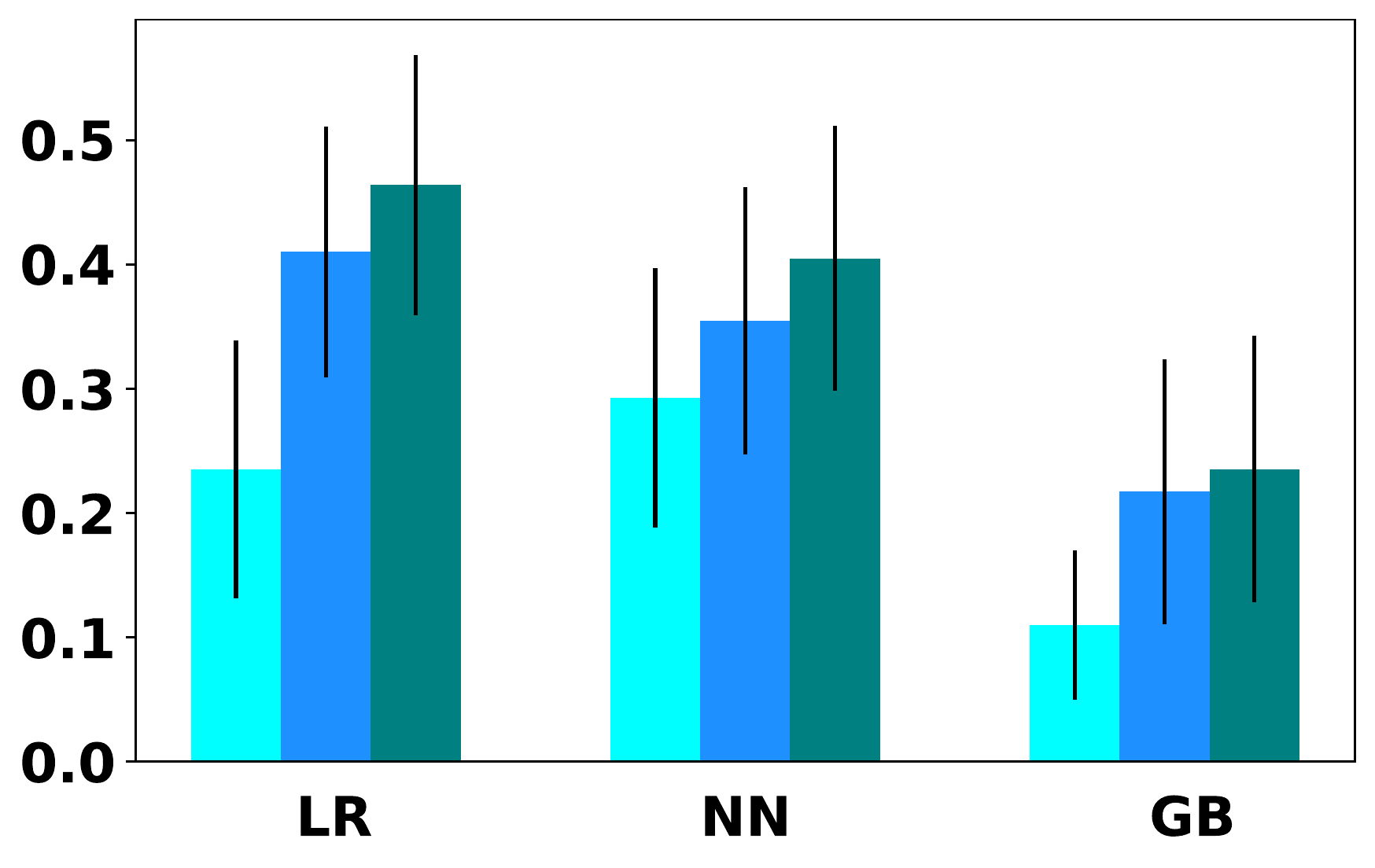} 
  %\vspace{-15pt}
  \caption{BlogCatalog}
  \label{fig:sub-second}
\end{subfigure}
\begin{subfigure}{.24\textwidth}
  %\centering
  % include second image
  \includegraphics[width=\textwidth]{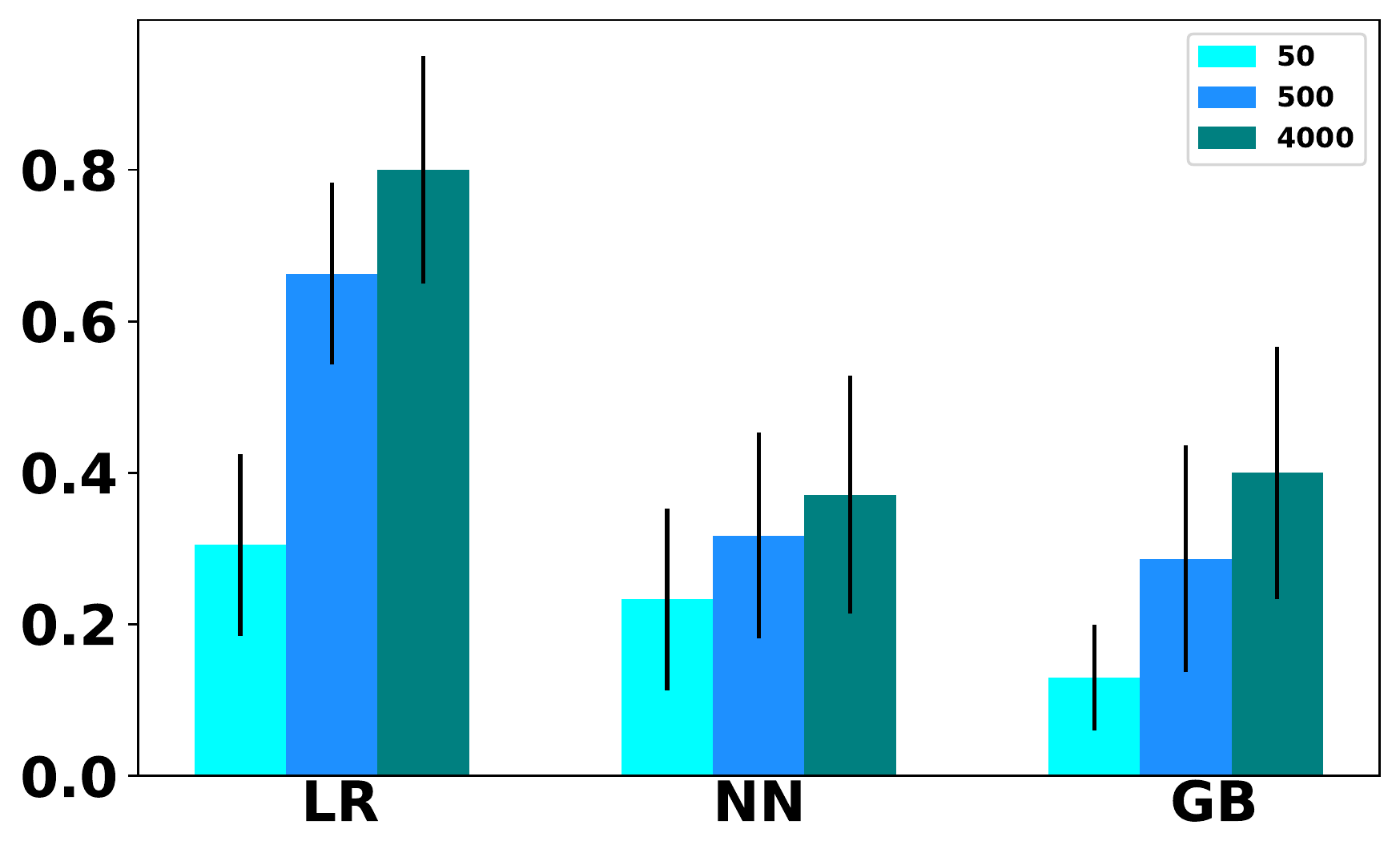}  
  %\vspace{-15pt}
  \caption{Flickr}
  \label{fig:sub-second}
\end{subfigure}
%\vspace{-10pt}
\caption{RMSE of ACE in ProEmb with varying embedding vector dimensions in four datasets and $\beta_u \sim \mathcal{N}(0,3)$ employing mean() activation function. The x-axis represents different types of base-learners used in the counterfactual component of the ProEmb framework.}
%\vspace{-15pt}
\label{fig:AE_dim_mean}
\end{figure*}
\subsection{Results}
\subsubsection{Contagion effect estimation evaluation for dyadic data}
In this section, we present the results of a comparative analysis among various methods used to estimate peer contagion effects in the SAH and Hateful Users datasets, based on dyad data with distinct $\beta_u$ vectors. These findings align consistently with the results from the network data. As depicted in Figure \ref{fig:TSLE_dyad},  \textit{TSLS} exhibits the highest estimation error when compared to other baseline methods, whereas \textit{T-LR} demonstrates more favorable performance. The results illustrated in Fig. \ref{fig:SAH_dyad} further show that \textit{PE-LR} and \textit{PE-G} methods outperform other approaches in terms of reducing contagion effect estimation errors.

In terms of sensitivity to the dimension of the embedding, as Fig. \ref{fig:AE_dim_dyad} shows, by increasing the number of embedding dimensions from 20 to 4,000 in SAH and from 50 to 4000 in the Hateful Users dataset, the estimation error of the ProEmb method increases. This finding confirms the necessity of applying dimension-reduction techniques when dealing with high-dimensional data.

\subsubsection{Contagion effect estimation evaluation for network data}
In this experiment, we use mean() activation function for h(). Figure \ref{fig:SAH_mean} demonstrates that \textit{TSLS} receives the highest bias and variance in both SAH and Hateful Users datasets. Similar to experiments with max() activation function, we observe that our framework, \textit{PE-GB} achieves the best performance with different feature representation techniques. Regarding sensitivity to the dimension of the embedding, we obtain consistent results with data on dyads and network data with max() activation function. In all datasets, by increasing the dimension of the emebeddings, the estimation error of the method increases (Fig. \ref{fig:AE_dim_mean}).

\end{document}